\documentclass[12pt, a4paper, logo, copyright]{googledeepmind}

\usepackage[sort&compress, square, numbers]{natbib}

\usepackage[utf8]{inputenc} 
\usepackage[T1]{fontenc}    
\usepackage{graphicx}
\usepackage[font=small]{caption}  
\usepackage{hyperref}       
\usepackage{url}            
\usepackage{booktabs}       
\usepackage{amsfonts}       
\usepackage{nicefrac}       
\usepackage{microtype}      
\usepackage{xcolor}         
\usepackage{wrapfig}
\usepackage{subcaption}
\usepackage{tabularx}
\usepackage{amsmath}
\usepackage{amssymb}
\usepackage{algorithm}
\usepackage{algpseudocode}
\usepackage{mathtools}
\usepackage{copyrightbox}
\usepackage{bbm}
\usepackage{amsthm}
\usepackage{chngcntr}
\usepackage{enumitem}
\usepackage{fancyvrb,fvextra}
\usepackage{xspace}
\usepackage{comment}
\usepackage{placeins}
\usepackage{titlesec}
\usepackage[capitalise]{cleveref}
\crefname{section}{Sec.}{Secs.}
\crefname{table}{Tab.}{Tabs.}
\usepackage{setspace}
\usepackage{lineno}
\usepackage{xr}

\titlespacing{\paragraph}{0pt}{0pt}{0.5em}
\titlespacing{\section}{0pt}{0pt}{0pt}
\titlespacing{\subsection}{0pt}{0pt}{0pt}

\newcommand{\finaledit}[1]{#1}

\usepackage{amsmath,amsfonts,bm}

\def\eqref#1{equation~\ref{#1}}

\def\1{\bm{1}}

\def\vb{{\bm{b}}}

\def\vx{{\bm{x}}}

\def\vz{{\bm{z}}}

\def\mS{{\bm{S}}}

\def\mW{{\bm{W}}}
\def\mX{{\bm{X}}}
\def\mY{{\bm{Y}}}
\def\mZ{{\bm{Z}}}

\DeclareMathAlphabet{\mathsfit}{\encodingdefault}{\sfdefault}{m}{sl}
\SetMathAlphabet{\mathsfit}{bold}{\encodingdefault}{\sfdefault}{bx}{n}

\newcommand{\softmax}{\mathrm{softmax}}

\DeclareMathOperator*{\argmax}{arg\,max}
\DeclareMathOperator*{\argmin}{arg\,min}

\newcommand{\norm}[1]{\left\lVert#1\right\rVert}

\title{Aligning Machine and Human Visual Representations across Abstraction Levels}

\correspondingauthor{lukas.muttenthaler@tu-berlin.de; klaus-robert.m\"uller@tu-berlin.de, lampinen@google.com}


\reportnumber{} 

\author[1,2,3,4,*]{Lukas Muttenthaler}
\author[1]{Klaus Greff}
\author[2,3,5]{Frieda Born}
\author[5,6]{Bernhard Spitzer}
\author[7]{Simon Kornblith}
\author[1]{Michael C. Mozer}
\author[1,2,3,8,9,*]{Klaus-Robert M\"uller}
\author[1]{Thomas Unterthiner}
\author[1,*]{Andrew K. Lampinen}

\affil[1]{Google DeepMind}
\affil[2]{Machine Learning Group, Technische Universit\"at Berlin}
\affil[3]{BIFOLD, Berlin Institute for the Foundations of Learning and Data, Berlin, Germany}
\affil[4]{Max Planck Institute for Human Cognitive and Brain Sciences, Leipzig, Germany}
\affil[5]{Max Planck Institute for Human Development, Germany}
\affil[6]{TUD Dresden University of Technology, Germany}
\affil[7]{Anthropic}
\affil[8]{Department of Artificial Intelligence, Korea University, Seoul}
\affil[9]{Max Planck Institute for Informatics, Saarbr\"ucken, Germany}
\affil[*]{corresponding authors}

\begin{abstract}
Deep neural networks have achieved success across a wide range of applications, including as models of human behavior and neural representations in vision tasks~\cite{khaligh2014deep, peterson2019}. However, neural network training and human learning differ in fundamental ways, and neural networks often fail to generalize as robustly as humans do~\cite{lake2017building, geirhos2018generalisation}, 
raising questions regarding the similarity of their underlying representations. \emph{What is missing for modern learning systems to exhibit more human-\finaledit{aligned} behavior?} We highlight a \emph{key misalignment} between vision models and humans: whereas human conceptual knowledge is hierarchically organized from fine- to coarse-scale distinctions \finaledit{\citep[e.g.][]{hebart2020revealing}}, model representations do not accurately capture all these levels of abstraction.
To address this misalignment, we first train a teacher model to imitate human judgments, then transfer human-\finaledit{aligned} structure from its representations to refine the representations of pretrained state-of-the-art vision foundation models via finetuning.
These human-aligned models more accurately approximate human behavior and uncertainty across a wide range of similarity tasks, including a new dataset of human judgments spanning multiple levels of semantic abstractions. They also perform better on a diverse set of machine learning tasks, increasing generalization and out-of-distribution robustness. Thus, infusing neural networks with additional human knowledge yields a \emph{best-of-both-worlds representation} that is both more consistent with human \finaledit{cognitive judgments} and more practically useful, thus paving the way toward more robust, interpretable, and human-\finaledit{aligned} artificial intelligence systems.
\end{abstract}

\definecolor{coarse-green}{RGB}{137, 190, 146}
\definecolor{fine-blue}{RGB}{37, 108, 137}
\definecolor{boundary-red}{RGB}{133, 36, 89}

\begin{document}
\maketitle
\doublespacing 


\section{Introduction}
\label{sec:introduction}

\begin{figure}[t!]
    \centering
    \includegraphics[width=\textwidth]{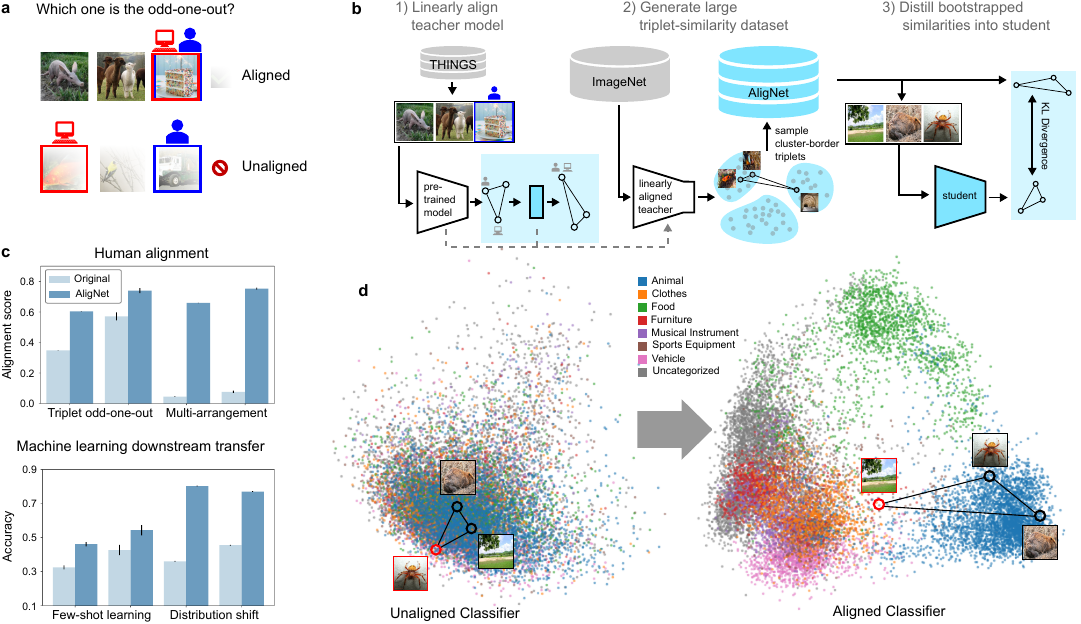}
    \caption[Overview]{\doublespacing\textbf{a}: An example of the triplet odd-one-out task where a human and a neural network model choose the same (top) and a different (bottom) odd-one-out image respectively. \textbf{b}: The different parts of the AligNet framework depicted from end to end. First, we develop a teacher model of human judgments using the THINGS dataset (first panel). Second, we apply this model to ImageNet and cluster its latent representations into semantically meaningful categories (second panel). This allows us to generate arbitrarily many similarity judgments. Third, the obtained human-\finaledit{aligned} similarity structure information is distilled into a student vision foundation model using a novel loss function (third panel). \textbf{c}: Representative human alignment (top panel) and machine learning downstream (bottom panel) results show significant performance improvements of the aligned over the non-aligned version of the ViT-B classifier (up to 123\% of relative improvement for ML downstream transfer). \textbf{d}: 2D latent space projection for visualizing the change in the representations after alignment. While the representations of a standard ViT-B classifier model are unstructured and categories overlap, after alignment the representations are grouped into meaningful categories.
    }
    \label{fig:fig1}
\end{figure}

While deep learning has recently driven rapid progress in areas of artificial intelligence such as natural language processing \citep{lecun2015deep, brown2020language}
and computer vision 
\citep{radford2021learning, dosovitskiy2020image, zhai2023sigmoid, kirillov2023segment},
even the best of these systems often fail in ways that humans would not \citep{geirhos2018generalisation, lapuschkin2019unmasking, geirhos2020shortcut, thrush2022winoground,  bowers2022deep, muttenthaler2023improving}.
These failures have led to renewals  \citep{lake2017building,bowers2022deep} of older arguments \citep{fodor1988connectionism,tenenbaum2011grow} that neural networks lack essential ingredients of human intelligence. 
How can we build systems that produce more human-\finaledit{aligned} behavior?

Human perception is robust 
and generalizes across different visual settings 
\citep{lake2015human-level, roads2017improving, geirhos2018generalisation}.
However, model performance declines---often drastically---if the data distribution \finaledit{shifts} between training and test sets 
\citep[e.g.,][]{hendrycks2021faces}.
This lack of robustness in \finaledit{vision model} representations poses a challenge for downstream applications that require generalization 
\citep[e.g.,][]{pooch2019trust}. In addition, humans tend to be well-calibrated---for example, when they are asked to judge visual similarity \citep{roads2017improving}---that is, humans' (un)certainty tends to correlate with their (in)accuracy.
AI systems, on the other hand, are often \emph{overconfident} and show high certainty 
even when their predictions are incorrect \citep{minderer2021revisiting}. Thus, many differences remain to be reconciled before we can ultimately achieve human-like artificial intelligence.

\finaledit{Here}, we highlight a key misalignment between humans and deep learning models that may underlie some of these differences: model representations tend to fail to capture the \finaledit{full} 
multi-level conceptual structure of human \finaledit{knowledge}. Although model representations successfully encode the \finaledit{\emph{local}} human-perceived similarity structure among \finaledit{closely-related} entities that share 
\finaledit{(e.g. different dog breeds)}, the \emph{global} relationships between concepts \finaledit{with more abstract semantic relations} 
(e.g., dogs and fish, which are both animate but \finaledit{visually dissimilar}) are modeled much less systematically.
%
%
Human neural representations, however, are organized by global features like animacy 
\citep{cichy2019,hebart2020revealing}, and at multiple finer scales that capture nuanced semantic relationships \citep{connolly2012representation, cichy2019, king2019}. Could the lack of global organization in the representations of deep learning models across levels of the conceptual hierarchy contribute to the aforementioned weaknesses of these models?

\finaledit{A challenge for addressing this misalignment is that collecting representative datasets of human judgments is challenging and expensive. We therefore propose a method for synthesizing simulated (approximately) human-aligned similarity judgments via a surrogate teacher model---a large foundation model which we align using an affine transformation \citep{muttenthaler2023human} and uncertainty distillation on a small existing dataset \citep{hebart2019things}. We use this surrogate to produce the \emph{AligNet Dataset}---human-aligned pseudolabels \citep[cf.][]{lee2013pseudo-label} from the surrogate model for triplets sampled from ImageNet \citep{russakovsky2014imagenet} using a novel clustering-based data grouping method. We finetune various vision foundation models on AligNet using a novel similarity-space distillation objective. These models show substantially more human-aligned predictions on various cognitive science tasks---including \emph{Levels}, a novel dataset of human semantic judgments reflecting multiple levels of abstraction. Furthermore, these aligned models show improved accuracy and out-of-distribution robustness across many downstream machine learning tasks, thus showing the improved generalizability of the aligned model representations. We release our aligned models and training and evaluation datasets.}

In summary, our work contributes to better understanding \finaledit{a} key difference between artificial and natural intelligence. Moreover, our results illustrate a principle for aligning models \finaledit{to} humans---focusing on the multi-\finaledit{scale} relational structure of human knowledge---that may \finaledit{contribute} to the more general problem of achieving human-\finaledit{aligned} artificial intelligence.


\section{Results}

\finaledit{
To build foundation models with more human-\finaledit{aligned} behavior we inject additional supervision about human behavior into the model representations, using a surrogate \emph{teacher} model: a vision foundation model whose representations are linearly transformed to approximate human judgments and uncertainty on the THINGS dataset \citep{hebart2019things}. We use clusters from this teacher model's representations to sample triplets from ImageNet \citep{russakovsky2014imagenet} and soft-label them using distances in the teachers representation space, then distill these soft-labels into the student representations via a Kullback-Leibler divergence objective. For details see Sec.~\ref{sec:methods_alignet}.

\subsection{Toward more human-aligned models}
\label{sec:human_alignment}

\begin{figure}[ht!]
    \centering 
    \includegraphics[width=\textwidth]{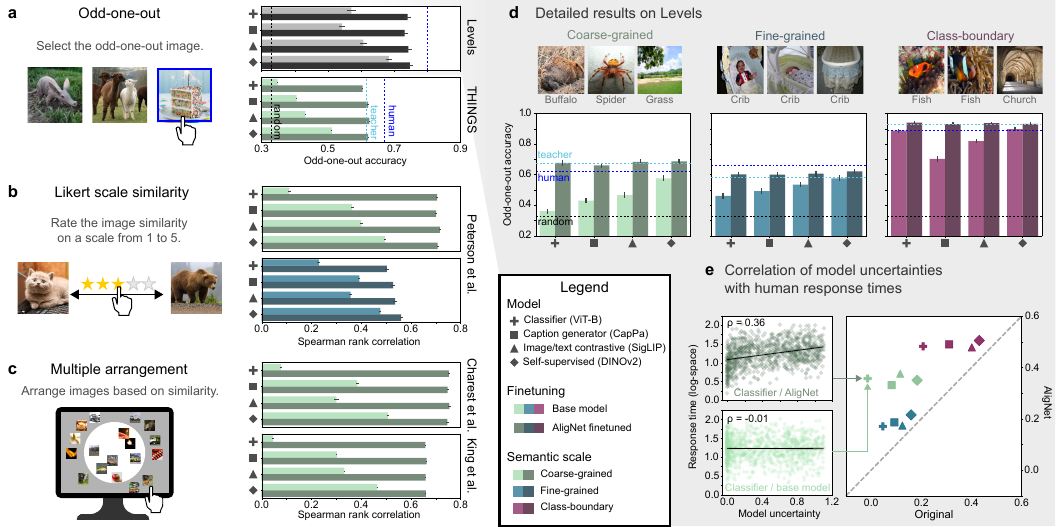}
    \caption[Human alignment results]{\doublespacing Human alignment results. \textbf{a}: Odd-one-out accuracies on the THINGS dataset and performance averaged across all three levels of abstraction for Levels. \textbf{b}: Spearman rank correlations for the human response datasets from \citet{peterson2018evaluating} for the {\color{coarse-green}{coarse-grained}} \texttt{various} category and averaged across all {\color{fine-blue}{fine-grained}} single domain categories. \textbf{c}: Spearman rank correlations for the multi-arrangement datasets from \citet{king2019} and \citet{cichy2019} respectively. \textbf{d}: Odd-one-out accuracies on our datasets shown individually for the three levels of abstraction. \textbf{e}: Spearman rank correlation of model uncertainties and human response times. Model uncertainties are modeled as discrete Shannon entropy of the pairwise similarities in a triplet.}
    \label{fig:fig2}
\end{figure}

\noindent\textbf{THINGS Triplet Odd-one-out.}
We first validate that our teacher model performs well on the test data for the THINGS dataset \citep{hebart2019things} used in training; as expected, the teacher achieves high performance (61.7\% accuracy, close to the human noise ceiling of 66.67\%). We then align a variety of student models---trained with objectives ranging from image captioning to classification or self-supervised learning---using this teacher's representations; all models show substantially improved human alignment on the THINGS tests (relative performance increases from 21.28 - 74.47\%).

\noindent\textbf{Other cognitive tasks.}
{\color{black}{Our findings generalize across various object similarity tasks that are commonly used in the cognitive sciences---triplet odd-one-out task (relative performance increases up to $73.35\%$), Likert scale similarity ratings (up to $6.3$ fold increase in Spearman rank correlation coefficient), and multiple-arrangement tasks (up to $14.47$ fold increase in Spearman rank correlation coefficient). All performance increases are statistically significant at $\alpha=0.05$; for details see the SI.}} 

\subsubsection{Alignment at multiple levels of abstraction}
\label{sec:alignment_levels_abstraction}

\noindent\textbf{\emph{Levels} dataset.}
Because prior cognitive datasets were not specifically targeted for assessing knowledge of vision foundation models across levels of abstraction, we collected a novel dataset of human judgments---which we call \emph{Levels}---that is based on the triplet odd-one-out task, but stratified across different levels of the semantic hierarchy.} 
Specifically, we collect {\color{coarse-green}{\emph{global coarse-grained semantic}}}, which require deciding on the odd-one-out among broadly different categories; {\color{fine-blue}{\emph{local fine-grained semantic}}}, involving discerning subtle distinctions within the same category; and {\color{boundary-red}{\emph{class-boundary}}}, testing the capacity to identify category boundaries. For details see Sec.~\ref{sec:methods_levels}. %

\finaledit{
\noindent\textbf{Alignment at multiple levels.}}
The levels dataset allows us to systematically study discrepancies between human and model decisions across \finaledit{these different levels. We find that our soft-alignment method reduces these discrepancies at all levels, but especially for the global coarse-grained judgments, as we hypothesized. Specifically:
}

\noindent{\bf {\color{coarse-green}{Global coarse-grained}}}.
\finaledit{This level shows the largest improvements.} The base models achieved low accuracies of $36.09\%$ (ViT-B) $-$ $57.38\%$ (DINOv2 ViT-B). \finaledit{AligNet models significantly improved; all models performed well, with accuracies of $65.70\%$ (DINOv1 ViT-B) $-$ $68.56\%$ (DINOv2 ViT-B)--- \emph{above}} the human-to-human reliability score of $61.92\%$ (see \cref{fig:fig2}d, leftmost column). That is, the AligNet models' responses were more similar to average human responses (since each triplet response is the majority response of the subjects) than the level of agreement among the human participants. AligNet models' relative improvements ranged from  $19.48\%$ (DINOv2 ViT-B) $-$ $93.51\%$ (ViT-L).

\noindent{\bf {\color{fine-blue}{Local fine-grained}}}. \finaledit{Most base models did not strongly align to human responses for fine-grained semantics either; 
all models achieved poor alignment scores of $46.04\%$ (ViT-B) $-$ $57.72\%$ (DINOv2 ViT-B), except for DINOv1 ViT-B which performed significantly better($62.92\%$; near the human noise ceiling of $65.92\%$; see SI Tab.~\ref{tab:levels_results}). AligNet models achieved increased accuracies of $58.93$ (ViT-S) $-$ $62.92\%$ (DINOv1 ViT-B); relative improvements $7.84\%$ (DINOv2 ViT-B) - $46.03\%$ (ViT-L) (see \cref{fig:fig2}d, center column).

\noindent{\bf {\color{boundary-red}{Class-boundary}}}. Supervised classifiers and image/text contrastive \finaledit{base} models performed close to the noise ceiling; \finaledit{accuracies ranged from} $81.96\%$ (SigLIP ViT-B) to $93.67\%$ (ViT-L). Others performed worse; the CapPa captioning model achieved $70.37\%$. AligNet fine-tuning brought all models to a similar level, \finaledit{achieving accuracies up to $93.24\%$ (ViT-L), higher than} the human noise ceiling of $89.21\%$ (see \cref{fig:fig2}d, rightmost column). 
Relative improvements were $0.62\%$ (ViT-L) - $32.29\%$ (CapPa ViT-B). 
For more \finaledit{performance details see SI} Tab.~\ref{tab:levels_results}.

\noindent{\bf AligNet model uncertainties correspond to human latencies}.\label{sec:alignment_uncertainties} \finaledit{We also collected (continuous) human response times (RTs), which we use as} a proxy of the participants' uncertainty \citep{ratcliff1978theory, kiani2014choice}. We measured \finaledit{model uncertainty as the} 
entropy of the three pairwise similarities within each triplet. Base model uncertainties were not correlated with human RTs for the coarse ($\rho=-0.014-0.184$) and fine-grained settings ($\rho=0.047-0.160$) and moderately correlated for the class boundary setting ($\rho=0.208-0.432)$.
All AligNet models showed substantially increased uncertainty alignment across all levels (see \cref{fig:fig2}e), especially the coarse-grained abstraction level ($\rho=0.479-0.506)$.

{\color{black}{
\finaledit{\noindent{\bf Evaluating other model classes}}. To confirm that the aforementioned representational weaknesses are present in other classes of deep learning models, we evaluated \finaledit{two} state-of-the-art natively-multimodal LLM, Gemini 2.0 Flash \finaledit{and Gemini 2.5 pro~\citep{comanici2025gemini}}, on Levels. \finaledit{These VLMs perform similarly to---or slightly better than---}the better pretrained vision models across all levels; however, \finaledit{they} still substantially underperform our AligNet fine-tuned models (see SI~\ref{appx:vlm_eval_levels}). Additionally, we evaluated models trained on Ecoset~\cite{ecoset}, an ecologically motivated natural image dataset, and find that Ecoset models have severe difficulties matching the human similarity judgments 
(see SI~\ref{appx:ecoset_on_levels}). 
These results confirm that the AligNet fine-tuning offers greater improvements in human alignment than merely incorporating language modeling or ecological training.}
}

}

\subsection{Aligned models reflect the conceptual hierarchy}
\label{sec:conceptual_hierarchy}

\begin{figure}[ht!]
    \centering
    \includegraphics[width=\textwidth]{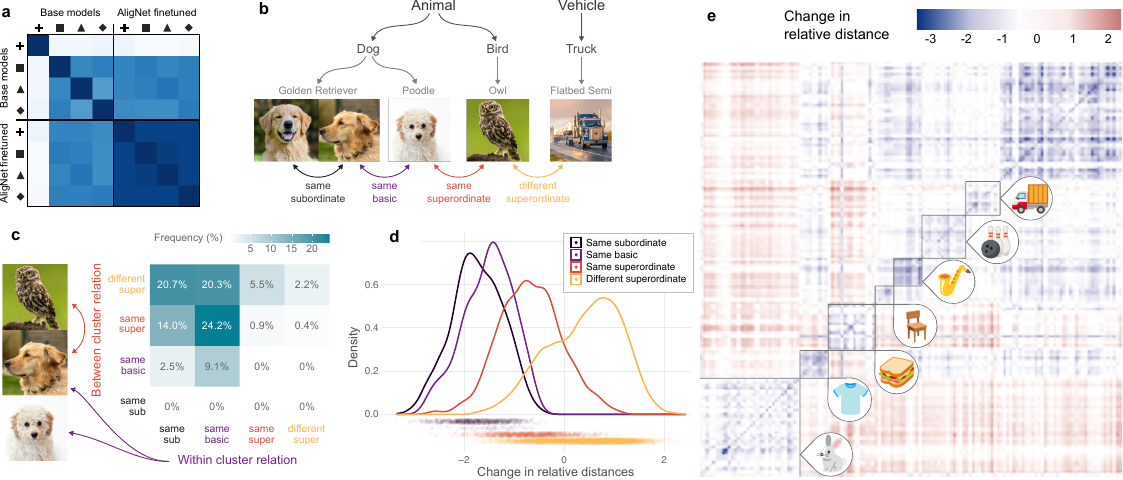}
    \caption[Aligned models reflect the semantic hierarchy.]{\doublespacing Aligned models reflect the semantic hierarchy. \textbf{a}: Before alignment, models trained with different losses have dissimilar representation structures---particularly those trained for supervised classification. After alignment, however, model representation structures are much more similar to each other. \textbf{b}: To understand the alignment, we study how the models' representations change across the semantic hierarchy, from relations between images within the same subordinate category to relations across superordinate categories. \textbf{c}: The cluster-driven triplet sampling tends to produce triplets where two images have a closer relation than the third. \textbf{d}: The result is that the relations between image representations change following the semantic hierarchy---images from the same {\color{black}subordinate,} basic, or superordinate category tend to move closer together, while those from different superordinate categories move farther apart. \textbf{e}: Visualizing the distance changes in more detail, with some superordinate categories boxed on the diagonal and labelled with icons. (Panels d-e are for the representations of ViT-B.)}
    \label{fig:representation_structure_changes}
\end{figure}

\finaledit{
How do model representations change after soft-alignment? In \cref{fig:representation_structure_changes} we show that while the model representations are dissimilar before alignment, they become more aligned with each other after soft-alignment. This convergence is driven by models aligning better with the human conceptual hierarchy \citep[\cref{fig:representation_structure_changes}b; cf.][]{rosch1976basic}. Our soft-alignment procedure embeds this global structure at two levels: first in the cluster-based sampling (\cref{fig:representation_structure_changes}c), and then in the soft-labels (see above). Because of these factors, the relationships between image representations change during alignment according to their semantic relationship; representations of images from the same basic category tend to move closer together, those of images from the same superordinate category tend to move somewhat closer, and those from different superordinate categories tend to move apart (all effects are highly-significant; \(t\)s\(>3.93\), \(p\)s\(<0.001\)).

As an illustrative example, in a base ViT-B the representations of lizards are close to those of some plants and fruits due their similarity in texture, color, or background; after alignment, they become similar to representations of other animals and more distant from those of other, unrelated superordinate categories. This reorganization yields better generalization, e.g. when a lizard image is used as an example depicting an abstract category like animals.

SI~\ref{appx:eval_hierarchy} presents detailed analyses, including reorganization at high levels such as living vs. nonliving, across layers, and in other models. Furthermore, SI~\ref{appx:alignet_unalignet_disagreement_humans} shows that where the AligNet model and a baseline unaligned model disagree, human judgments are strongly correlated with those of AligNet, but not the ablation model---in fact, emph{every} human participant in the study agreed more with AligNet.
}

\subsection{Alignment improves generalization and out-of-distribution robustness}
\label{sec:ml_results}

\begin{figure}[ht!]
    \centering
    \includegraphics[width=1.0\textwidth]{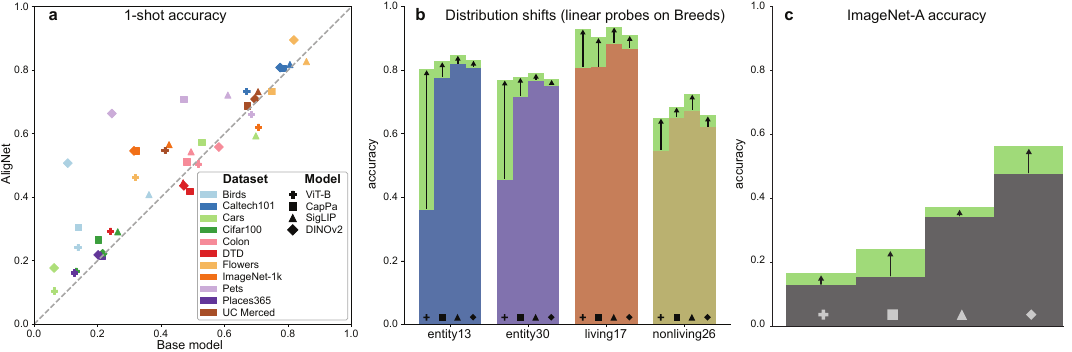}
    \caption[Downstream machine learning results]{\doublespacing Machine learning downstream results. We evaluated how AligNet finetuning affected the downstream task performances of various pre-trained VFMs using liner probing. \textbf{a}: 1-shot accuracy before ($x$-axis) vs. after ($y$-axis) AligNet finetuning on various image datasets. \textbf{b}: Accuracy improvements on the four BREEDS distribution shift benchmark datasets. \textbf{c}: Accuracy improvements on the ImageNet-A dataset that evaluates model robustness.}
    \label{fig:ml-results}
\end{figure}

\finaledit{How do human-aligned representations affect performance on machine learning tasks? We investigated how alignment improves generalization and out-of-distribution robustness across a variety of downstream tasks.

\noindent{\bf One-shot classification}. We first test an extremely-challenging generalization setting: classifying images given only a single labelled example per class. In \cref{fig:ml-results} we show one-shot performance before and after soft-alignment on ten image-classification datasets from varied domains, such as fine-grained bird~\cite[Birds;][]{welinder2010caltech-ucsd} and flower~\cite[Flowers;][]{nilsback2008automated} types classification, multi-domain natural image classification~\cite[ImageNet;][]{deng2009imagenet}, and scene recognition~\cite[Places365;][]{zhou2018places}. The majority of cases (32 of 40) show an improvement, sometimes by a substantial margin (e.g. DINOv2 shows a \(2.7\times\) increase on the Pets dataset); overall, alignment significantly increases the generalization performance on these tasks (p < 0.05). 
These results show how human-aligned representations support strong generalization from little data.
See SI~\ref{appx:few_shot} for additional results, including the complementary benefits of combining our method with other approaches to few-shot generalization.

\noindent{\bf Distribution shift}. A long-standing problem for applying machine learning algorithms is distribution shift \citep[cf.,][]{sugiyama2012machine, farahani2021brief}: in deployment, data often differ in subtle ways from the training data, leading to unexpected model failures.
To evaluate whether the global structure induced by alignment helps ameliorate this issue, we evaluated our models on the BREEDS benchmarks~\citep{santurkar2020breeds}---which specifically tests generalization under input distribution shifts, using datasets where training and test data points are sampled from different subpopulations.
\cref{fig:ml-results}b shows that AligNet fine-tuning consistently improves performance significantly across all benchmarks and models types (especially the image classifier, ViT-B). See SI~\ref{appx:breeds_detailed_results} for further results.

\paragraph{Model robustness.} Alignment also improves robustness. We evaluate on ImageNet-A \citep{hendrycks2021natural}, a challenging dataset of natural images that are adversarial (i.e., models tend to misclassify them, but humans perform better).
Again, alignment improves accuracy for all models (cf. \cref{fig:ml-results}c), with improvements of up to $9.5$ percentage points ($1.6\times$ improvement for CapPa). While our method is not designed for improving OOD robustness, its improvements are comparable to state-of-the-art methods designed for precisely this problem (see Tab.~\ref{tab:ood_robustness_comparison} in the SI).

Togerther, these machine learning results corroborate that AligNet fine-tuning improves the generalization, transfer, and robustness of model representations.
}

\section{Discussion}
\label{sec:discussion}

The differences between natural intelligence and the capabilities of neural networks are the subject of long-standing debates \citep{fodor1988connectionism, lake2017building}. Despite the dramatic \finaledit{recent} progress in AI,
these discussions persist, because 
deep learning systems still seem to fail in non-human-like ways 
\citep{lapuschkin2019unmasking, geirhos2020shortcut}.

\finaledit{Here}, we have highlighted---and addressed---a key deficiency in a broad class of vision foundation models: their representations do not adequately represent the multi-level conceptual structure of \finaledit{human} semantic knowledge (see Sec.~\ref{sec:human_alignment}).
\finaledit{We demonstrate this deficiency through \emph{Levels}, a new dataset of human similarity judgments across multiple levels of abstraction.} 
To address this deficiency, we established a methodological framework for \finaledit{aligning deep learning models' representations with human judgments to create more human-aligned systems.} 
%
\finaledit{This framework involves bootstrapping from a small quantity of human data to train a surrogate teacher model, and using this teacher to create a large new synthetic dataset (\emph{AligNet}), which we use to fine-tune various vision-foundation models to inject human-aligned structure}

\finaledit{This approach yields} significantly \emph{increased alignment} to human judgments on cognitive science tasks (Sec.~\ref{sec:human_alignment}), \finaledit{and \emph{better} generalization and robustness} on representative machine learning tasks (Sec.~\ref{sec:ml_results}). \finaledit{Thus}, soft-alignment helps to reduce the brittleness of machine learning models under changing environments.
Moreover, our results illustrate how the broader paradigm of studying representational alignment \citep{kriegeskorte2008representational,sucholutsky2023getting} can not only yield \emph{insights} \finaledit{about \emph{how systems relate}}, but can be leveraged to actively align model representations with human \finaledit{knowledge} 
to \emph{improve the models' generalization}.


These results contribute to long-standing debates over \finaledit{which features of human intelligence neural networks may lack} 
\citep{fodor1988connectionism,tenenbaum2011grow,lake2017building}.
In particular, 
\finaledit{one line of critique argues that neural networks lack the capability to appropriately represent abstract relations like same and different \citep{marcus1998rethinking,holyoak2014proper}, or to organize knowledge into hierarchies of concepts \citep{tenenbaum2011grow}}. While aspects of these critiques have been refuted in simple synthetic settings \citep[e.g.][]{geiger2023relational}, similar criticisms persist for modern foundation models
\citep{bowers2022deep}. Our results show that, while standard {\color{black} training} objectives do not adequately capture \finaledit{hierarchical category relations}, these relations can be distilled into the models---
which improves the models' 
resilience under the \finaledit{distribution shifts highlighted in} prior critiques. {\color{black} These results \finaledit{show} that hierarchical representations may emerge from a system that is neither explicitly hierarchically structured, nor trained explicitly on the hierarchy.}

Although we focused on vision, similar global misalignments likely arise in other areas of research. For instance, in natural language processing, models are similarly trained with objectives that focus on distinguishing between close matches (e.g., prediction objectives that primarily distinguish words \finaledit{that are likely to occur}, rather than considering their relations to less probable concepts).
Applying alignment techniques may therefore analogously help to better capture the global structure of semantic and syntactic relationships among language inputs that these objectives might miss.

%


\finaledit{More broadly,} AI systems have been successfully adopted in many areas. 
However, these deployments lead to \emph{practical} \citep{paleyes2023challenges} and \emph{conceptual} \citep{amodei2016concrete} concerns about trustworthiness 
\finaledit{and safety}.
It is therefore increasingly important to identify the reasons why these systems occasionally fail 
and how to alleviate these failures.
Our work advances the understanding of the deficiencies of vision model representations, and simultaneously shows a viable path for ameliorating these deficiencies by alignment with human judgments.

\finaledit{Our work has several} limitations that could be addressed in future efforts. First, the models we used neither account for context in similarity judgments nor higher-order relations. 
Second, 
human representations may vary systematically across individuals, cultures, and so on. 
Finally, human judgment is full of flaws, intrinsic contradictions and discrepancies. Given these issues, perfect alignment to human performance may not always be desirable for a technical system. Thus, future work could explore how to best learn from human knowledge without adopting human imperfections.

In summary, we have provided an initial \finaledit{approach to distill} global, human-\finaledit{aligned} similarity structures into the representations of modern deep neural networks. We demonstrated an efficient path toward a \emph{best-of-both-worlds representation} that is both more consistent with human \finaledit{judgments} and more practically useful, paving the way toward more robust, interpretable, and human-\finaledit{aligned} artificial intelligence systems. We hope that our work will inspire more general approaches to (softly) aligning models by distilling \finaledit{human} priors into their representations.

\section{Methods summary}
\label{sec:methods}

\subsection{AligNet}
\label{sec:methods_alignet}

\noindent{\bf{THINGS dataset objects}}. 
{\color{black}{We use the THINGS dataset \citep{hebart2020revealing} --- a behavioral dataset of human responses in a triplet odd-one-out task for object images --- to build a \emph{surrogate model} (denoted as  \emph{teacher}) that approximates a human object similarity space. This similarity space is determined by the human odd-one-out responses from which various human concepts can be inferred \citep{hebart2020revealing, muttenthaler2022vice}.}} The THINGS dataset can formaly be defined as $\mathcal{D} \coloneqq (\{a_{s}, b_{s}\}\mid\{i_{s},j_{s},k_{s}\})_{s=1}^{n}$ which denotes a dataset of $n$ object triplets and corresponding human odd-one-out responses, where $\{a, b\} \subset \{i,j,k\}$ and $\{a, b\}$ is the object pair that was chosen by a human participant to have the highest similarity. Let $\mX \in \mathbb{R}^{m\times p}$ be the teacher model representations for the $m=1854$ objects in the THINGS dataset. Note that each category in the THINGS dataset is represented by one object image. From $\mX$ we can construct a similarity matrix for all object pairs $\mS \coloneqq \mX\mX^{\top} \in \mathbb{R}^{m\times m}$, where $S_{ij} = \vx_{i}^{\top}\vx_{j}$ is the representational similarity for objects $i,j$.

\noindent{\bf Linear transformation}. Our goal is to learn an affine transformation into the THINGS human object similarity space of the form: $\vx^{\prime} =\mW\vx+\vb$. Here, $\mW \in \mathbb{R}^{p \times p}$ is a learned transformation matrix, $\vb \in \mathbb{R}^{p}$ is a bias, and $\vx \in \mathbb{R}^{p}$ is the neural network representation for a single object image in the THINGS dataset. We learn the affine transformation for the representation of the image encoder space of the SigLIP-So400m teacher model (see the Methods for details about the teacher model). Using this affine transformation, an entry in the pairwise similarity matrix $\mS^{\prime}$---which represents the similarity between two object images $i,j$---can now be written as $S^{\prime}_{i,j} \coloneqq \left(\mW\vx_{i} + \vb\right)^{\top} \left(\mW\vx_{j} + \vb\right)$.

\noindent{\bf Uncertainty distillation}. \label{sec:method_ud} We mainly follow the optimization process introduced in \citet{muttenthaler2023improving}. However, we modify their approach by injecting uncertainty measures about human odd-one-out responses into the representation space of the teacher, using a recent approximate Bayesian inference method for learning object concepts from human behavior \citep{muttenthaler2022vice}. Thus, we replace the negative log-likelihood of the discrete human odd-one-out choices---which we refer to as \emph{hard-alignment}---with the negative log-likelihood of the probabilities for the pairwise triplet similarities obtained from the Bayesian inference model---referred to as \emph{soft-alignment}. Hard-alignment is defined as,
\begin{align}
    \mathcal{L}_{\mathrm{hard-align}}(\mS^{\prime}) \coloneqq -\frac{1}{n}\sum_{s=1}^n\log \underbrace{q\left(\{a_{s}, b_{s}\}|\{i_{s}, j_{s}, k_{s}\},\mS^{\prime}\right)}_{\text{\color{black}{odd-one-out prediction}}},
\label{eq:discrete_align_loss}
\end{align}
where the affine transformation of the teacher's representation space is optimized to match the (discrete) human responses. In contrast, soft-alignment gives the following KL divergence between the human uncertainties $p^{*}$ and the teacher model probabilities $q$ obtained from applying a $\softmax$ function to the pairwise similarities of $\mS$,
\begin{align}
    \mathcal{L}_{\mathrm{soft-align}}\left(\mS^{\prime}\right) \coloneqq \frac{1}{n}\sum_{s=1}^n\underbrace{{\color{black}{p^{*}_{s}\left( \{i_{s},j_{s},k_{s}\} \right)}} \log p^{*}_{s}\left( \{i_{s},j_{s},k_{s}\} \right) }_{\text{{\color{black}{human uncertainty}}}}~-~\underbrace{p^{*}_{s}\left( \{i_{s},j_{s},k_{s}\} \right)  \log q_{s} \left(\{i_{s},j_{s},k_{s}\}, \mS^{\prime} \right)}_{\text{{\color{black}{cross-entropy}}}},
\label{eq:soft_align_loss}
\end{align}
The final objective for learning the \emph{uncertainty distillation} (UD) transformation is defined as
\begin{equation}
    \argmin_{\mW,\vb}~ \underbrace{\mathcal{L}_{\mathrm{soft-align}}\left(\mX, \mW,\vb\right)}_{\text{\color{black}{soft alignment}}} + \lambda \norm{\textstyle \mW-\left(\sum_{j=1}^{p}{\mW_{jj}}/p \right) I}_{\mathrm{F}}^{2},
\label{eq:ud_objective}
\end{equation}
where $I \in \mathbb{R}^{p \times p}$ is the identity matrix. The right-hand side of the above objective is an $\ell_{2}$-regularization whose aim is to preserve the nearest neighbor information (or equivalently, the local similarity structure) of the pretrained representations while learning an affine transformation into the THINGS human object similarity space. The above equation is minimized using standard SGD.

\noindent{\bf Superordinate clusters}. \label{sec:method_clustering}
{\color{black}{After we find a transformation that best matches the human responses}}, we embed the ImageNet train set in the transformed representation space of the teacher and cluster the representations into superordinate categories using $k$-Means clustering. We find the optimal number of clusters $k$ using the Elbow criterion. We use the clusters for generating triplets of distinct ImageNet images by always sampling two images from the same cluster and one image from a different cluster. For all triplets that we generate we identify their odd-one-out choice using the representations of the surrogate teacher model.

\noindent{\bf Human-\finaledit{aligned} responses}. The responses of the surrogate model simulate a dataset of human-\finaledit{aligned} triplet odd-one-out responses. In addition to the discrete odd-one-out choices, the dataset includes the exact relationships among all pairwise similarities in a triplet obtained from the probability space of the teacher model. Thus, we now have access to soft choices.

\noindent{\bf AligNet objective}. To distill the pairwise similarity structure of the teacher into a different student network, we introduce a novel Kullback-Leibler divergence based objective function similar to Eq.~\ref{eq:soft_align_loss} that facilitates the distillation process. This loss function is defined as
\begin{align*}
    \mathcal{L}_{\mathrm{alignet}}(\mS^{\prime}, \mS^{\dagger}) \coloneqq \frac{1}{B}\sum_{s=1}^{B}{\color{black}{\sigma\left(\left[S^{\prime}_{i,j},S^{\prime}_{i,k},S^{\prime}_{j,k}\right], \tau^{\prime}\right)_{s}}}\log \sigma \left(\left[S^{\prime}_{i,j},S^{\prime}_{i,k},S^{\prime}_{j,k}\right], \tau^{\prime}\right)_{s} - \\ \sigma\left(\left[S^{\prime}_{i,j},S^{\prime}_{i,k},S^{\prime}_{j,k}\right], \tau^{\prime}\right)_{s}  \log \sigma\left(\left[S^{\dagger}_{i,j},S^{\dagger}_{i,k},S^{\dagger}_{j,k}\right], \tau^{\dagger}\right)_{s},
\end{align*}
where $\mS^{\prime}$ and $\mS^{\dagger}$ are the pairwise similarity matrices of the teacher and the student representation spaces respectively, $B$ is the number of image triplets in a batch, $\sigma$ is a $\softmax$ function that transforms the similarities into probabilities, and $\tau^{\prime}$ and $\tau^{\dagger}$ are temperature values for the teacher and student that we find via grid search. The final AligNet objective is defined as
\begin{equation*}
    \argmin_{{\theta}^{\dagger}}~\mathcal{L}_{\mathrm{alignet}}(f_{{\theta}^{\dagger}}) + \lambda \norm{\textstyle \theta^{*}-\theta^{\dagger}}_{2}^{2},
\end{equation*}
where $\theta^{*}$ are the parameters of the pretrained student before fine-tuning and $\theta^{\dagger}$ are the parameters of the fine-tuned student. The right-hand side is similar to the $\ell_{2}$-regularization employed for learning the UD transformation. It tries to preserve the (fine-grained) structure of the pretrained representation space as a function of $\lambda$ which determines the strength of the regularization. 

\subsection{Representational Similarity Analysis}
\label{sec:methods_rsa}


Representational Similarity Analysis (RSA) is a well-established method for comparing neural network representations---extracted from an arbitrary layer of the model---to representations obtained from human behavior \citep{kriegeskorte2008representational}. In RSA, one first obtains representational similarity matrices (RSMs) for the human behavioral judgments and for the neural network representations (more specific details can be found in the SI). These RSMs measure the similarity between pairs of examples according to each source. As in previous work \citep{cichy2019, king2019, hebart2020revealing, muttenthaler2023human, muttenthaler2023improving}, we flatten the upper triangular of human and model RSMs respectively and quantify their similarities using use the Spearman rank correlation coefficient. In contrast to the Pearson correlation coefficient, the Spearman rank correlation is scale-invariant and thus better suited to measure similarities of judgments obtained from different sources.

\subsection{Levels}
\label{sec:methods_levels}

We collected a new multi-level similarity judgment dataset from \(N = 473\) human participants, which we named \emph{Levels}. The dataset contains odd-one-out judgments on three different types of triplets: {\color{coarse-green}{\emph{coarse-grained semantic}}}, which require deciding on the odd-one-out in broadly different categories; {\color{fine-blue}{\emph{fine-grained semantic}}}, which involved discerning subtle within category distinctions; and {\color{boundary-red}{\emph{class-boundary}}}, which tested for category boundary detection. Consistent selection of the same odd-one-out image (e.g., \(i\)) in multiple participants indicated that the remaining two images (e.g., \(j\) and \(k\)) were closer to each other in the participants’ concept space than either was to the odd-one-out (see the SI for details about the data collection). 
\emph{Levels} allowed us to evaluate model-human alignment for the same set of stimuli on various levels of abstraction, and to assess how well the models capture the inherent uncertainty in human judgments, inferred from response latencies. 

\noindent{\bf Human-to-human alignment}. We computed the human noise ceiling for each abstraction setting in Levels using a Leave-One-Out (LOO) cross-validation approach. In LOO, the agreement level for a triplet is computed as the average match rate between a held-out participant's response and the majority response of the remaining population. Thus, for a triplet that was used in five participants, on each LOO iteration, one participant response is held out and the remaining four comprise the population. The human-to-human reliability score is then calculated as the average agreement level across all triplets in the dataset.

\subsection{Alignment with conceptual hierarchy}
\label{sec:methods_conceptual_hierarchy}

When analyzing alignment with the conceptual hierarchy, we use the original ImageNet category labels for the images \citep{deng2009imagenet}. ImageNet is structured according to the WordNet hierarchy, from which we extract basic- and superordinate-categories that align with the prior cognitive work. We quantify these changes with mixed-effects linear regressions, see SI~\ref{appx:eval_hierarchy}.

\finaledit{
\section*{Code Availability}

The training code is available at \url{https://github.com/google-deepmind/alignet}. The aligned model checkpoints are  publicly available at \url{https://console.cloud.google.com/storage/browser/alignet}.
Both the experiment and analysis code for Levels and for the human validation of the representational differences between AligNet and UnaligNet are hosted on GitHub and archived on the Zenodo platform (see \url{https://zenodo.org/records/15554034} and \url{https://zenodo.org/records/15554174}).

\section*{Data Availability}
The synthetically created AligNet data are publicly available at~\url{https://console.cloud.google.com/storage/browser/alignet}.
The Levels data are available on GIN prior to publication: \url{https://doi.org/10.12751/g-node.hg4tdz}.
}

\section*{Acknowledgements}
L.M., F.B., and K.R.M. were in part supported by the German Ministry for Education and Research (BMBF) under Grants 01IS14013A-E, 01GQ1115, 01GQ0850, 01IS18025A, 031L0207D, and 01IS18037A. K.R.M.\ was partly supported by the Institute of Information \& Communications Technology Planning \& Evaluation (IITP) grants funded by the Korean government (MSIT) (No. 2019-0-00079, Artificial Intelligence Graduate School Program, Korea University and No. 2022-0-00984, Development of Artificial Intelligence Technology for Personalized Plug-and-Play Explanation and Verification of Explanation). B.S. and F.B. were supported by European Research Council Consolidator Grant ERC-2020-COG-101000972 and Deutsche Forschungsgemeinschaft (DFG) Grant 462752742. Correspondence should be addressed to L.M., K.R.M., and  A.L. The authors would like to thank Aravindh Mahendran, Mehdi S. M. Sajjadi, Robert Geirhos, and Xiaohua Zhai for fruitful discussions and Ishita Dasgupta, Katherine Hermann, Shakir Mohamed \finaledit{and the editors and reviewers} for helpful comments on earlier versions of the manuscript.

\section*{Author contributions} 

L.M.: \textit{Initiated and led the project. Played a major role in every part of the project.} Writing—original draft, conceptualization, investigation, writing—review and editing, methodology, data curation, validation, formal analysis, software, project administration and visualization. 
K.G.: \textit{Helped leading the project and played a critical role in major parts (e.g., running experiments, creating figures) of the project.} Writing—original draft, conceptualization, investigation, validation, formal analysis, software, project administration, and visualization.
F.B.: \textit{Led the Levels data collection and was in full charge of everything concerned with it.} Writing—original draft, investigation, validation, writing—review and editing,  data curation, software, and visualization.
B.S.: Writing—review and editing, validation, and supervision. 
S.K.: Writing—review and editing, methodology, and supervision.
M.C.M.: Writing—review and editing, validation, resources, methodology, and supervision.
K.R.M.: \textit{Helped leading the project and shaping some of its ideas.}  Writing—original draft, conceptualization, writing—review and editing, methodology, funding acquisition, resources, supervision, and project administration.
T.U.: \textit{Hosted L.M. during his first internship at GDM and helped leading the project.} Conceptualization, investigation, writing—review and editing, methodology, validation, supervision, software, project administration, and visualization.
A.L.: \textit{Played a major role in administering and supervising the project and shaping some of its ideas.} Writing—original draft, conceptualization, investigation, writing—review and editing, methodology, validation, supervision, project administration and visualization.


\bibliography{references}


\clearpage
\appendix
\section*{Supplementary Material}
\section{Methods}
\subsection{Soft-alignment}
\label{appx:method}

This section is organized as follows: We start by describing how we transform model representations into a space that matches human similarity judgments about coarse-grained semantic object relations. We introduce a novel affine transformation that matches human similarity judgments and injects the uncertainties that humans assign to their triplet odd-one-out choices into a model's representation space. Using these human-aligned representations, we sample triplets of ImageNet \citep{deng2009imagenet} images differently than uniform random sampling by clustering the representations into superordinate categories and using those clusters for data partitioning. Finally, after having created AligNet triplets, we can fine-tune models with a novel triplet loss object function.

\subsubsection{Representational alignment}
\label{appx:methods:rsa}

\noindent {\bf Data.} \label{appx:method-data}
To increase the degree of alignment between human and neural network similarity spaces, we use the publicly available THINGS dataset, which is a large behavioral dataset of $4.7$ million unique triplet responses from $12{,}340$ human participants for $m=1854$ natural object images \citep{hebart2023things-data} from the public THINGS object concept and image database \citep{hebart2019things}. Let $\mathcal{D}_{\mathrm{things}} \coloneqq (\{a_{s}, b_{s}\}\mid\{i_{s},j_{s},k_{s}\})_{s=1}^{n}$ be the dataset of object triplets and corresponding human responses, where $\{a, b\} \subset \{i,j,k\}$ and $\{a, b\}$ is the object pair that was chosen by a human participant to have the highest similarity.

\noindent {\bf Odd-one-out accuracy.} \label{appx:method-ooo}
The triplet odd-one-out task is a commonly used task in the cognitive sciences to measure human notions of object similarity \citep{fukuzawa1988, robilotto2004limits, hebart2020revealing, muttenthaler2022vice}. To measure the degree of alignment between human and neural network similarity judgments in the THINGS triplet task, we embed the $m=1854$ THINGS images into the representation space of a neural network with $\mX \in \mathbb{R}^{m \times p}$. Given vector representations $\vx_1$, $\vx_2$, and $\vx_3$ of the three images in a triplet, we first construct a similarity matrix $\mS \in \mathbb{R}^{3\times 3}$ where $S_{i,j} \coloneqq \vx_i^\top \vx_j$ is the dot product between a pair of image representations. We identify the closest pair of images in the triplet as $\argmax_{i, j > i} S_{i,j}$ with the remaining image being the odd-one-out. We define odd-one-out accuracy as the fraction of triplets where the odd-one-out ``chosen by a model'' is identical to the human odd-one-out choice. Thus, we want to find an affine transformation of the general form ${\vx}^{\prime} = \mW\vx + \vb$ with ${\vx}^{\prime} \in \mathbb{R}^{p}$ that maximizes this accuracy. The pairwise similarity can then be defined as $S^{\prime}_{ij} = \left(\mW\vx_{i} + \vb\right)^{\top}\left(\mW\vx_{j} + \vb\right)$.

\noindent {\bf Hard alignment loss.} \label{appx:method-discrete-loss}
Given a similarity matrix of neural network representations $\mS$ and a triplet $\{i,j,k\}$, the likelihood of a particular pair, $\{a, b\}\subset \left\{i,j,k\right\}$, being most similar with the remaining object being the odd-one-out, is modeled by the $\softmax$ of the object similarities,
\begin{align}
\sigma(\mS, \tau) \coloneqq \exp\left(S_{a,b}/\tau\right) / \left(\exp\left(S_{i,j}/\tau\right)+\exp\left(S_{i,k}/\tau\right)+\exp\left(S_{j,k} / \tau \right)\right).
\label{eq:triplet-likelihood}
\end{align}
We can then define the probability of the neural network model to choose the most similar pair (according to the human subjects) to be $q\left( \{a, b\}|\{i, j, k\},\mS \right) \coloneqq \sigma(\mS, \tau)$ with $\tau=1$.
For $n$ triplet responses, the discrete negative log-likelihood is defined as follows, 
\begin{align*}
    \mathcal{L}_{\mathrm{hard-align}}(\mS^{\prime}) \coloneqq -\frac{1}{n}\sum_{s=1}^n\log \underbrace{q\left(\{a_{s}, b_{s}\}|\{i_{s}, j_{s}, k_{s}\},\mS^{\prime}\right)}_{\text{\color{black}{odd-one-out prediction}}}.
\end{align*}

\noindent{\bf Modelling human uncertainties.} \label{appx:method-uncertainty}
Since each triplet response is a discrete choice, we don't have direct access to the uncertainties of a human participant over the objects in a triplet. Thus, the above loss function optimizes a transform to match the human choice but does not take into account the uncertainties over the three odd-one-out alternatives. However, it is possible to model these uncertainties using Variational Interpretable Concept Embeddings \citep[VICE;][]{muttenthaler2022vice}, a recently proposed, approximate Bayesian inference method for learning an interpretable object concept space from human similarity judgments. VICE has shown remarkable performance in predicting the (dis-)agreement in human similarity judgments for multiple similarity judgment datasets, including THINGS \citep{muttenthaler2022vice}. 

We train a VICE model on the official THINGS train triplet dataset using the (default) hyperparameters recommended by the authors. To capture the uncertainties in human triplet responses, VICE learns a mean, $\mu \in \mathbb{R}^{m \times d}$, and a variance, $\sigma \in \mathbb{R}^{m \times d}$, for each object image and each object dimension respectively. Therefore, the set of VICE parameters is defined as $\theta = \{\mu, \sigma\}$. VICE uses the reparameterization trick \citep{kingma2013auto-encoding, rezende2014stochastic} to generate an embedding matrix $\mY \in \mathbb{R}^{m \times d}$, $\mY_{\theta, \epsilon} = \mu + \sigma \odot \epsilon$, where
$\epsilon \in \mathbb{R}^{m\times d}$ is entrywise $\mathcal{N}(0,1)$, and $\odot$ denotes the Hadamard (element-wise) product. 

After convergence, we can use a VICE model to obtain a posterior probability distribution for each triplet in the data. We approximate the probability distribution using a Monte Carlo estimate \citep{graves2011practical, blundell2015weight, blei2017variational} from $R$ samples $\mY^{(r)} = \mY_{\hat{\theta},\epsilon^{(r)}}$ for $r=1,\hdots,R$, yielding
\begin{equation*}
    \hat{p}\left(\{y_{s}, z_{s}\}|\{i_{s}, j_{s}, k_{s}\}\right) \coloneqq \frac{1}{R} \sum_{r=1}^R \underbrace{p\left(\{y_{s}, z_{s}\}|\{i_{s}, j_{s}, k_{s}\},\mY^{(r)}\right)}_{\text{\color{black}{Monte-Carlo estimate(s)}}},
\label{eq:choice_distribution}
\end{equation*}
where we set $R=50$ because we found it to yield the best predictive performance on the official THINGS validation set. This gives a representative probability estimate for each of the three pairs in a triplet to be selected as the most similar pair.

\noindent {\bf Soft alignment loss.} \label{appx:method-soft-loss}
Using the posterior probability estimates obtained from VICE, we transform the original THINGS triplet dataset of discrete triplet choices into a triplet dataset of probability distributions that reflect the human uncertainties of the triplet alternatives. Let $\mathcal{D}^{\dagger} \coloneqq \left(p^{*}_{s}\left( \{i_{s},j_{s},k_{s}\} \right) \right)_{s=1}^{n}$ be the transformed triplet dataset, where
\begin{equation*}
p^{*}_{s}\left( \{i_{s},j_{s},k_{s}\} \right) \coloneqq \hat{p}\left(\{y_{s}, z_{s}\} \mid \{i_{s},j_{s},k_{s}\}\right)~\forall~\{y, z\} \subset \{i, j, k\}.
\end{equation*}
Now, for $n$ triplet responses we can define the negative log-likelihood for the soft alignment loss as,
\begin{align*}
    \mathcal{L}_{\mathrm{soft-align}}(\mS^{\prime}) \coloneqq \frac{1}{n}\sum_{s=1}^n \underbrace{{\color{black}{p^{*}_{s}\left( \{i_{s},j_{s},k_{s}\} \right)}} \log p^{*}_{s}\left( \{i_{s},j_{s},k_{s}\} \right) }_{\text{{\color{black}{human uncertainty}}}}~-~\underbrace{p^{*}_{s}\left( \{i_{s},j_{s},k_{s}\} \right)  \log q_{s} \left(\{i_{s},j_{s},k_{s}\}, \mS^{\prime} \right)}_{\text{{\color{black}{cross-entropy}}}},
\end{align*}
where $q_{s}\left( \{i_{s},j_{s},k_{s}\}, \mS \right) \coloneqq q\left(\{y_{s}, z_{s}\} \mid \{i_{s},j_{s},k_{s}\}, \mS \right)~\forall~\{y, z\} \subset \{i, j, k\}$.

\noindent {\bf Uncertainty distillation.} \label{appx:method-uncertainty-distill}
Using the above soft alignment loss function instead of the vanilla hard alignment loss proposed by \citet{muttenthaler2023human} for aligning the model representations with human global object similarity, the final objective function for the \emph{uncertainty distillation (UD)} transformation is defined as follows,
\begin{equation*}
    \argmin_{\mW,\vb}~ \underbrace{\mathcal{L}_{\mathrm{soft-align}}(\mX, \mW,\vb)}_{\text{\color{black}{soft alignment}}} + \lambda \norm{\textstyle \mW-\left(\sum_{j=1}^{p}{\mW_{jj}}/p \right) I}_{\mathrm{F}}^{2}.
\end{equation*}
Although this expression is similar to the global transform defined in \citet{muttenthaler2023improving}, we find it to yield equally strong downstream task performance as the gLocal transform proposed in \citet{muttenthaler2023improving} while predicting human uncertainties better than the global transform. It appears as if there is barely any trade-off between representational alignment and downstream task performance for using the UD, whereas \citet{muttenthaler2023improving} found that the global transform yields slightly better human alignment but worse downstream task performance compared to the gLocal transform. We use the UD transformation for generating human-like similarity judgments by transforming a model's representation space with UD.

\subsubsection{Data generation} \label{appx:data-gen}

In the following section, we describe the AligNet data generation process. We start by introducing the data that we use for constructing the triplets. We continue with a detailed description of the different sampling strategies that we consider in our analyses. Finally, we explain how we collect model responses using transformed representations and define the objective function for fine-tuning models on AligNet.

\noindent {\bf Image data.}  \label{appx:imagenet} For creating AligNet, we use the publicly available ImageNet database \citep{deng2009imagenet}. ImagNet is a natural image dataset with $\approx10^{6}$ training data points and $1000$ image categories \citep{russakovsky2014imagenet}. The categories are almost equally distributed in the data with small variations in the number of images between the different classes. Hence, ImageNet can be considered a highly balanced dataset. ImageNet has been the dominant image dataset for training large computer vision models until the advent of image/text multi-modal training a few years ago. Although to date larger image datasets exists, ImageNet is still is one of the largest open-source and most widely used image datasets in the field of Computer Vision.

\noindent {\bf Triplet sampling.} \label{appx:alignet-triplet-sampling}
For generating triplets of images, we employ three different sampling strategies: \emph{random}, \emph{class-border}, and \emph{cluster-boundary} sampling. Let $m^{\prime}$ be the number of images in the data where $m^{\prime}=1,281,167$ and $C$ be the number of classes with $C=1000$. Let $\mathcal{D}_{\mathrm{image}} \coloneqq \left(x_{i}, y_{i}\right)_{i=1}^{m^{\prime}}$ be the ImageNet dataset of $m^{\prime}$ image-label pairs.

\noindent {\bf Random.} Uniform random sampling is the vanilla sampling approach and was used to create the THINGS datasets (see Sec.~\ref{appx:method-data}). In random sampling, three images are chosen uniformly at random without replacement from all of the $m$ images in the data to create a triplet. Since there are $C=1000$ classes and each class has approximately $1000$ images, most of the triplets generated with this approach contain three images from three different classes. The number of triplets different from triplets with images from three distinct classes is negligible. Note that this is the same sampling approach that was used to generate the THINGS triplets \citep{hebart2020revealing}. A triplet generated via random sampling can be defined as the following triplet set $\mathcal{S} \coloneqq \{x_{i}, x_{j} ,x_{k}\}$ with the constraint $\left(y_{i} \neq y_{j} \neq y_{k}\right)$.

\noindent {\bf Class-boundary.} Another way to sample image triplets is to exploit the label information associated with each data point.
Instead of three random images from three distinct classes, we determine class-boundary triplets to contain two images from the same class and one image from a different class.
This is similar to the approach introduced in \citet{muttenthaler2024set} where each odd-$k$-out set of images contains a majority class and $k$ odd class singletons.
This sampling approach allows models to learn class boundaries similar to the standard supervised learning setting.
A triplet generated via class-boundary sampling can be defined as the following triplet set $\mathcal{S} \coloneqq \{x_{i}, x_{j}, x_{k}\}$ with the constraint $\left(y_{i} = y_{j} \neq y_{k}\right) \lor \left(y_{i} \neq y_{j} = y_{k}\right) \lor \left(y_{i} = y_{k} \neq y_{j}\right)$ where the original labels are used for data partitioning.

\noindent {\bf Cluster-boundary.} Since we want to introduce a general approach that does not rely on label information, we employ a third sampling strategy that is in principle similar to the class-boundary approach but does not require labels. Let $\mZ \in \mathbb{R}^{m^{\prime} \times p}$ be the stacked representations of a neural network model for every image in $\mathcal{D}_{\mathrm{image}}$. The representations can essentially be computed for any layer of a model. Here, we use the image encoder for image/text models and the \texttt{cls} token representation of the penultimate layer for any other model (since we only use ViT based models).

We cluster the representation spaces $\mZ$ and $\mZ^{\prime} \coloneqq \left( \mW\mZ^{\top} + \left(\vb_{1}, \ldots, \vb_{m^{\prime}}\right) \right)^{\top}$\footnote{Each of the $m$ column vectors on the right-hand side is the same $\vb$.} respectively (see Sec.~\ref{eq:ud_objective} for how the transformation variables $\mW$ and $\vb$ are computed) into $c$ representation clusters where $c$ can be regarded as similar to $C$, the number of labels in the original dataset. We use the Elbow criterion to select $c$. For all of our main experiments we set $c=500$. Hence, the ImageNet dataset is transformed into a dataset of image-cluster instead of image-label pairs. Let $\mathcal{D}_{\mathrm{image}}^{\prime} \coloneqq \left(x_{i}, y^{\prime}_{i}\right)_{i=1}^{m^{\prime}}$ be the modified ImageNet dataset of image and cluster pairs. After the clustering, we apply the same sampling method as for class-boundary triplets: for each triplet we choose uniformly at random two images without replacement from one cluster and one image from a different cluster. Thus, a triplet generated via cluster-boundary sampling can be defined as the following set $\mathcal{S} \coloneqq \{x_{i}, x_{j} ,x_{k}\}$ with the constraint $\left(y^{\prime}_{i} = y^{\prime}_{j} \neq y^{\prime}_{k}\right) \lor \left(y^{\prime}_{i} \neq y^{\prime}_{j} = y^{\prime}_{k}\right) \lor \left(y^{\prime}_{i} = y^{\prime}_{k} \neq y^{\prime}_{j}\right)$ where instead of the original labels we use the cluster labels for partitioning the data.

\paragraph{Triplet response generation.}
\label{appx:alignet-response-gen}

Let $\mathcal{D}_{\mathrm{triplets}} \coloneqq \left(\{x_{i}, x_{j} ,x_{k}\}_{s}\right)_{s=1}^{n^{\prime}}$ be the dataset of sampled ImageNet triplets for which we want to collect responses using transformed model representations. Note that we can sample an arbitrary number of triplets --- upper-bounded by the binomial coefficient $\binom{n^{\prime}}{k}$ with $k=3$ --- and can thus set $n^{\prime}$ to essentially any natural number. For the experiments that we report in the main text we set $n^{\prime}=10^{7}$ because we found a larger $n^{\prime}$ to not yield any downstream task improvements. For now, we regard our surrogate model as a blackbox model with transformed ImageNet representations $\mZ^{\prime} \coloneqq \left( \mW\mZ^{\top} + \left(\vb_{1}, \ldots, \vb_{m^{\prime}}\right) \right)^{\top} \in \mathbb{R}^{m^{\prime} \times p}$  where the affine transformation was found via UD optimization (see Eq.~\ref{eq:ud_objective}).

Given transformed representations $\vz^{\prime}_{1}$, $\vz^{\prime}_{2}$, and $\vz^{\prime}_{3}$ of the three images in a triplet, we can construct a similarity matrix $\mS^{\prime} \in \mathbb{R}^{3\times 3}$ where $S^{\prime}_{i,j} \coloneqq {\vz^{\prime}}_{i}^{\top} \vz^{\prime}_{j}$ is the dot product between a pair of representations. Similarly to how we do this for learning the UD transformation (see Sec.~\ref{appx:method-ooo}), we identify the closest pair of images in a triplet as $\argmax_{i, j > i} S^{\prime}_{i,j}$ with the remaining image being the odd-one-out. Let $\mathcal{D}_{\mathrm{align}} \coloneqq (\{x_{a}, x_{b}\}_{s}\mid\{x_{i},x_{j},x_{k}\}_{s})_{s=1}^{n^{\prime}}$ then constitute the final AligNet dataset of ImageNet triplets and corresponding model responses, where $\{x_{a}, x_{b}\} \subset \{x_{i}, x_{j}, x_{k}\}$ and $\{x_{a}, x_{b}\}$ is the image pair that was chosen by the transformed model representations to have the highest pairwise similarity. The model choices are the closest approximation to the human choices due to the UD transformation.

\subsubsection{Objective function}
\label{appx:alignet-objective}

Let $f_{\theta}$ be a neural network function parameterized by $\theta$, the set of its weights and biases. For every input image $x$ that the function $f_{\theta}(x)$ processes it yields a representation $f_{\theta}(x) = z$. Here, $z$ refers to the image encoder representation of image/text models or the \texttt{cls} token representation before the final linear layer for other model types. From the representations of the three images in a triplet, we can again construct a similarity matrix $\mS^{\dagger} \in \mathbb{R}^{3\times 3}$ where $S^{\dagger}_{i,j} \coloneqq z_{i}^{\top} z_{j}$ is the dot product between a pair of image representations. The AligNet loss function is defined as the following KL divergence between teacher and student triplet probabilities,
\begin{align}
    \mathcal{L}_{\text{alignet}}(\mS^{\prime}, \mS^{\dagger}) \coloneqq \frac{1}{B}\sum_{s=1}^{B}{\color{black}{\sigma\left(\left[S^{\prime}_{i,j},S^{\prime}_{i,k},S^{\prime}_{j,k}\right], \tau^{\prime}\right)_{s}}}\log \sigma \left(\left[S^{\prime}_{i,j},S^{\prime}_{i,k},S^{\prime}_{j,k}\right], \tau^{\prime}\right)_{s} - \notag \\ \sigma\left(\left[S^{\prime}_{i,j},S^{\prime}_{i,k},S^{\prime}_{j,k}\right], \tau^{\prime}\right)_{s}  \log \sigma\left(\left[S^{\dagger}_{i,j},S^{\dagger}_{i,k},S^{\dagger}_{j,k}\right], \tau^{\dagger}\right)_{s},
\label{eq:alignet_kld_loss}
\end{align}
where $\tau^{\prime}=1$ and $\tau^{\dagger}>1$ and $B$ is the batch size. We find $\tau^{\dagger}$ via grid search and set it to $\tau^{\dagger}=100$ for all of our experiments. Recall that $\sigma$ is a $\softmax$ function that models the probabilities over the three image similarity pairs (see Eq.~\ref{eq:triplet-likelihood}). The final AligNet objective is defined as the following minimization problem,
\begin{equation}
\argmin_{\theta}~\mathcal{L}_{\mathrm{alignet}}(f_{\theta}) + \lambda \norm{\textstyle \theta^{*}-\theta^{\dagger}}_{2}^{2},
\label{eq:alignet_objective}
\end{equation}
where $\theta^{*}$ are the parameters of the pretrained base student model and $\theta^{\dagger}$ are the parameters of the fine-tuned student model. This $\ell_{2}$-regularization, which we refer to as \emph{weight decay to initialization}, encourages the fine-tuned set of parameters to stay close to its base during training. It is similar to the regularization employed for learning the UD transformation (see Eq.~\ref{eq:ud_objective}) but adapted to the set of all model parameters rather than a linear transform.

\subsubsection{Surrogate teacher model}
\label{appx:teacher_model}

\citet{muttenthaler2023human} showed that image/text models and models trained on large, diverse datasets are better aligned with human similarity judgments than vision models trained with a self-supervised learning objective or supervised models trained on ImageNet.
Thus, we use the best-performing image/text model according to various computer vision benchmarks at the time of writing this manuscript as our teacher model.
This model is called SigLIP \citep{zhai2023sigmoid}.
SigLIP, similar to CLIP \citep{radford2021learning} and ALIGN \citep{jia2021scaling}, is trained via contrastive language-image pretraining using millions of image/text pairs.
The difference between CLIP and SigLIP is that SigLIP uses a paired sigmoid loss instead of the standard $\softmax$ function usually used for pretraining image/text models via cross-entropy.
Image/text pretraining allows the model to learn an aligned representation space for images and text; thus, adding more semantic information about the objects in an image to the model representations.

We use the SigLIP-So400m variant of SigLIP as our teacher model. The So400m variant uses an optimized ViT backbone whose performance is similar to one of the largest ViTs, ViT-g/14 \citep{alabdulmohsin2023getting} while having fewer parameters and thus being smaller. The number of parameters of SoViT-400m/14 lies somewhere between ViT-L/16 and ViT-g/14.
The output dimensionality of the image and text encoder representations of SoViT-400m/14 is $p$=$1152$ each.
We align the image encoder representations with human odd-one-out choices using the UD optimization outlined in Eq.~\ref{eq:alignet_objective}.
This allows us to increase the triplet odd-one-out accuracy of SigLIP-So400m from $44.24\%$ to $61.7\%$ (see rightmost column in Tab.~\ref{tab:rsa_results_ckh}) which is close to the human-noise ceiling of $66.67\%$ for THINGS \citep[cf.][]{hebart2020revealing} and thus among the best human-aligned models without AligNet fine-tuning \citep[cf.][]{muttenthaler2023human}. Note that this is a relative increase in performance of $39.47\%$. Throughout this manuscript we use the human-aligned version of SigLIP-So400m as the \emph{surrogate teacher model} for generating human-like similarity judgments and distilling human-like similarity structure into student VFMs.
We select a diverse and representative set of student VFMs.

\subsubsection{Student Models}
\label{appx:student_models}

Since previous research has demonstrated that a model's architecture has no significant impact on the degree of alignment with human similarity judgments \citep{muttenthaler2023human, muttenthaler2023improving}, we use the same architecture for all student models that we fine-tune on AligNet.
Specifically, we use the Vision Transformer \citep[ViT;][]{dosovitskiy2020image} for the backbone of each student model.
We use the ViT rather than CNN-based model because ViTs have recently emerged as the dominant neural network architecture for computer vision application and VFMs.
Every large VFM that is used in practice is based on the ViT \citep{tschannen2023image, chen2023pali-3, zhai2023sigmoid, beyer2024paligemma}. Unless otherwise mentioned, we use the base model size, i.e., ViT-B.
ViT-B has $12$ attention layers and an internal (hidden) representation size of $p=768$. It has been shown that both the training data and the objective function have a substantial impact on the degree of alignment with human behavior.
Thus, we use student models that were trained on different pretraining task with different training data and objective functions.

\emph{Supervised} pre-training is still the prevailing mode of training computer vision models.
Therefore, we trained ViT-B on the popular ImageNet dataset consisting of 1.4M natural images\,\citep{deng2009imagenet}. To examine how model performance changes as a function of the size of a model, we train ViT instances of three different sizes on ImageNet: ViT-S/16, ViT-B/16, ViT-L/16.
The image patch size is the same for each of those models.
To evaluate the effect of AligNet on \emph{self-supervised} pretraining, we use pretrained DINOv1\,\citep{caron2021emerging} and DINOv2\,\citep{oquab2023dinov2} models of which DINOv1 was pretrained on ImageNet and DINOv2 was pretrained on a different, larger image dataset as denoted below.
In addition, we investigate \emph{multimodal training of vision models} that add textual information both in the form of \emph{image captioning} via the CapPa model\,\citep{tschannen2023image}, as well as \emph{contrastive language-image pretraining} (CLIP) via SigLIP\,\citep{zhai2023sigmoid}.
The latter model is considered \emph{state-of-the-art} on many downstream computer vision applications and is used as the image embedding model in modern large visual-language models\,\citep[VLMs;][]{chen2023pali-3,beyer2024paligemma}.
The full list of student models that we consider in our analyses is the following:
\begin{itemize}
    \item ViT-\{S,B,L\}
    \begin{itemize}
        \item \emph{Training data}: ImageNet \citep{deng2009imagenet}
        \item \emph{Objective}: Supervised learning
    \end{itemize} 
    \item {{CLIP (ViT-B) (see \cref{appx:clip})
    \begin{itemize}
        \item \emph{Training data}: WebImageText \citep{radford2021learning}
        \item \emph{Objective}: CLIP \citep{radford2021learning}
    \end{itemize}}}
    \item SigLIP (ViT-\{B,SO400M\})
    \begin{itemize}
        \item  \emph{Training data}: WebLI \citep{zhai2023sigmoid}
        \item \emph{Objective}: CLIP \citep{radford2021learning}
    \end{itemize} 
     \item {{SigLIP2 (ViT-B) (see \cref{appx:siglip2})
    \begin{itemize}
        \item \emph{Training data}: WebLI \citep{zhai2023sigmoid}
        \item \emph{Objective}: Combination of various objectives (see \citet{tschannen2025siglip} for details)
    \end{itemize}}}
    \item DINOv1 (ViT-B)
    \begin{itemize}
        \item  \emph{Training data}: ImageNet \citep{deng2009imagenet}
        \item \emph{Objective}: Self-supervised image pretraining \citep{caron2021emerging}
    \end{itemize} 
    \item DINOv2 (ViT-B)
    \begin{itemize}
        \item  \emph{Training data}:  DINOv2 data (see \citet{oquab2023dinov2} for details)
        \item \emph{Objective}: Self-supervised teacher-student distillation \citep{oquab2023dinov2}
    \end{itemize}
    \item CapPa (ViT-B)
    \begin{itemize}
        \item  \emph{Training data}: JFT-3B (Google proprietary dataset)
        \item \emph{Objective}: Multi-modal image captioning \citep{tschannen2023image}
    \end{itemize}
    \item {{Randomly initialized ViT-B (see \cref{appx:scratch})
    \begin{itemize}
        \item \emph{Training data}: AligNet
        \item \emph{Objective}: AligNet objective (see \cref{eq:alignet_kld_loss} in \cref{appx:alignet-objective})
    \end{itemize}}}
\end{itemize}

\subsection{Representational Similarity Analysis (RSA)}
\label{appx:rsa}

\subsubsection{Multi-arrangement task}
\label{appx:multi_arrangement}

Human similarity judgments for \citet{king2019} and \citep{cichy2019} were obtained by using a multi-arrangement task. In a multi-arrangement task, participants are presented with a computer screen showing images of several different objects. The participants are asked to arrange the images into semantically meaningful clusters, given the instruction that images of objects that lie close together are considered more similar. From this arrangement, one can infer pairwise (dis-)similarities of the objects and average those across all participants to obtain a representative (dis-)similarity matrix.

\subsubsection{Likert scale}
\label{appx:likert}

In \citet{peterson2016,peterson2018evaluating}, pairwise similarity judgments were obtained by asking human participants to rate the similarity of pairs of objects on an ordinal scale that ranges from 0 (``not similar at all'') to 10 (``very similar''). The pairwise similarity ratings can be averaged across the different participants which in turn yields a matrix of similarities between pairs of objects.

\subsubsection{Neural network representations} 
\label{appx:rsa_neural_reps}

Representation Similarity Matrices (RSMs) for neural network representations are obtained by first embedding the same set of images that were presented to the human participants in the $p$-dimensional latent space of a model. The latent space could be any layer of a neural network. For the base models, we use the representations of the image encoder for SigLIP and the \texttt{CLS} token of the penultimate layer for CapPa, DINOv2, and ViT-B. We do this because previous work has shown that the penultimate layer space and the image encoder space of image/text models respectively yield the highest similarity to human behavior \citep{peterson2018evaluating,peterson2019, muttenthaler2023human}. 

After embedding the images into the neural net's latent space, we get a representation matrix $\mX \in \mathbb{R}^{n \times p}$ for the $n$ images in the data. Instead of simply computing the dot-product similarity matrix $\mS \coloneqq \mX\mX^{\top}$, in RSA one typically uses either a cosine similarity or a Pearson correlation kernel to compute the affinity matrix,
\begin{align*}
&~\text{cos}(\vx_{i}, \vx_{j}) \coloneqq \frac{\vx_{i}^{\top}\vx_{j}}{||\vx_{i}||_{2} ||\vx_{j}||_{2}};
&~\phi(\vx_{i}, \vx_{j})\coloneqq \frac{\left(\vx_{i}-\bar{\vx_{i}}\right)^{\top}\left(\vx_{j}-\bar{\vx_{j}}\right)}{||\left(\vx_{i}-\bar{\vx_{i}}\right)||_{2} ||\left(\vx_{j}-\bar{\vx_{j}}\right)||_{2}},
\end{align*}
where the cosine similarity kernel function $\text{cos}(\vx_{i}, \vx_{j})$ or the Pearson correlation kernel function $\phi(\vx_{i}, \vx_{j})$ is applied to every ($\vx_{i}, \vx_{j}$) vector pair of the matrix $\mX$ for obtaining the final representational similarity matrix $\mS^{\prime} \in \mathbb{R}^{n \times n}$. Here, we use the Pearson correlation kernel function  $\phi(\vx_{i}, \vx_{j})$ to obtain a neural net's RSM. Pearson correlation is the centered version of cosine similarity and the ranking of the obtained similarities does not differ between the two kernel functions but Pearson correlation first centers the vectors to have zero mean and is therefore a more robust measure. For obtaining RSMs with transformed representations, the transforms are first applied to $\mX$ before computing $\mS^{\prime}$.

\subsection{Levels data}
\label{appx:levels}

\noindent{\bf Participants}. We recruited $N=508$ participants (209 female, 289 male, 3 diverse, \(N = 7\) missing demographic information due to revocation of study consent; mean age = 31.75 $\pm$ SD: 8.04  years) online via Prolific Academic (\href{https://www.prolific.ac}{https://www.prolific.ac}). The eligibility criteria were that participants had to be between 18 and 50 years old, fluent in English, have a normal or corrected-to-normal vision, no colorblindness, and have a minimum approval rating of 95\% on Prolific. Participants provided informed consent before starting the experiment. The experiment lasted approximately 45 minutes. Participants were reimbursed with £7.7 for completing the experiment and received an additional bonus payment of £0.77. Partial payments were made if the experiment was not completed due to technical issues (\(N = 6\)) or early termination by the participant (\(N = 1\)). Participants performing below 90\% correct on catch trials (\(N = 19\), 3 female, 16 male), or failing to respond in the allotted time window (15s) in more than 10 trials (\(N = 9\), 4 female, 4 male, 1 diverse) were excluded. Thus, (\(N = 473\)) participants remained in the dataset (202 female, 269 male, 2 diverse; mean age = 31.82 $\pm$ SD = 8.03 years). Of these participants, (\(N = 448\)) were each tested with a different selection of triplets, while ensuring that each triplet was presented \(N = 5\) times across the entire sample of participants (see information on stimuli sampling below). Due to a server glitch during trial assignment, the remaining (\(N = 25\)) participants shared their exact triplet selection with one other participant in the sample. These (\(N = 25\)) participants were excluded from the response times (RT) and uncertainty estimation (see Sec.~\ref{sec:alignment_levels_abstraction}) to restrict analysis to participants with different sets of triplets.  The experiment was approved by the internal review board (IRB) of the Max Planck Institute for Human Development.

\noindent{\bf Stimuli}. The experimental stimuli were images taken from the ImageNet dataset \citep{deng2009imagenet}. Another 9 images were used for instructions only and depicted natural objects selected from the Bank of Standardized Stimuli \citep[BOSS;][]{brodeur2014bank}, available at \href{https://drive.google.com/drive/folders/1FpnEFkbqe_huRwfsCf7gs5R1zuc1ZOkn}{BOSS}. We grouped the visual stimuli presented in the triplets according to different levels of abstraction: {\color{coarse-green}{\emph{coarse-grained semantic}}}, which comprised three images from three different categories; {\color{fine-blue}{\emph{fine-grained semantic}}}, showing three images from the same category; and {\color{boundary-red}{\emph{class-boundary}}}, where two images were from the same and one from a different category.

Instead of randomly sampling triplets---which would reproduce dataset biases---we stratified sampling by superclasses. ImageNet classes follow the WordNet hierarchy \citep{deng2009imagenet, russakovsky2014imagenet}, which includes higher-level classes. For instance, all dog breeds can be summarized as a single dog superclass. To avoid presenting dogs, birds, and other fine-grained classes that are overrepresented in ImageNet more frequently to the participants than other categories, we grouped the ImageNet classes into $717$ coarse-grained WordNet superclasses. We uniformly at random sampled images from those $717$ superclasses to construct the different kinds of triplets. Note that for all superclasses with more than one class, we uniformly at random chose one subclass and either uniformly at random sampled one image, two images (without replacement), or three images (without replacement) from that subclass, depending on the triplet type. For most superclasses that were comprised by a single subclass only, i.e., a \emph{one-to-one-mapping}, we could skip the subclass sampling part. Triplet sampling resulted in (\(N = 450\)) predefined experiment trial sets, of which (\(N = 448\)) were used for testing. Across these, each triplet was presented within (\(N = 5\)) different experiment files. This sampling process ensured a balanced distribution of triplets across the sample, and the repetition of each triplet in five different participants allowed for the calculation of an uncertainty distribution for each triplet.


\noindent{\bf The triplet odd-one-out task}. On each trial, participants were presented with a triplet of images (\(i, j, k\)). Participants were asked to select the image that was the most different from the other two, i.e., the odd-one-out. During the instructions, participants saw different triplets with increasing ambiguity regarding which image would likely be picked as the odd-one-out. Participants were given explanations for potential odd-one-out choices, clarifying that decisions could be based on different criteria, such as semantic or perceptual features of the shown images. 

\noindent{\bf Procedure}. The experiment was run online using  \href{https://www.jspsych.org/7.3/}{jsPsych v7.3.3} and custom plugins. Participants were asked to provide demographic information, including their age and gender. Thereafter, they viewed written instructions about the task and performed six practice trials (2 trials per triplet level of abstraction). Participants were free to repeat the instructions until they felt confident to perform the experiment. The experiment proper comprised $N = 330$ experiment trials. Each trial started with a fixation cross (1s), followed by the presentation of a triplet (max. 15s). Participants were asked to select the ``odd-one-out'' using the right, left, or downward facing arrow keys on their keyboard. Responses could be entered between 1-15s after triplet onset, after which the next trial started. Trials in which participants failed to submit a response were rare (M = 0.27\% of trials; min = 0.00\%, max = 6.06\%). The serial order of triplet types (e.g., fine-grained or coarse-grained semantic) and ImageNet classes (e.g., dogs or birds) was counterbalanced across the experiment. We additionally counterbalanced the serial position of trial types across participants using a Latin-Square approach \citep[LSD;][]{grant1948latin}. Participants could take short breaks (self-paced for up to 2 minutes) after \(N = 50, 150,\) and, \(200\) experiment trials; Experimental trials were interleaved with \(N = 16\) catch trials (class border triplets), which were predefined based on low model uncertainty and 100\% agreement among participants on these specific triplets during piloting. Catch trial performance was used as an indicator of adequate task engagement (see participant inclusion criteria above).

\noindent{\bf Preprocessing of human response times and uncertainty estimation}. Descriptive statistics on response times (RT) and uncertainty estimation (see Sec.~\ref{sec:alignment_levels_abstraction}) were calculated based on participants with unique experimental trial sets (\(N = 448\)). The RT data were log transformed (\(\log(\text{RT})\)), in accordance with current best practices for RT analysis. Trials with RTs longer than 10 seconds were excluded from analysis (on average M = 2.64\% of trials per participant). Since responses could be given no earlier than 1 s after triplet onset (see \textit{Procedure} above), no lower bound was set for RT-exclusion. To estimate uncertainty (in terms of the level of (dis-)agreement among observers) for each triplet, we used the discrete (Shannon) entropy of the response distribution across participants.
\newpage

\section{Additional/Detailed Results}

In addition to the results that we presented in the main paper, here, we report the same set of results in more detail and add further analyses that we touched upon in the main part but did not go into much detail. We start by presenting findings from a comprehensive set of evaluations that we performed to test the alignment of model representations with human similarity judgments both for similarity judgments collected in previous efforts \citep[cf.][]{peterson2018evaluating,cichy2019,king2019,hebart2020revealing} and for our own similarity judgment dataset Levels (see Sec.~\ref{appx:levels}). Subsequently, we demonstrate benefits of our method for machine learning downstream tasks that test generalization, such as \emph{few-shot learning} and \emph{out-of-distribution detection}. Finally, we perform a rigorous qualitative analysis of the changes in the model representations after applying our alignment framework to the various student models.

\begin{figure}[ht!]
    \centering
    \includegraphics[width=1.0\textwidth]{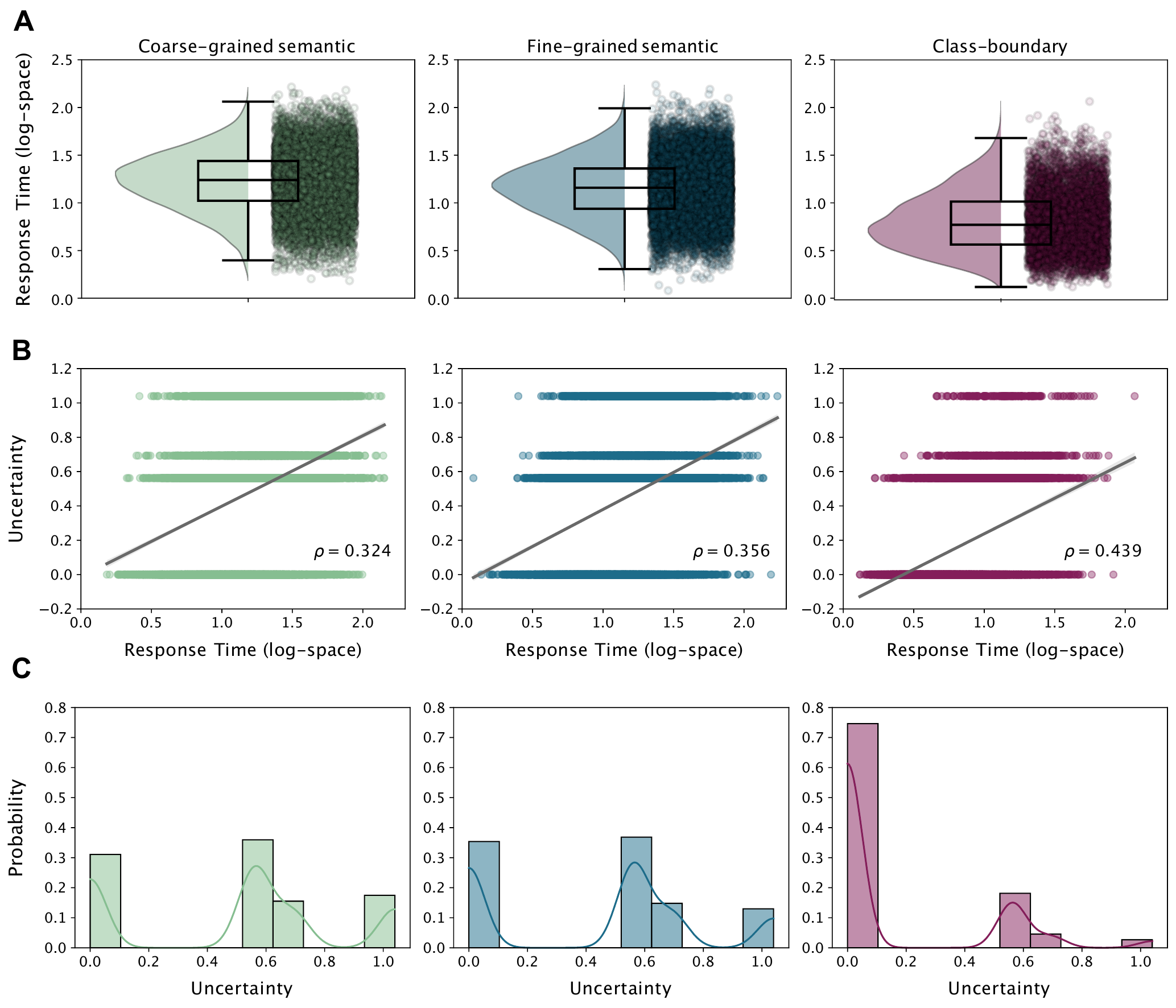}
    \caption{Human response times and disagreement levels for the different triplet types. \textbf{A}: Distribution of human response times (in log-space) for the three different triplet type settings. \textbf{B}: Spearman rank correlation and linear regression fit of the participant disagreement levels/uncertainties with the participants' response times (in log-space). \textbf{C}: Probability distributions of the participant disagreement levels/uncertainties for the three triplet type settings.}
    \label{fig:human_uncertainties_and_rts}
\end{figure}

\subsection{Human Alignment}
\label{appx:human_alignment}

In this section, we present additional evaluation results regarding the alignment of model representations with human similarity judgments, complementing Sec.~\ref{sec:human_alignment}. For every student vision foundation model, we consider four settings of model representations and their refinements,
\begin{itemize}
    \item \emph{Original}: We use the pretrained representations of every student model without any changes to their representation space.
    \item \emph{Distillation without alignment}: We perform AligNet fine-tuning without the first step (see \cref{fig:fig1}) of aligning the teacher model with the human similarity judgments in THINGS. That is, we distill the teacher similarity structure of the pretrained, non-aligned SigLIP-So400m model (see Methods) 
    into a student VFM.
    \item \emph{Uncertainty Distillation (UD)}: We learn a linear transform of the models' pretrained representation space  into a human global object similarity space (using the THINGS triplet odd-one-out choices) while preserving the model's local similarity structure (see Eq.~\ref{eq:ud_objective}). We remark that this does not involve any step of the AligNet framework. The weights of the student models are frozen during the UD optimization.
    \item \emph{AligNet/Soft-alignment}: We learn a new representation space using the full AligNet framework as outlined in Sec.~\ref{sec:methods_alignet}. This can be referred to as \emph{soft-alignment}.
\end{itemize}

\subsubsection{Detailed results for RSA}

\begin{figure}[ht!]
    \centering
    \includegraphics[width=\textwidth]{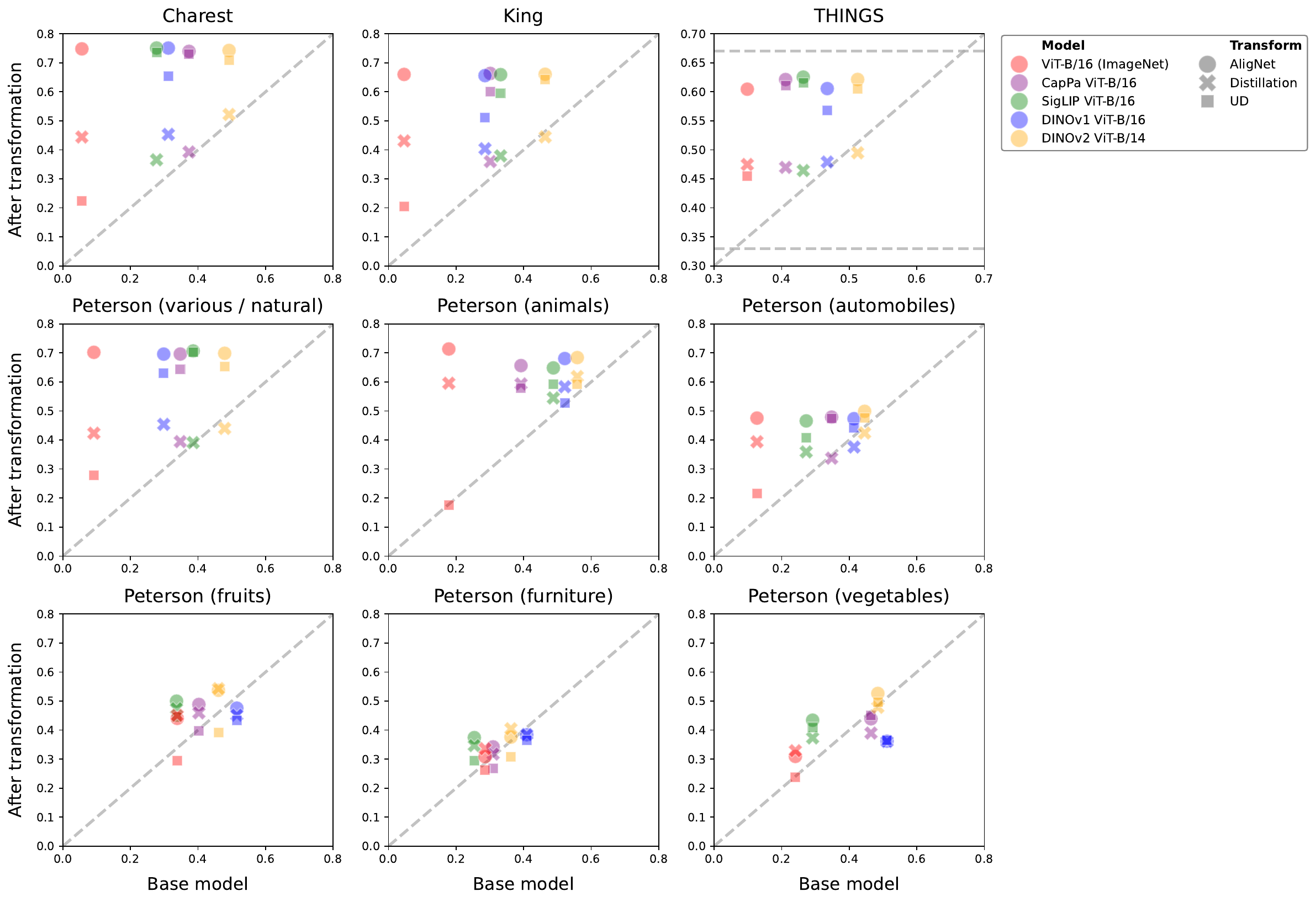}
    \caption{Alignment of model representations with human similarity judgments for the four different representation settings \emph{Original}, \emph{Distillation}, \emph{UD}, and \emph{AligNet} for all models with a ViT-B backbone. The $x$-axis shows the degree of alignment for the base model, i.e., the \emph{original} representations, whereas the $y$-axis displays the degree of alignment after linearly transforming the representations via UD or fine-tuning the models using distillation (without alignment) or the full AligNet framework. For all datasets but THINGS alignment is measured as the Spearman rank correlation coefficient between human and model RSMs. For THINGS, alignment is measured as triplet odd-one-out accuracy, i.e., the fraction of triplets for which models selected the same odd-one-out object as the human participants. Dashed horizontal lines for THINGS depict random guessing ($0.333$) and the human noise-ceiling/reliability score ($0.667$) respectively.}
    \label{fig:rsa_scatter_main_models}
\end{figure}

In \cref{tab:rsa_results_ckh} we present the average alignment scores for all student models that we considered in our analyses (see Methods) 
and each of the four representation settings defined above for the datasets from \citet{cichy2019}, \citet{king2019}, and \citet{hebart2020revealing}. For the former two datasets, alignment is measured via RSA using Spearman rank correlation between the upper triangulars of the pairwise similarity matrices. For those datasets, human similarity judgments were collected using a multi-arrangment task where participants were asked to arrange (natural) objects on a computer screen (see Sec.~\ref{appx:multi_arrangement} for details). For the THINGS dataset, alignment is measured via triplet odd-one-out accuracy, i.e., the fraction of triplets for which models selected the same odd-one-out object as the human participants. 

\begin{table}[ht!]
\centering
\resizebox{1\textwidth}{!}{%
\begin{tabular}
{l|cccc|cccc|cccc}
\toprule
& \multicolumn{4}{c}{\citet{cichy2019}} & \multicolumn{4}{c}{\citet{king2019}} & \multicolumn{4}{c}{\citet{hebart2020revealing}} \\
Model $\setminus$ Fine-tuning & Original & Distillation & UD & AligNet & Original & Distillation & UD & AligNet & Original & Distillation & UD & AligNet \\
\midrule
ViT-S & $0.083$ & $0.520$ & $0.367$ & $0.743$ & $0.143$ & $\mathbf{0.486}$ & $0.353$ & $\mathbf{0.665}^{\dagger}$ &  $39.28\%$ &  $\mathbf{49.97}\%$ & $50.01\%$ &  $59.44\%$ \\ 
ViT-B & $0.056$ & $0.444$ & $0.225$ &  $0.748$ & $0.046$ & $0.431$ & $0.204$ &  $0.660$ &  $34.92\%$ & $47.47\%$ & $45.47\%$ & $60.45\%$ \\ 
ViT-L & $0.041$ & $\mathbf{0.540}$ & $0.227$ &  $\mathbf{0.751}^{\dagger}$ & $0.095$ & $0.463$ & $0.201$ &  $0.664$ & $34.84\%$ & $49.30\%$ &  $44.40\%$ & $60.75\%$ \\ 
CapPa ViT-B & $0.373$ & $0.393$ & $0.729$ & $0.739$ & $0.301$ & $0.360$ & $0.599$ &  $0.663$ &  $40.58\%$ & $46.97\%$ & $61.04\%$ &  $62.11\%$ \\ 
DINOv1 ViT-B & $0.312$ & $0.453$ & $0.653$ &  $\mathbf{0.751}^{\dagger}$ & $0.286$& $0.404$ & $0.512$ & $0.656$ &  $46.75\%$ & $47.91\%$ & $56.80\%$ &  $60.56\%$ \\ 
DINOv2 ViT-B &  $\mathbf{0.492}$ & $0.522$ & $0.710$ &  $0.743$ & $\mathbf{0.464}$ & $0.445$ & $\_$ & $0.661$  & $\mathbf{51.24}\%$ & $49.49\%$ & $60.51$\% &  $62.13\%$ \\ 
SigLIP ViT-B &  $0.277$ & $0.366$ & $\mathbf{0.735}$ & $\mathbf{0.751}^{\dagger}$ & $0.332$ & $0.379$ & $0.595$ & $0.659$ & $43.20\%$ & $46.44\%$ & $61.53\%$ & $\mathbf{62.54}\%^{\dagger}$ \\
\midrule
SigLIP So400m (Teacher) &  $0.226$ & \_  & $0.723$  &  \_  & $0.267$  & \_  & $\mathbf{0.638}$  & \_&  $44.24\%$ & \_ & $\mathbf{61.70}\%$ &  \_ \\
\bottomrule
\end{tabular}%
}

\caption{Human alignment results for three different human similarity judgment datasets. The datasets from \citet{cichy2019} and \citet{king2019} were collected using a multi-arrangement task. For those datasets, alignment with human similarity judgments is measured via RSA using the Spearman rank correlation. The dataset from \citet{hebart2020revealing} used a triplet odd-one-out task for collecting human judgments. Here, alignment with human choices is measured via triplet odd-one-out accuracy (\%). Bold face indicates highest performance within a single column and $\dagger$ indicates best performance for a dataset overall.}
\label{tab:rsa_results_ckh}
\end{table}

\begin{figure}[ht!]
    \centering
    \includegraphics[width=\textwidth]{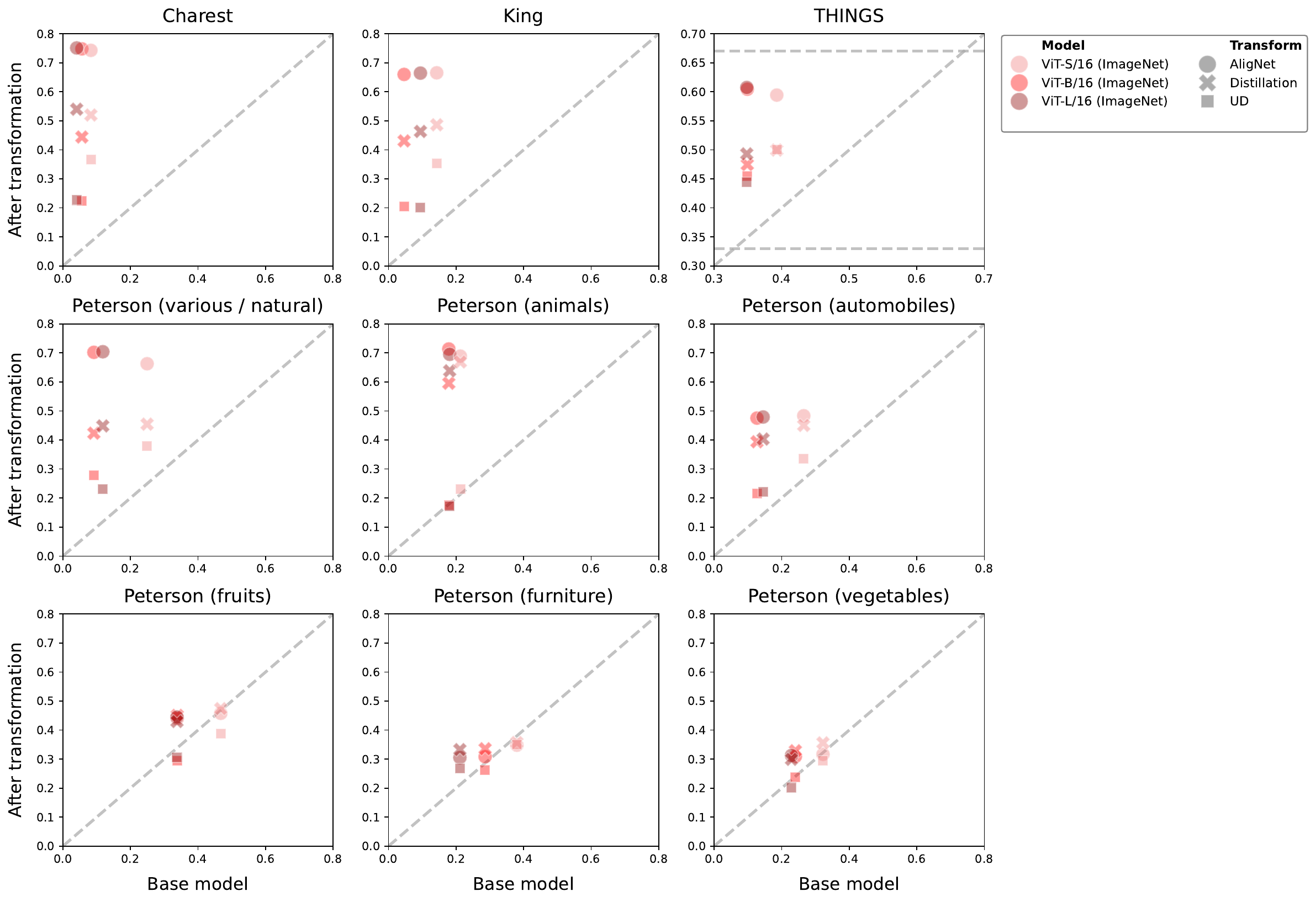}
   \caption{Alignment of model representations with human similarity judgments for the four different representation settings \emph{Original}, \emph{Distillation}, \emph{UD}, and \emph{AligNet} for the three supervised ImageNet-trained ViT models. The $x$-axis shows the degree of alignment for the base model, i.e., the \emph{original} representations, and the $y$-axis displays the degree of alignment after linearly transforming the representations via UD or fine-tuning the models using distillation (without alignment) or the full AligNet framework. For all datasets but THINGS alignment is measured as the Spearman rank correlation coefficient between human and model RSMs. For THINGS, alignment is measured as triplet odd-one-out accuracy, i.e., the fraction of triplets for which models selected the same odd-one-out object as the human participants. Dashed horizontal lines for THINGS depict random guessing ($0.333$) and the human noise-ceiling ($0.667$) respectively.}
    \label{fig:rsa_scatter_vits}
\end{figure}

We find that base models---that is the \emph{original} representation setting (see above)---performed poorly for almost every human similarity judgment dataset (see Tab.~\ref{tab:rsa_results_ckh} and \cref{fig:rsa_scatter_main_models}). The differences among the models in this setting is significant.
While DINOv2 ViT-B achieves a Spearman rank correlation coefficient of $\rho=0.492, p<0.001$ for the dataset in \citet{cichy2019} and an odd-one-out accuracy of $51.21\%$ for the THINGS dataset, ViT-L is not correlated with the human similarity judgments from \citet{cichy2019} and \citet{king2019} ($\rho=0.041$ and $\rho=0.095$ respectively) and shows a close to chance-level odd-one-out accuracy of $34.84\%$ for the THINGS dataset (see Tab.~\ref{tab:rsa_results_ckh} and $x$-axis in \cref{fig:rsa_scatter_vits}).
AligNet fine-tuning significantly improved the alignment scores of all models for all datasets.
In addition, it minimized the differences between the models to an extent that their differences in the degree of alignment are not statistically significant.
Thus, AligNet fine-tuning models are equally well aligned with human similarity judgments irrespective of their architecture, pretraining data, and objective function.

In the top row of \cref{fig:rsa_scatter_main_models} we show the performance of the transformed representations as a function of the base model performance for those datasets. We can see that AligNet fine-tuning significantly improves upon the base model representations ($x$-axis). It yields the most human-aligned models across all three datasets (depicted by the circles) and often substantially improves upon the linearly aligned representations (depicted by the squares).

\begin{table}[ht!]
\footnotesize
\begin{tabular}{l|cccc|cccc}
\toprule
& \multicolumn{4}{c}{\citet{peterson2016,peterson2018evaluating} ({\color{coarse-green}{coarse-grained}})} & \multicolumn{4}{c}{\citet{peterson2016,peterson2018evaluating} ({\color{fine-blue}{fine-grained}})} \\
Model $\setminus$ Fine-tuning & Original & Distillation & UD & AligNet & Original & Distillation & UD & AligNet \\
\midrule
ViT-S & $0.249$ & $\mathbf{0.454}$ & $0.379$ & $0.663$ & $0.330$ & $0.461$ & $0.320$ & $0.459$ \\ 
ViT-B & $0.092$ & $0.423$ & $0.278$ &  $0.702$ & $0.234$ & $0.421$ & $0.237$ &  $0.449$ \\ 
ViT-L & $0.118$ & $0.448$ & $0.231$ &  $0.704$ & $0.221$ & $0.421$ &  $0.234$ & $0.447$ \\ 
CapPa ViT-B & $0.347$ & $0.394$ & $0.644$ & $0.696$ & $0.383$ & $0.419$ & $0.434$ &  $0.481$ \\ 
DINOv1 ViT-B & $0.299$ & $\mathbf{0.454}$ & $0.631$ &  $0.696$ & $\mathbf{0.474}$ & $0.431$ & $0.426$ & $0.475$ \\ 
DINOv2 ViT-B &  $\mathbf{0.479}$ & $0.440$ & $0.653$ &  $0.699$ & $0.462$ & $\mathbf{0.494}$ & $\mathbf{0.452}$ & $\mathbf{0.524}^{\dagger}$  \\ 
SigLIP ViT-B &  $0.386$ & $0.391$ & $\mathbf{0.703}$ & $\mathbf{0.707}^{\dagger}$ & $0.329$ & $0.419$ & $0.430$ & $0.484$ \\ 
\bottomrule
\end{tabular}
\caption{Human alignment results for the dataset from \citet{peterson2016,peterson2018evaluating}. Human pairwise similarity judgments were collected using an ordinal Likert scale. Scores reflect the average Spearman rank correlation coefficients with the human pairwise similarity judgments. To obtain an average alignment score for the fine-grained setting, we averaged the performances across the five single category settings: \texttt{animals}, \texttt{automobiles}, \texttt{fruits}, \texttt{furniture}, and \texttt{vegetables}. Bold face indicates highest performance within a single column and $\dagger$ indicates best performance for a dataset overall.}
\label{tab:rsa_results_peterson}
\end{table}

Human similarity judgments from the dataset introduced in \citet{peterson2016, peterson2018evaluating} were collected using an ordinal Likert scale (see Sec.~\ref{appx:likert} for details). The authors collected pairwise similarity ratings for \texttt{various} natural objects with one image per category, similar to the other three human similarity judgment datasets. Thus, those pairwise similarity ratings reflect global {\color{coarse-green}{coarse-grained}} semantic structure. In addition, the authors collected pairwise similarity ratings for multiple images from a single category that reflect local {\color{fine-blue}{fine-grained}} semantics. The specific categories for which similarity ratings were collected are \texttt{animals}, \texttt{automobiles}, \texttt{fruits}, \texttt{furniture}, and \texttt{vegetables}. We average model performances across those five single category pairwise similarity rating datasets to obtain a single measure for fine-grained semantic and refer to it as the fine-grained dataset of \citet{peterson2016,peterson2018evaluating}. In Tab.~\ref{tab:rsa_results_peterson}, we report Spearman rank correlation coefficients of the model RSMs with the RSMs obtained from the human similarity ratings for all student models and the four representation settings. AligNet fine-tuning significantly improves performance across the board, even for the fine-grained setting. While the best model for the coarse-grained setting is AligNet fine-tuned SigLIP ViT-B ($\rho=0.707, p<0.001$), AligNet fine-tuned DINOv2 ViT-B is most human-aligned model for the fine-grained setting ($\rho=0.524, p<0.001$). AligNet fine-tuning significantly improves the Spearman rank correlation coefficients for all models compared to their original representations in the coarse-grained setting.

In the bottom two rows of \cref{fig:rsa_scatter_main_models} and \cref{fig:rsa_scatter_vits} respectively we show model performances individually for the single category pairwise similarity rating datasets. We observe that AligNet fine-tuning significantly improved the degree of alignment for all models for the \texttt{various}, \texttt{animals}, and \texttt{automobiles} categories but there does not appear to be a difference in performance between the different model transformation settings for the categories \texttt{fruits}, \texttt{furniture}, and \texttt{vegetables}. We hypothesize that this difference in alignment benefits stems from the fact that animal- and automobile-related concepts are more frequently represented in the THINGS dataset than fruits-, furniture-, and vegetable-related concepts \citep[cf.][]{hebart2019things, hebart2020revealing}. Moreover, THINGS is a human similarity judgment dataset that reflects global coarse-grained semantic, similar to the coarse-grained setting in Levels. Thus, it is rather surprising to find benefits for human local fine-grained semantic by either UD or fine-tuning on AligNet.

\subsubsection{Larger models benefit more from AligNet}

In \cref{fig:rsa_scatter_vits} we compare supervised ImageNet-trained ViTs of different sizes. Interestingly, the smallest version of the ViTs, ViT-S, achieved the best base model performance across the board. That is, their original representation space was better aligned with the different human similarity judgments than the representation spaces of the larger ViT backbones (see \cref{fig:rsa_scatter_vits} and Tables~\ref{tab:rsa_results_ckh},~\ref{tab:rsa_results_peterson}). However, after fine-tuning the models on AligNet the larger ViT versions were either better or equally-well aligned than ViT-S.

\finaledit{
\subsubsection{Descriptive statistics for human response times and uncertainty measures on Levels}

We examined the human response time (RT) differences across the three abstraction settings as an indicator of cognitive processing demands during odd-one-out selection (see \cref{fig:human_uncertainties_and_rts}A). As expected, responses were fastest for the class-boundary triplets (Mdn = 2.0\textcolor{blue}{4} s, SE = 0.03), significantly slower for fine-grained semantic triplets (Mdn = 3.0\textcolor{blue}{7} s, SE = 0.0\textcolor{blue}{5}; \(t(447) = 4\textcolor{blue}{5.920}\), \(p < 0.001\)), and slowest for coarse-grained semantic triplets (Mdn = 3.25 s, SE = 0.0\textcolor{blue}{6}; \(t(447) = 1\textcolor{blue}{4.160}\), \(p < 0.001\) compared to fine-grained). Unlike the simpler choices enabled by clear semantic boundaries, longer RTs at higher levels of semantic abstraction likely reflected more complex decisions, where participants may have weighed information across multiple  perceptual and semantic dimensions. Consistent with these RT results, we also found higher levels of cross-participant agreement (see \cref{fig:human_uncertainties_and_rts}C) in the class boundary condition compared to both the coarse-grained semantic and the fine-grained semantic conditions.

Participants' RTs were used as a proxy for decision uncertainty. RTs also correlated positively with cross-participant disagreement within each abstraction levels (coarse-grained semantic: \(r = 0.325\), \(p < 0.001\); fine-grained semantic: \(r = 0.364\), \(p < 0.001\); class-boundary: \(r = 0.443\), \(p < 0.001\); see SI Sec.~\ref{appx:levels} and \cref{fig:human_uncertainties_and_rts}B), suggesting that human RTs provided a reasonable approximation of decision uncertainty for model-to-human comparison.
}

\subsubsection{Detailed results for Levels}

In Tab.~\ref{tab:levels_results} we present triplet odd-one-out accuracies for every model that we considered in our analyses for each of the three levels of abstraction in the Levels dataset. 

\noindent{\bf {\color{coarse-green}{Global coarse-grained}}}. In this setting, AligNet fine-tuning improved alignment with human similarity judgments most significantly. Before fine-tuning, models achieved poor odd-one-out accuracies of $35.45\%$ (ViT-L) $-$ $57.38\%$ (DINOv2 ViT-B). Fine-tuning models on AligNet closed this gap and significantly improved the alignment between human and model responses to the extent that all models performed with odd-one-out accuracies of $65.70\%$ (DINOv1 ViT-B) $-$ $68.60\%$ (ViT-L) above the human-to-human reliability score of $61.92\%$. Interestingly, the most poorly performing model before fine-tuning, ViT-L, achieved the highest odd-one-out accuracy after fine-tuning on AligNet, with a relative improvement of almost $93.51\%$. Even the worst performing AligNet-tuned model, DINOv1 ViT-B, performed with an odd-one-out accuracy of $65.70\%$ significantly better than the best performing base model, DINOv2 ViT-B with $57.38\%$, and better than the best performing model after vanilla distillation without alignment, ViT-S with $58.82\%$, (see the {\color{coarse-green}{coarse-grained}} column in Tab.~\ref{tab:levels_results}). The relative performance improvements ranged from  $19.48\%$ (DINOv2 ViT-B) $-$ $93.51\%$ (ViT-L). Three models---SigLIP ViT-B, DINOv2 ViT-B, and ViT-L---performed even better than the teacher model which is the strongest baseline in that setting that exists at the time of writing this manuscript.

\noindent{\bf {\color{fine-blue}{Local fine-grained}}}. The responses of most base models' did not strongly correspond to human responses for fine-grained semantics either and, thus, model performances were far from the human reliabiltiy score. The human noise-ceiling was with $65.92\%$ similar to the noise-ceiling of the coarse-grained setting. Prior to fine-tuning, models achieved poor alignment scores of $40.75\%$ (ViT-L) $-$ $57.72\%$ (DINOv2 ViT-B), except for DINOv1 ViT-B whose base model performance ($62.92$\%) was significantly better than the performance of the other models (see Tab~\ref{tab:levels_results}).

Soft-alignment improved this mismatch to some degree but not as significantly as it did for the coarse-grained abstraction setting. AligNet models achieved odd-one-out accuracies of $58.93$ (ViT-S) $-$ $62.92\%$ (DINOv1 ViT-B) (see the {\color{fine-blue}{fine-grained}} column in Tab.~\ref{tab:levels_results}). Here, relative increases in performance ranged from $7.84\%$ (DINOv2 ViT-B) - $46.03\%$ (ViT-L). For a single model, DINOv1 ViT-B, the alignment score did not change after soft-alignment. Again, the best performing soft-aligned model, DINOv1 ViT-B, performed with an odd-one-out accuracy of $62.92\%$ better than any other model---whether linearly aligned using UD or fine-tuned via teacher distillation---in that setting. Note that UD and fine-tuning via distillation decreased the performance of DINOv1 ViT-B. This model performed closest to the human reliability score of $65.92\%$. Every soft-aligned model performed better than the teacher in that abstraction setting.

It is rather surprising that AligNet models were better aligned than their base models for this setting because the THINGS triplets reflect coarse-grained semantic structure and, hence, neither teacher nor student representations have ever explicitly learned anything about a fine-grained human object similarity space. We hypothesize that the fusion of information about human coarse-grained semantic and the teacher model's local similarity structure leads to a representation space that better reflects human object similarity structure in general---irrespective of the level of granularity.

\noindent{\bf {\color{boundary-red}{Class-boundary}}}. Supervised classifiers and image/text contrastive models performed close to the noise ceiling for class-boundary triplets prior to any fine-tuning. Their odd-one-out accuracies ranged from $81.96\%$ (SigLIP ViT-B) to $93.67\%$ (ViT-L) (see Tab.~\ref{tab:levels_results}). However, the caption generator model (CapPa) often responded differently from the human participants in that setting and achieved a significantly lower odd-one-out accuracy score compared to the other student models. It selected the same odd-one-out image as the human participants in $70.37\%$ of the triplets. 

AligNet fine-tuning changed the representation spaces of all models to be equally well aligned with the human respondents. The differences between AligNet models was significantly smaller compared to the other settings, with odd-one-out accuracies of $93.09\%-94.24\%$. Surprisingly, the odd-one-out accuracies of models fine-tuned on AligNet were higher than the human reliability score of $89.21\%$ (see \cref{fig:fig2}D, rightmost column) with the best model achieving an impressive alignment score of $94.24\%$ (ViT-L). This means that the responses of AligNet models were more similar to the average human responses---since each triplet response is the majority response of the subject population---than the level of agreement among the human subjects themselves. The performances of all models improved by fine-tuning them on AligNet. The relative increases in performance were between $0.62\% - 32.39\%$.

Simply performing \emph{distillation without alignment} did not improve the models' alignment with humans as significantly. For some models (e.g., ViT-L) simply distilling the similarity structure of the teacher model into the student without aligning the teacher's representations even decreased the degree of alignment with the human odd-one-out responses (see the {\color{boundary-red}{class-boundary}} column in Tab.~\ref{tab:levels_results}). Similarly, applying the UD transformation to a model's representation space sometimes slightly improved the human-model fit (e.g. ViT-B) and sometimes significantly decreased model performance (e.g. ViT-S) with large differences between models. Moreover, AligNet fine-tuning was the only transformation technique that could significantly decrease the variance in performance across the different student models. The difference between the worst (CapPa ViT-B) and the best (ViT-L) AligNet fine-tuned model was just $1.15\%$

\begin{table}[ht!]
\centering
\resizebox{1\textwidth}{!}{%
\begin{tabular}
{l|cccc|cccc|cccc}
\toprule
& \multicolumn{4}{c}{{\color{coarse-green}{Coarse-grained}}} & \multicolumn{4}{c}{{\color{fine-blue}{ Fine-grained}}} & \multicolumn{4}{c}{{\color{boundary-red}{Class-boundary}}} \\
Model $\setminus$ Fine-tuning & Original & Distillation & UD & AligNet & Original & Distillation & UD & AligNet & Original & Distillation & UD & AligNet \\
\midrule
ViT-S &  40.08\% & \textbf{58.82}\% & 46.29\% & 67.09\% & 51.30\% & 58.80\% & 43.41\% &  58.93\% &  92.23\% &  92.64\% & 85.31\% &  93.80\% \\ 
ViT-B &  36.10\% & 56.15\% & 40.96\% &  67.72\% & 46.04\% & 57.04\% & 42.66\% &  60.02\% &  88.73\% & 91.68\% & \textbf{93.31}\% & 93.99\% \\ 
ViT-L &  35.45\% & 57.75\% & 41.84\% &  \textbf{68.60}\%$^\dagger$ & 40.75\% & 57.47\% & 39.34\% &  59.51\% &  \textbf{93.67}\% & \textbf{93.18}\% & \textbf{93.31}\% &  \textbf{94.24}\%$^\dagger$ \\ 
CapPa-B ViT-B &  42.99\% & 53.55\% & 66.73\% &  66.06\% & 49.36\% & 57.15\% & 53.07\% &  60.21\% &  70.37\% & 90.85\% & 87.92\% &  93.09\% \\
DINOv1 ViT-B & 52.44\% & 55.86\% & 57.82\% &  65.70\% & \textbf{62.92}\%$^\dagger$ \ & \textbf{59.76}\% & \textbf{60.91}\% & \textbf{62.92}\%$^\dagger$ &   90.09\% & 92.39\% & 88.84\% &  94.03\% \\ 
DINOv2 ViT-B &  \textbf{57.38}\%  & 49.37\% & 66.09\% &  68.56\% & 57.72\% & 55.31\% & 55.72\% & 62.24\%&   89.74\% & 88.46\% & 90.13\% &  93.45\% \\ 
SigLIP ViT-B &  46.88\%  & 51.72\% & \textbf{68.17}\% &  68.47\% & 53.44\% & 57.39\% & 57.71\% & 60.94\% &  81.96\% & 90.81\% & 91.32\% & 93.75\% \\ 
\midrule
SigLIP So400m (Teacher) &  50.24\%  & \_ & 68.03\% &  \_  & 57.27\% & \_  & 58.86\% &  \_	&  90.42\% & \_ & 93.11\% &  \_ \\
\bottomrule
\end{tabular}%
}
\caption{Human alignment results for Levels. Here, we show triplet odd-one-out accuracies (in \%)---measured as the fraction of triplets for which models selected the same odd-one-out image as the majority of the human participants for the three levels of abstraction---\emph{{\color{coarse-green}{coarse-grained}}}, \emph{{\color{fine-blue}{fine-grained}}}, and \emph{{\color{boundary-red}{class-boundary}}}---in the Levels dataset that we collected via online crowdsourcing. Bold face indicates highest performance within a single column and $\dagger$ indicates best performance for a triplet type setting overall.}
\label{tab:levels_results}
\end{table}

{{
\subsubsection{Human alignment depends on the abstraction level}
\label{sec:alignment_abstraction}

We found that the best model before soft-alignment at a particular abstraction level was often not the best model after soft-alignment. While DINOv2 was the best aligned model before AligNet fine-tuning for the  fine-grained (57.38\%) and coarse-grained ({57.72\%) abstraction settings (see Tab.~\ref{tab:levels_results}), it remained the best aligned model only for the fine-grained setting (62.24\%) but did not improve as significantly as the other models for the coarse-grained (relative improvement of 18.78\%) and class-boundary settings (relative improvement of 4.13\%). On the other hand, ViT-L was the worst-aligned model before fine-tuning for the coarse-grained setting (35.45\%), but after AligNet fine-tuning, it became the best-aligned model, achieving the highest odd-one-out accuracy (68.60\%) across all models. The relative improvement in performance was 93.51\%. In addition, ViT-L was both the best aligned model before (93.67\%) and after (94.24\%) AligNet fine-tuning for the class-boundary setting (both scores are higher than the human reliability score of 89.21\%).

We observed a similar phenomenon for the coarse- and fine-grained human responses datasets of \citet{peterson2016, peterson2018evaluating}. DINOv2 showed the highest Spearman rank correlation coefficient with the human similarity judgments prior to any fine-tuning for both abstraction levels but remained the best aligned model only for the fine-grained datasets after fine-tuning (see Tab.~\ref{tab:rsa_results_peterson} for details). This suggests that DINOv2 fine-tuned on AligNet best captured fine-grained human semantic structure whereas image/text contrastive and supervised models best reflected coarse-grained and class-boundary semantic structure respectively after the fine-tuning process.

\subsubsection{ImageNet vs. Ecoset models on Levels}
\label{appx:ecoset_on_levels}

Here, we evaluate vision models trained on Ecoset~\cite{ecoset} on the Levels data. Ecoset is an ecologically motivated natural image dataset, designed in the hope to produce models that better reflect human perception than models trained on ImageNet. The number of (training) data points is the same between Ecoset and ImageNet (approx. 1.3M) but the number of classes is differs between the two datasets: 565 (Ecoset) vs. 1000 (ImageNet). We find that models trained on Ecoset are worse aligned with human similiarty judgments than the same set of models trained on ImageNet in most abstraction settings (see Tab.~\ref{tab:ecoset_results}). This is in line with previous findings of \citet{muttenthaler2023human} who found that Ecoset models are worse aligned than ImageNet models with the human similarity judgment datasets from \citet{hebart2020revealing}, \citet{king2019}, and \citet{cichy2019}. The reason for this is most likely the higher number of classes in ImageNet which has been found to play a crucial role for the alignment of neural network representations with human semantic cognition~\cite{muttenthaler2023human}. Note that Ecoset is designed for supervised learning, and weakly-/self-supervised image/text model representations generally correspond better to human behavior and brain data than the representations of supervised ImageNet models~\cite{konkle2022can, muttenthaler2023human}.

\begin{table}[ht!]
\small
\centering
\begin{tabular}
{l|cc|cc|cc}
\toprule
& \multicolumn{2}{c}{{\color{coarse-green}{Coarse-grained}}} & \multicolumn{2}{c}{{\color{fine-blue}{ Fine-grained}}} & \multicolumn{2}{c}{{\color{boundary-red}{Class-boundary}}} \\
Model $\setminus$ Data & Ecoset & ImageNet & Ecoset & ImageNet & Ecoset & ImageNet \\
\midrule
AlexNet &  42.83\% & 38.41\% & 44.39\% & 44.31\% & 55.47\% & 49.81\% \\
VGG-16 &  42.65\% & 40.18\% & 45.16\% & 49.95\% & 60.34\% & 64.58\% \\ 
ResNet-50 &  44.98\% & 46.46\% & 44.80\% & 49.98\% & 63.85\% & 91.90\% \\ 
Inception v3 &  45.86\% & 36.08\% & 49.59\% & 35.76\% & 70.38\% & 82.28\% \\
\bottomrule
\end{tabular}
\caption{{Human alignment results for Ecoset vs. ImageNet models on the Levels data. Here, we show triplet odd-one-out accuracies (in \%)---measured as the fraction of triplets for which models selected the same odd-one-out image as the majority of the human participants for the three levels of abstraction---\emph{{\color{coarse-green}{coarse-grained}}}, \emph{{\color{fine-blue}{fine-grained}}}, and \emph{{\color{boundary-red}{class-boundary}}}---in the Levels dataset that we collected via online crowdsourcing. Representations of Ecoset models were extracted with the Python toolbox \texttt{thingsvision}~\cite{muttenthaler2021thingsvision}.}}
\label{tab:ecoset_results}
\end{table}

\subsubsection{Evaluating vision-language models on Levels}
\label{appx:vlm_eval_levels}

Although we focus on vision models in the main text, a growing area of research focuses on Vision-Language Models (VLMs). As language captures much of the detail of human semantic knowledge, it is natural to ask how these models perform on the Levels dataset. We therefore evaluated Gemini 2.0 Flash \citep{team2023gemini}\finaledit{ and Gemini 2.5 Pro \citep{comanici2025gemini}} on Levels, using a prompt that approximately follows the human instructions, except edited to describe a single trial rather than a full experimental procedure:
``I will show you three object images presented in sequence. These images show different things. Your task is to pick one image that is least similar to the other two. If you do not recognize what is shown to you in one of the images, simply base your judgement on your best guess at what the image might show. Which of these images is least similar to the others? Please give your answer ("first", "second", or "third") before explaining.''

After this prompt, we presented the three images, and then evaluated the model's response. As expected, the models\finaledit{---and especially the stronger 2.5 Pro model---}perform fairly well at the class-boundary judgments, and moderately well at the others; they are comparable to \finaledit{or better than} the stronger vision models before alignment. However, \finaledit{neither VLM achieves the performance of AligNet,} and they are especially weak on the coarse-grained judgments. These results suggest that the improvements offered by our methods are complementary to the benefits offered by the richer language supervision in VLMs.}}}

\begin{table}[ht!]
\small 
\centering
\begin{tabular}
{l|c|c|c}
\toprule
Model $\setminus$ Triplet type & \multicolumn{1}{c}{{\color{coarse-green}{Coarse-grained}}} & \multicolumn{1}{c}{{\color{fine-blue}{ Fine-grained}}} & \multicolumn{1}{c}{{\color{boundary-red}{Class-boundary}}} \\
\midrule
Gemini 2.0 Flash &  55.0\% & 58.8\% & 89.2\% \\
Gemini 2.5 Pro &  61.3\% & 61.3\% & 89.9\% \\

\bottomrule
\end{tabular}
\caption{{Gemini 2.0 Flash and Gemini 2.5 Pro evaluation performance (odd-one-out accuracy) on the Levels dataset. The models perform well on most categories, comparable to the original versions of some of the stronger vision models---with the stronger model even outperforming them on coarse-grained triplets---but neither VLM achieves as high of performance as AligNet, especially on the coarse-grained triplets. (cf. \cref{tab:levels_results}.)}}
\label{tab:gemini_levels_eval}
\end{table}


\subsubsection{Correlation of model output uncertainties with human RTs}

\begin{figure}[ht!]
\centering
\includegraphics[width=1.0\textwidth]{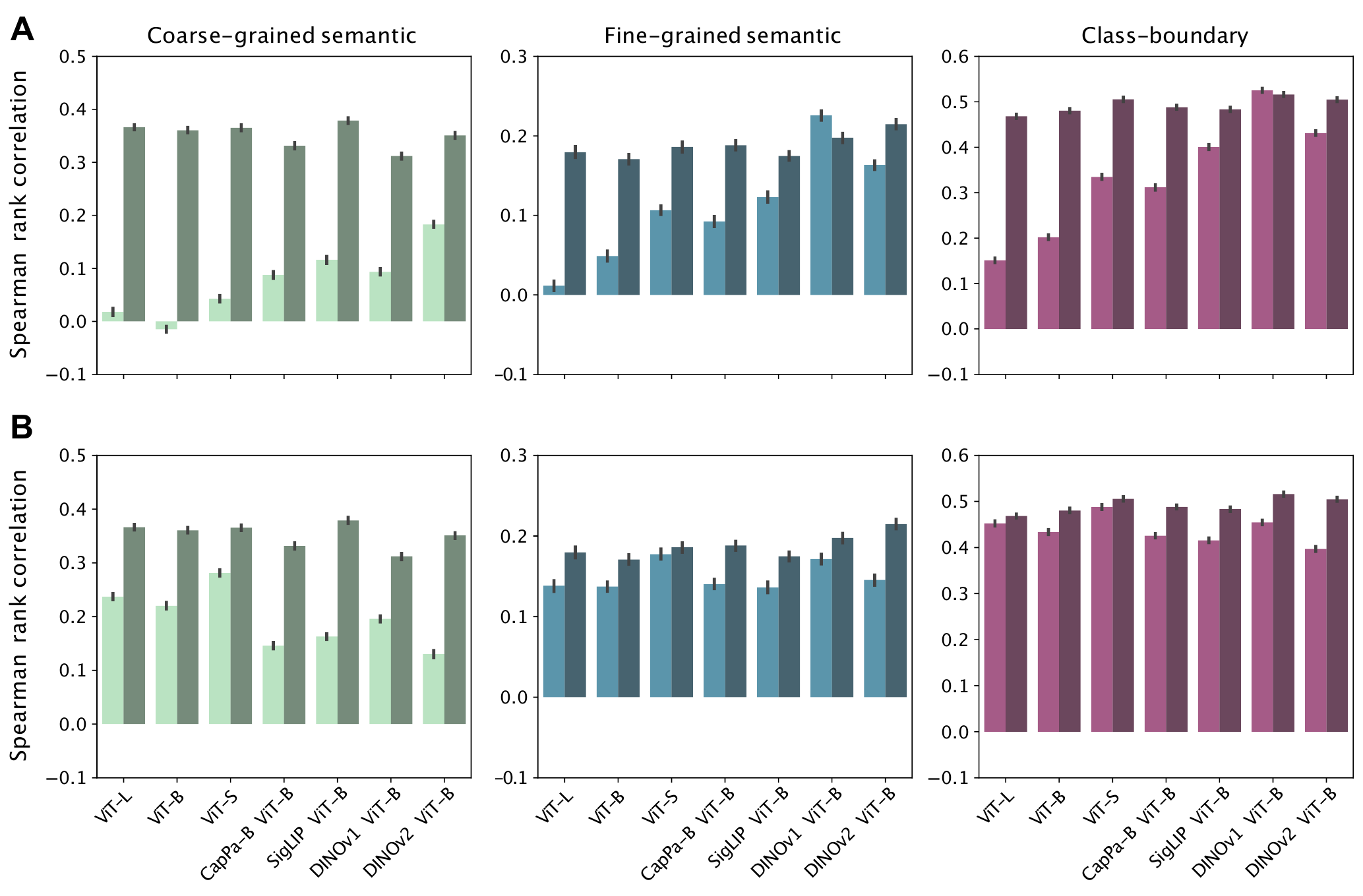}
\caption{Spearman rank correlation coefficients of model output uncertainties over the triplet odd-one-out choices (measured as discrete Shannon entropy) and the corresponding human response times (measured in log-space). The columns are partitioned into the three different triplet types: {\color{coarse-green}{coarse-grained}}, {\color{fine-blue}{fine-grained}}, and {\color{boundary-red}{class-boundary}}. \textbf{A}: AligNet fine-tuning (darker colors) in contrast to the base model representations (lighter colors). \textbf{B}: AligNet fine-tuning (darker colors) in contrast to distillation without alignment (lighter colors). Error bars reflect the standard deviation (SD) on the item level over $1000$ bootstraps.}
\label{fig:corrs_uncertainties_vs_rts}
\end{figure}

Here, we evaluate how the uncertainties of the model outputs---measured as the discrete Shannon entropy of the triplet probabilities---correlate with the human response times for the different triplets. For all base models, we observed a poor (close to zero) Spearman rank correlation between its output uncertainties over the triplet odd-one-out choices and the human response times for both the global coarse-grained and the local fine-grained semantic settings (see panel A in \cref{fig:corrs_uncertainties_vs_rts}). However, except for ViT-B and ViT-L, most base models showed a medium positive Spearman rank correlation ($\rho$=0.3-0.4) for the class-boundary setting (see rightmost column in panel A \cref{fig:corrs_uncertainties_vs_rts}). Note that the class-boundary setting is the easiest setting with the least variance/disagreement among the human participants (see Sec.~\ref{appx:levels} for details). AligNet-finetuning significantly improved the correspondence between model output uncertainties and human responses times for all models and triplet type settings. The improvements were most striking for the global coarse-grained semantic setting where every model achieved a medium to strong positive Spearman rank correlation of close to $\rho$=0.4. For the class-boundary setting, models fine-tuned on AligNet even achieved a strong positive Spearman rank correlation coefficient of close to $\rho=0.5$.

\begin{figure}[ht!]
\centering
\includegraphics[width=1.0\textwidth]{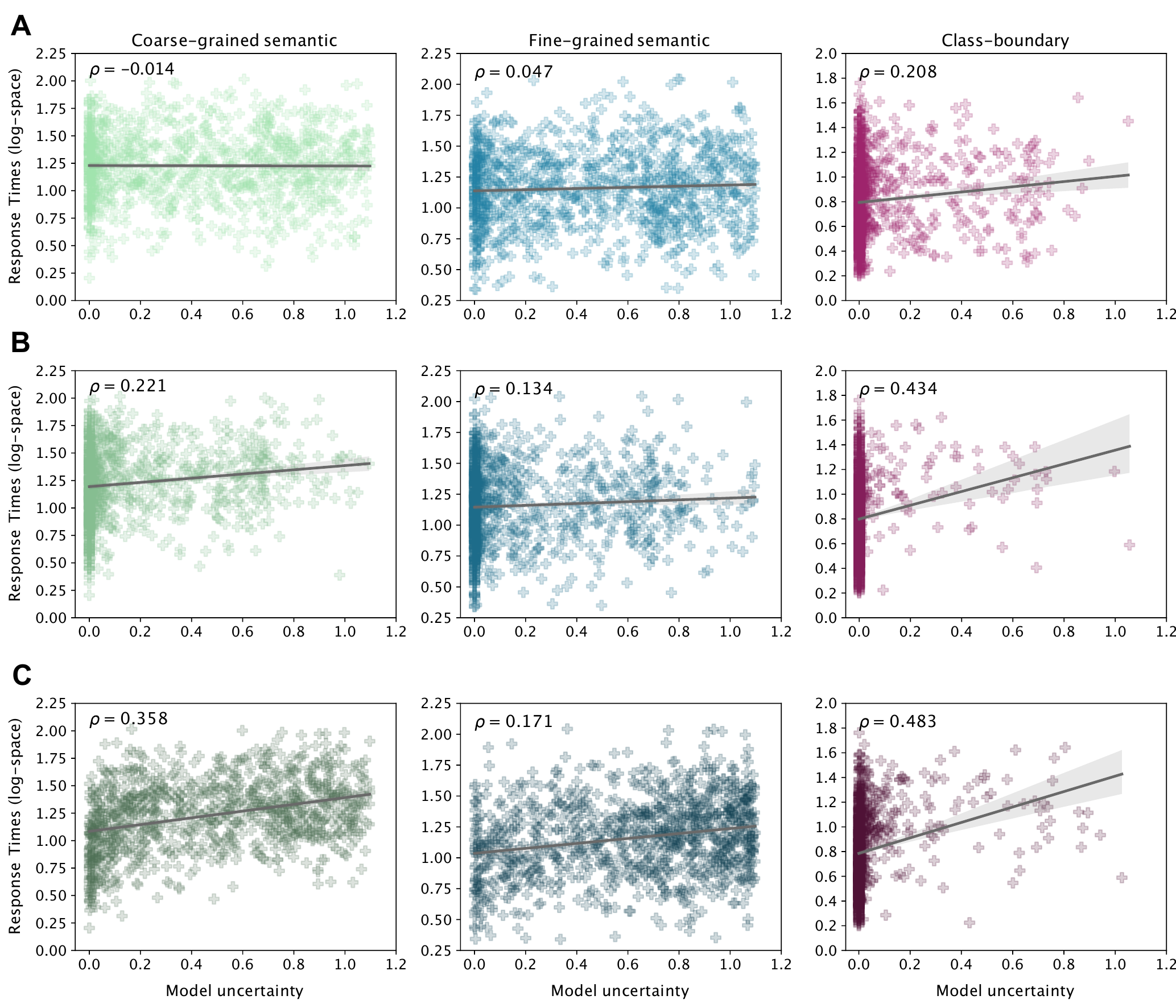}
\caption{Human response times (in $\log$-space) as a function of ViT-B's output uncertainty over the triplet odd-one-out choices (measured as the discrete Shannon entropy). The columns are partitioned into the three different triplet types: {\color{coarse-green}{coarse-grained}}, {\color{fine-blue}{fine-grained}}, and {\color{boundary-red}{class-boundary}}. \textbf{A}: Model uncertainties of \emph{original} representations. \textbf{B}: Model uncertainties of \emph{distilled} representations. \textbf{C}: Model uncertainties of \emph{AligNet-finetuned} representations. Correlation coefficients are Spearman rank correlations.}
\label{fig:corrs_uncertainties_vs_rts_vitb}
\end{figure}

\begin{figure}[ht!]
\centering
\includegraphics[width=1.0\textwidth]{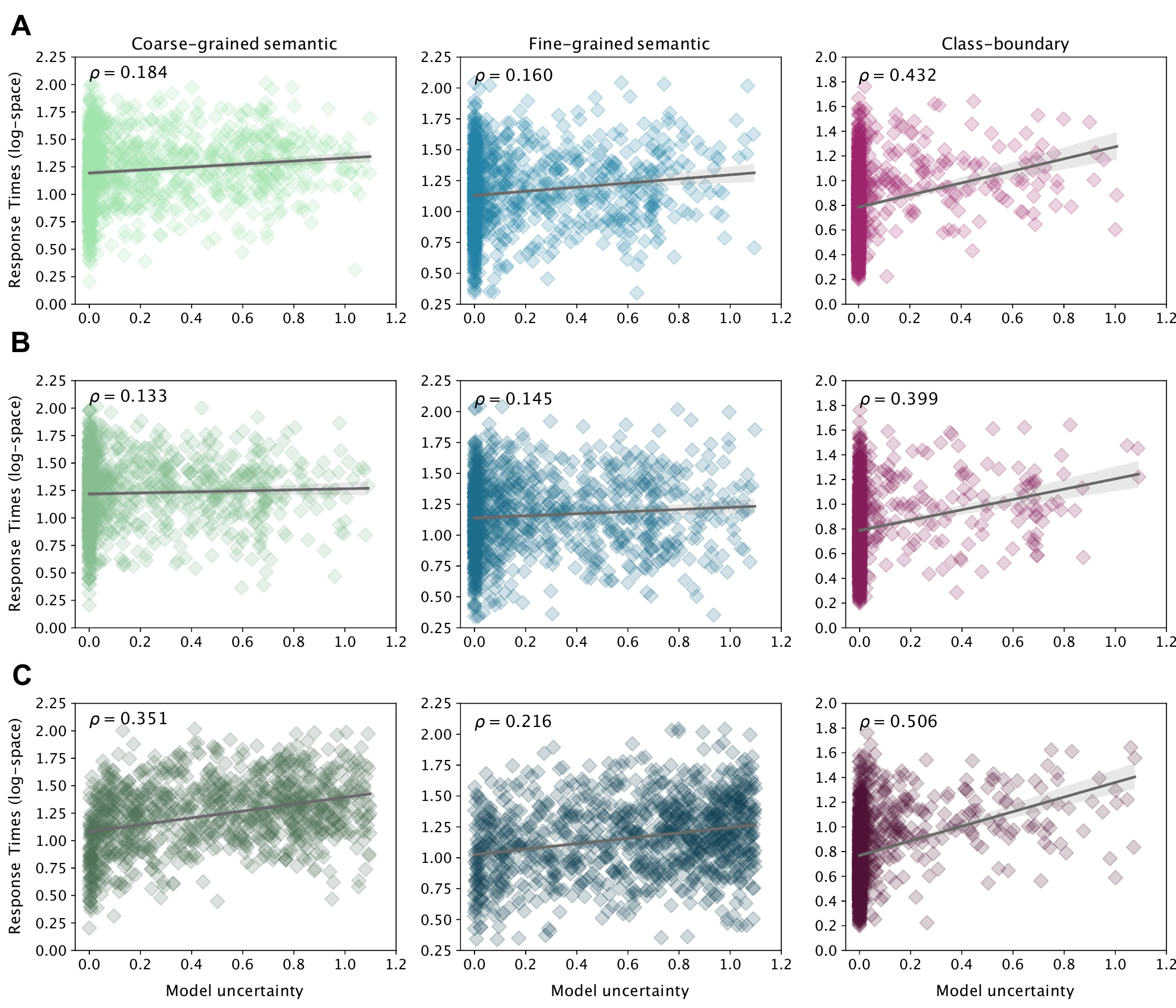}
\caption{Human response times (in $\log$-space) as a function of DINOv2's output uncertainty over the triplet odd-one-out choices (measured as the discrete Shannon entropy). The columns are partitioned into the three different triplet types: {\color{coarse-green}{coarse-grained}}, {\color{fine-blue}{fine-grained}}, and {\color{boundary-red}{class-boundary}}. \textbf{A}: Model uncertainties of \emph{original} representations. \textbf{B}: Model uncertainties of \emph{distilled} representations. \textbf{C}: Model uncertainties of \emph{AligNet-finetuned} representations. Correlation coefficients are Spearman rank correlations.}
\label{fig:corrs_uncertainties_vs_rts_dinov2}
\end{figure}

\begin{figure}[ht!]
\centering
\includegraphics[width=1.0\textwidth]{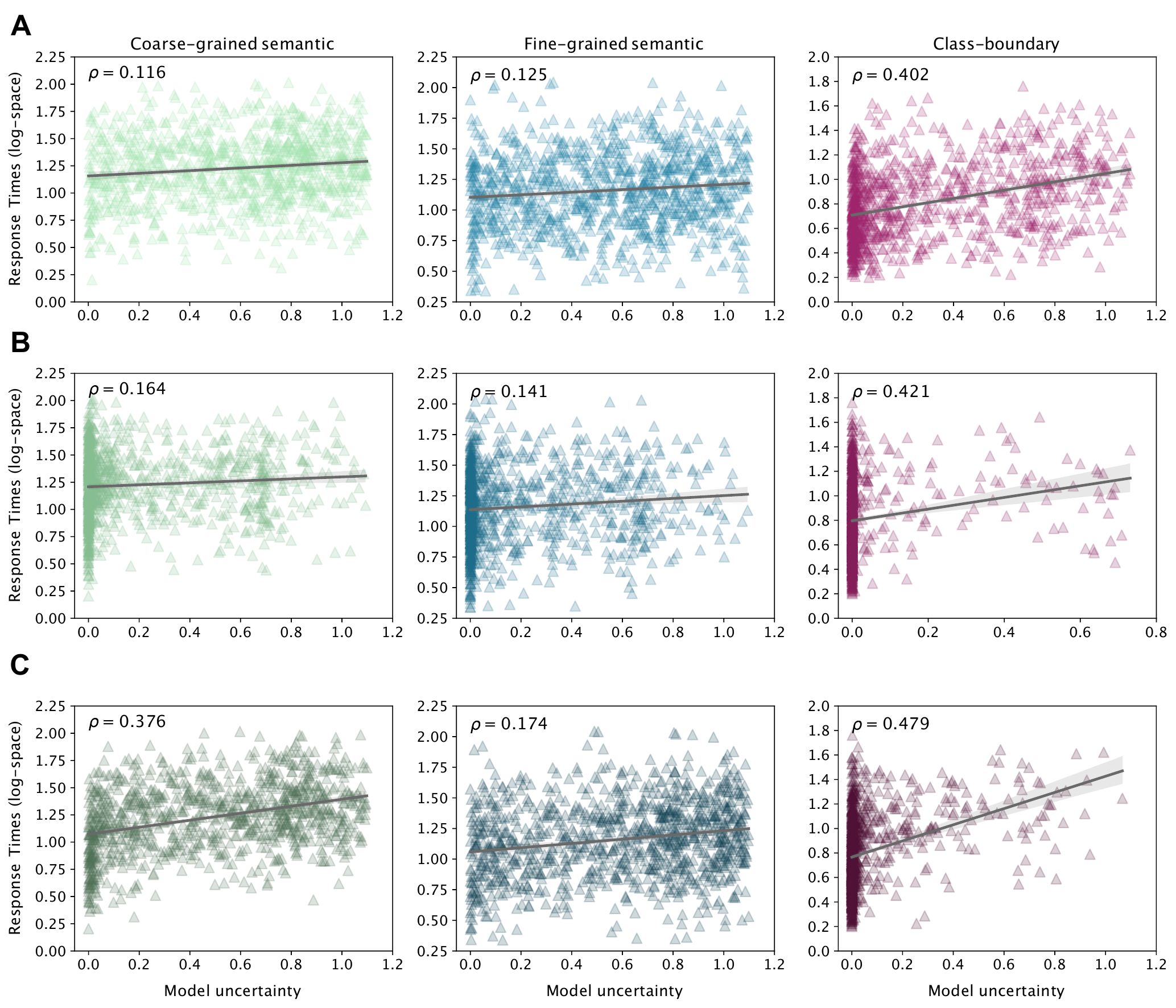}
\caption{Human response times (in $\log$-space) as a function of SigLIP ViT-B's output uncertainty over the triplet odd-one-out choices (measured as the discrete Shannon entropy). The columns are partitioned into the three different triplet types: {\color{coarse-green}{coarse-grained}}, {\color{fine-blue}{fine-grained}}, and {\color{boundary-red}{class-boundary}}. \textbf{A}: Model uncertainties of \emph{original} representations. \textbf{B}: Model uncertainties of \emph{distilled} representations. \textbf{C}: Model uncertainties of \emph{AligNet-finetuned} representations. Correlation coefficients are Spearman rank correlations.}
\label{fig:corrs_uncertainties_vs_rts_siglip_vitb}
\end{figure}

In panel B of \cref{fig:corrs_uncertainties_vs_rts}, we compare AligNet fine-tuned representations (darker colors) against performing distillation without alignment (lighter colors). We can see that simply performing distillation without alignment could improve the correspondence of model uncertainties and human response times over the base representations but the improvement was not as substantial compared to applying the full AligNet framework. We find that for every model and every triplet setting, AligNet fine-tuning yielded a significantly stronger Spearman rank correlation coefficient. In \cref{fig:corrs_uncertainties_vs_rts_vitb}, \cref{fig:corrs_uncertainties_vs_rts_dinov2}, and \cref{fig:corrs_uncertainties_vs_rts_siglip_vitb} we contrast the human response times (in $\log$-space) against the model's output uncertainties over the triplet odd-one-out responses for ViT-B, DINOv2 ViT-B, and SigLIP ViT-B respectively for each triplet individually. We chose those three models because they reflect a representative subset of the student models (see Sec.~\ref{appx:student_models} that we fine-tuned on AligNet. Each of those models has the same backbone but was pretrained on a different datasets with a different objective function. We observe that AligNet fine-tuning significantly improved the correspondence of the models' output uncertainties to the human responses times for all three models and triplet settings. Interestingly, distillation without alignment, decreased the Spearman rank correlation coefficient of DINOv2 for each of the three triplet settings (see panel B in \cref{fig:corrs_uncertainties_vs_rts_dinov2}). That was not the case for ViT-B and SigLIP ViT-B (see panel B in \cref{fig:corrs_uncertainties_vs_rts_vitb} and \cref{fig:corrs_uncertainties_vs_rts_siglip_vitb} respectively).

Taken together, we find that AligNet fine-tuning significantly improved the correspondence of models' output uncertainties with the human response times irrespective of the task and objective with which a model was (pre-)trained and regardless of the specific triplet type. The improvements were most striking for global coarse-grained semantic.

\subsection{Machine Learning}
\label{appx:machine_learning}

In this section we provide further details and evidence about the machine learning downstream task performances of our method. We start by demonstrating few-shot learning and out-of-distribution detection evaluations and end the section with showing various ablations.

\subsubsection{Few-shot learning}
\label{appx:few_shot}

Few-shot learning~\cite{wang2021generalizing} is a common way to evaluate how well neural network representations generalize to new tasks: Here, the goal is to learn a new task/dataset from only a few labelled examples (typically $\leq$10 examples per labelled class are used to train a classifier on top of the pretrained representations), and then evaluate its performance on the full test set for that task. Good few-shot performance is indicative of models that have a solid understanding of many different concepts, and that their representations are able to quickly integrate/expand to new or related concepts.

We run few-shot learning evaluations on eight different supervised learning tasks. ImageNet\,\citep{russakovsky2014imagenet}, Places365\,\citep{zhou2018places} and Cifar100\,\citep{krizhevsky2009learning} test for broad understanding of a wide variety of visual concepts, and gauge how easily a representation can adapt to a new learning task on concepts a (pretrained) model has likely encountered before.
We also test datasets that instead require that the model is able to differentiate between very fine-grained differences in related concepts. Concretely, we test  datasets such as Caltech-UCSD Birds 200 (CUB-200) Birds\,\citep{welinder2010caltech-ucsd}, Stanford Cars\,\citep{krause2013collecting}, Flowers\,\citep{nilsback2008automated}, Oxford-IIIT Pets\,\citep{parkhi2012cats} and the UC Merced Land Use Dataset\,\citep{yang2010bag-of-visual-words}. 
For our evaluation, we freeze the model weights, and train a linear classifier on top of the extracted image representation\,\citep{alain2016understanding}.
On all datasets, we test two few-shot settings: the extreme case where only a single example per class is provided (`1-shot`) and the more conservative, but still challenging `10-shot` challenge. 

\begin{figure}
    \centering
    \includegraphics[width=\textwidth]{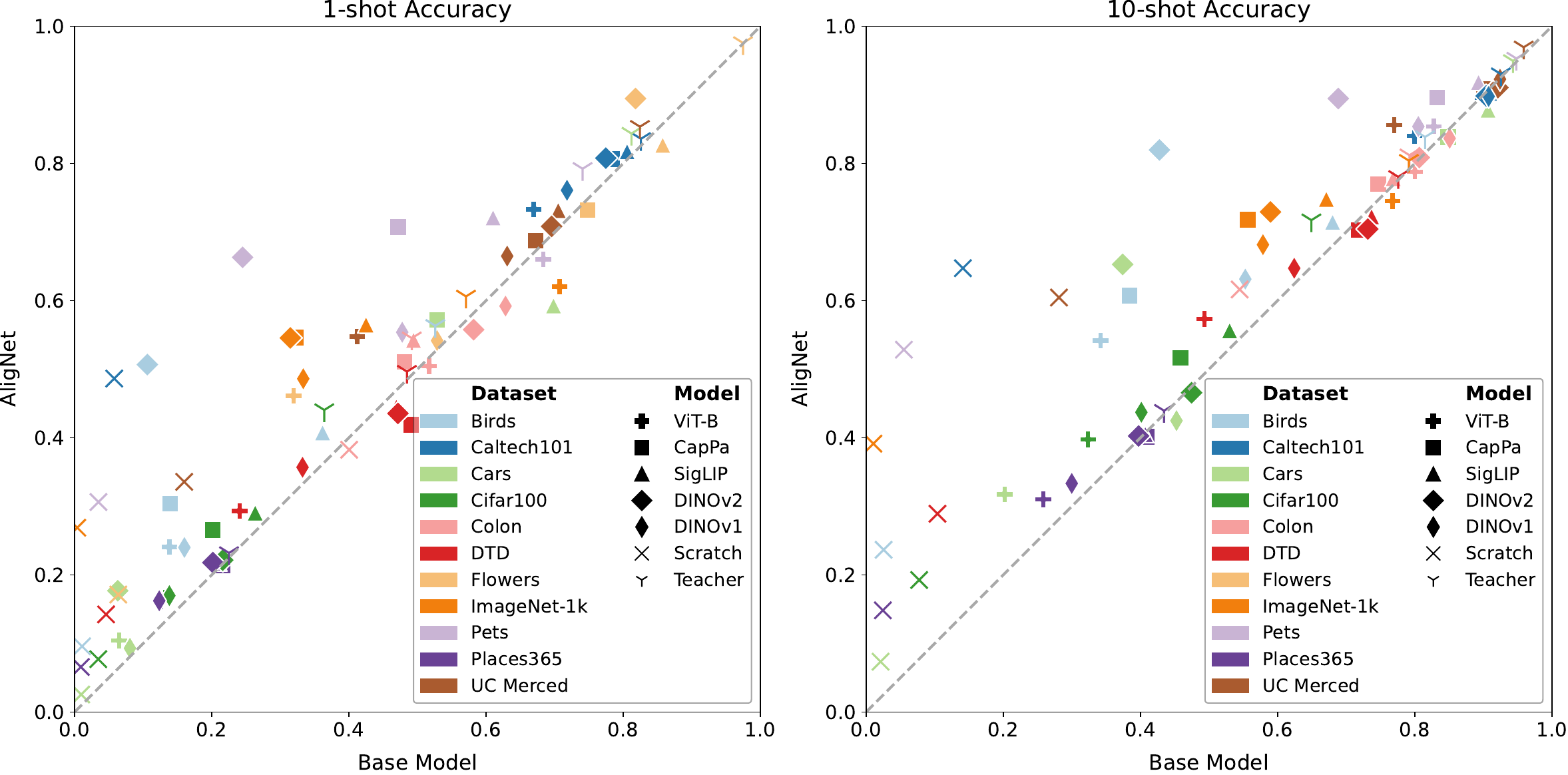}
    \caption{Results for 1-shot and 10-shot linear probing on a number of common few-shot evaluation datasets.}
    \label{fig:fewshot}
\end{figure}

In \cref{fig:fewshot}, we compare the few-shot performance of pre-trained models vs. their few-shot performance after being aligned-finetuned.
AligNet finetuning is clearly beneficial on most datasets in both 1-shot and 10-shot evaluations. 
Even models that have previously been trained on ImageNet data (e.g. DINOv1 ViT-B/16 and supervised ViT-B/16) show improvements on 1-shot ImageNet performance, which indicates that the benefits do not merely come from being exposed to ImageNet data, but also from the label information we distilled into AligNet. 

\paragraph{Tip-Adapter.} \citet{muttenthaler2023improving} have demonstrated that linearly transforming the pretrained representations of state-of-the-art image/text models into a human-aligned concept space can improve upon methods that were specifically designed for improving the performance of pretrained representations in few-shot learning tasks. One of these methods is Tip-Adapter~\cite{zhang2022tip}. Tip-Adapter is designed to yield effective few-shot classifiers for image/text models by linearly combining the output of two modules --- a \emph{zero-shot classifier} and a \emph{key-value cache model}. UD (see SI.~\ref{appx:method-uncertainty-distill}) and gLocal~\cite{muttenthaler2023improving} have highly similar learning objectives: both linearly transform the representations of pretrained vision foundation models into a (global) human-aligned concept space while preserving the models' nearest neighbor structure. Since we find that UD either performs equally well or better in matching the human similarity judgments and predicting the human RTs for the triplets in the Levels data (see SI.~\ref{appx:ablations}) and \emph{soft-aligned} models outperform the UD transformation across the different human object similarity tasks that we evaluated (see SI.~\ref{appx:human_alignment} for details), we can be confident that \emph{soft-aligned} image/text models (i.e., image/text models fine-tuned on AligNet) can improve upon Tip-Adapter. For CIFAR-100, the improvements in zero-shot performance of using gLocal in conjunction with Tip-Adapter compated to Tip-Adapter alone ranged from $6.29\%$ for a CLIP-RN50 up to $18.72\%$ for a CLIP ViT-L/14 (see Appendix D in \citet{muttenthaler2023improving}). For other datasets, the improvements in zero-shot performance were similarly high. One caveat with Tip-Adapter is that its application is restricted to models that have a text encoder module, i.e., image/text models. Soft-alignment, on the other hand, does not suffer from the same constraint and can be applied to any vision model.

\subsubsection{Out-of-distribution}
\label{appx:ood}

\begin{figure}
    \centering
    \includegraphics[width=\textwidth]{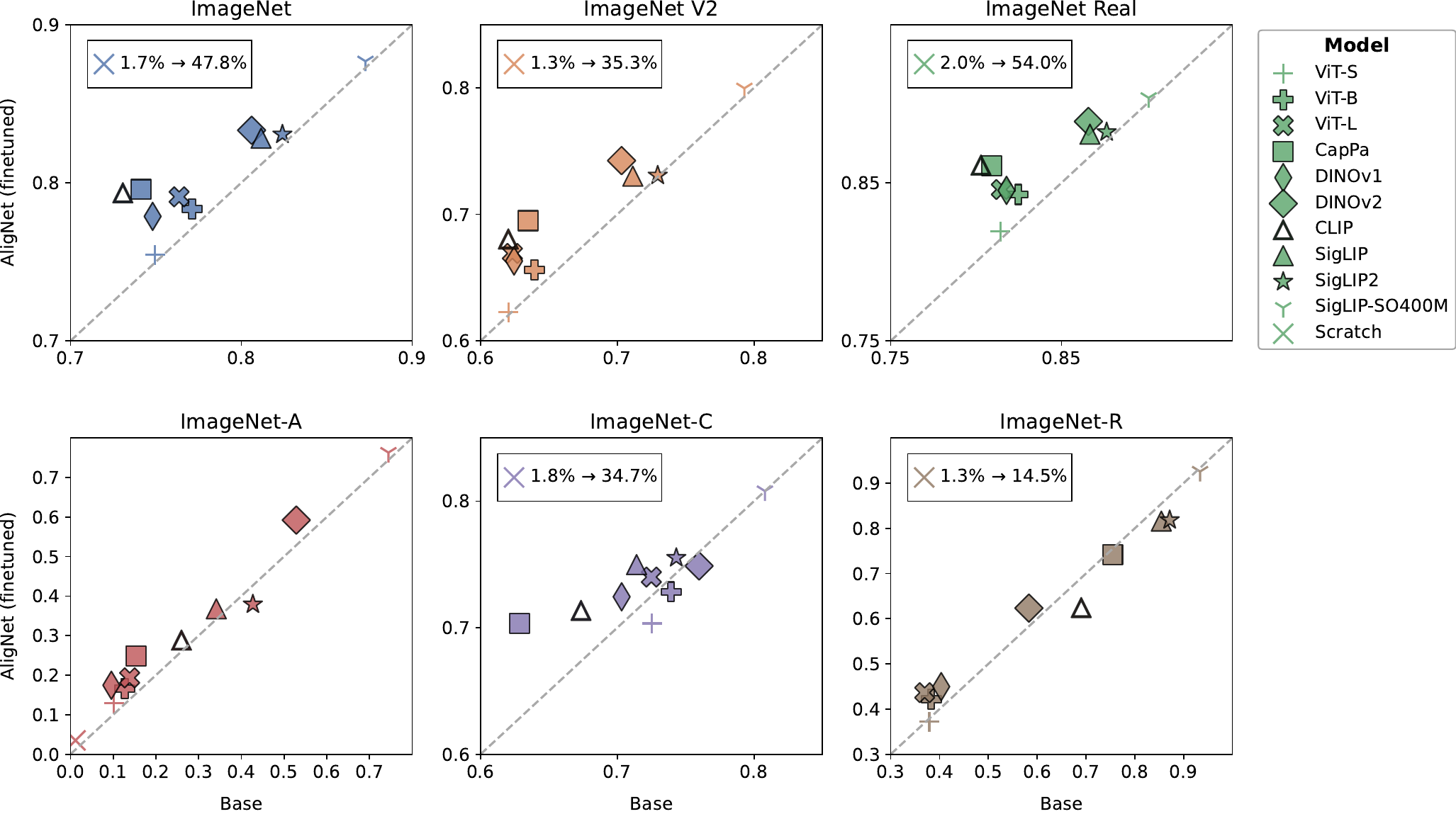}
    \caption{{
Out of distribution results on several popular variations of ImageNet.
    Accuracy of a linear readout head trained on top of the (frozen) base models is plotted on the $x$-axis, against the corresponding performance a linear readout head trained on top of the (frozen) AligNet-finetuned models on the $y$-axis.
    Every symbol above the diagonal indicates improved performance (through AligNet), while symbols below the diagonal indicate a performance degradation.
    Note that for better readability the range of the plot is zoomed in on the pre-trained models.
    The model trained from scratch ($\times$) is thus often out of bounds due to its substantially lower performance, and its accuracy before and after finetuning is instead written out in a small inset.
    }}
    \label{fig:imagenet_ood}
\end{figure}

In this section we investigate the how robustly a neural network has learned visual concepts: how well can they recognize objects under different, more challenging settings?
This is an important question for many applications, and gives an idea both of the generalization properties, as well as to the practical applicability of models.
To investigate robustness, we look at six different out-of-distribution testsets for the popular ImageNet dataset in \cref{fig:imagenet_ood}.
ImageNet ReAL\,\citep{beyer2020done} relabels the dataset to fix inaccuracies/oversights in the original data. ImageNet V2\,\citep{recht2019imagenet} is a new, updated test-set for the original dataset.
ImageNet-A\,\citep{hendrycks2021natural} consists of adversarial examples, whose aim is to confuse/trick modern systems.
ImageNet-C\,\citep{hendrycks2019benchmarking} tests how well models can deal with corrupted inputs, e.g. very noisy or blurry images.
Finally, ImageNet-R\,\citep{hendrycks2021faces} shows more abstract renditions of actual objects, for example scetches, sculptures or drawings.

As we see in the results, AligNet finetuning has a positive impact on the models under investigation, with most datasets showing slight improvements. On average over all models and across all datasets, accuracy went up by 2.6 percentage points, which on Imagenet-like datasets is considered a large performance boost\,\citep{beyer2020done}. As can be expected, the models that were pre-trained on imagenet already showed the smallest improvements. Even on ImageNet ReAL, which can be considered a more reliable measure of model improvement than the original Imagenet, on average the models improved by 1.5\%. 
The clearest advantage of AligNet can be seen for adversarial examples where AligNet-finetuned models exhibit much better behavior than their baseline counterparts: the average improvement was a jump of 5.6 percentage points.
For corruptions and renditions, the situation is mixed: while on average the results still show an improvement of 1.8\% on ImageNet C and 1.4\% on ImageNet R, the SigLIP models slightly deteriorated.
Overall, this fits with our understanding of what AligNet is meant to achieve: it allows models to learn clearer concepts that generalize better, but it is not improve performance in situations where image corruption or artistic renditions are the cause of misclassifications.

{\color{black}{In their comprehensive survey on the robustness of computer vision models, \citet{liu2024comprehensive} compare state-of-the-art training methods for improving the robustness of Vision Transformers (``ViTs’’), namely Adversarial Training (``AT’’), Masked Autoencoder (``MAE’’), and Contrastive Language-Image Pretraining (``CLIP’’). These methods have been shown to substantially improve robustness upon vanilla supervised or pure image self-supervised training. Similarly to AT, our method can be employed in addition to the other four methods evaluated in that survey. In our work we align SSL models similar to MAE, such as DINO and DINOv2, and models pretrained using CLIP (we use SigLIP which is a variant of CLIP). We find that soft-alignment performs comparably to AT (e.g., ImageNet-A, ImageNet-R). On average, it is neither significantly worse nor significantly better than AT and substantially improves upon the non-aligned base model representations (see Tab.~\ref{tab:ood_robustness_comparison}). Since ViT-B/16 is the current defacto standard architecture in computer vision, we use it as the basis for all of our soft-aligned models and compare against the numbers for ViT-B reported in \citet{liu2024comprehensive} (see Tab.~\ref{tab:ood_robustness_comparison}). We note that while our method is not primarily targeted toward optimizing OOD robustness (or has an adversarial training component), improved OOD robustness appears to be a useful side-effect that emerges from transforming the model representations into a more human-aligned and interpretable space via our method.}}

\begin{table}[ht!]
\scriptsize
\centering
\begin{tabular}
{l|l|cccc|cccc|c}
\toprule
& & \multicolumn{4}{c}{\textbf{IID}} & \multicolumn{4}{c}{\textbf{Real-world OOD}} & \multicolumn{1}{c}{\textbf{Synthesized OOD}}  \\
\textbf{Backbone} & \textbf{(Pre-)Training}  & IN-Val & IN-21K & IN-V2 & IN-Real & ON & IN-A & IN-R & IN-V & IN-C \\
\midrule
ViT-B/16 & Vanilla & 75.7 & \textbf{91.7} & 61.6 & 80.9 & 20.8 & 11.3 & 32.8 & 24.2 & 34.3 \\
 & MAE & \textbf{83.6} & 90.5 & 73.1 & \textbf{88.1} & 37.4 & 37.4 & 49.8 & 36 & 49.4 \\
 & CLIP & 68.4 & 69.8 & 61.9 & 75.1 & \textbf{45} & \textbf{50} & 77.7 & \textbf{39.4} & 29.3 \\
 & AT & 73.4 & 86.7 & 60.4 & 80.6 & 19.2 & 8.8 & 50.7 & 20.2 & 36.6 \\
\midrule
 & Vanilla + AligNet & 77.6 & --- & 65.4 & 83.6 & --- & 15.2 & 40.2 & --- & 72.3 \\
 & SigLIP + AligNet & 82.8 & --- & \textbf{73.3} & 87 & --- & 37 & \textbf{81.4} & --- & \textbf{74.8} \\
\bottomrule
\end{tabular}
\caption{Comparison of state-of-the-art (pre-)training methods for improving downstream task performance of computer vision models~\citep[cf.][]{liu2024comprehensive} and soft-aligned models (i.e., models fine-tuned on AligNet) for various ImageNet-based evaluation datasets, testing a model's performance either on i.i.d. data or its OOD robustness. Boldface indicates the best performance for a dataset.}
\label{tab:ood_robustness_comparison}
\end{table}

{{

\subsubsection{Distribution shift}
\label{appx:breeds_detailed_results}

\begin{figure}[ht!]
    \centering
    \includegraphics[width=\textwidth]{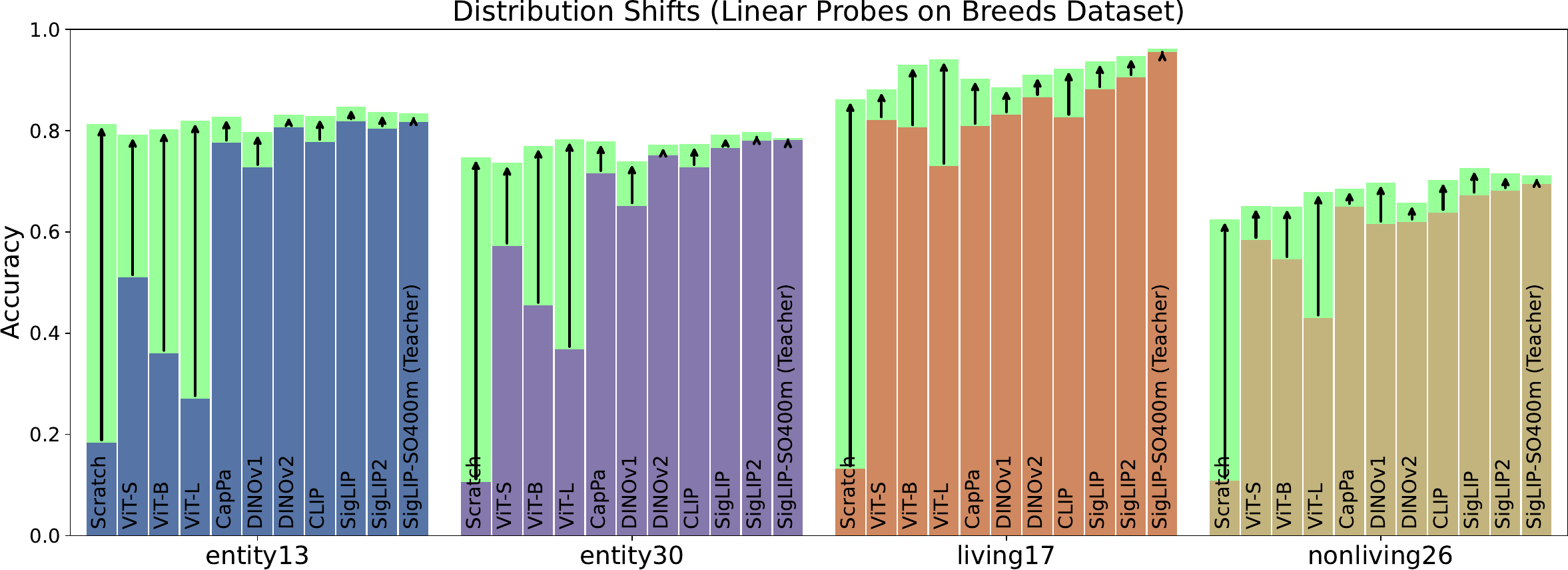}
    \caption{{Additional results for all student models on the four Breeds datasets: ``entity13'', ``entity30'', ``living17'', and ``nonliving26''. Student models are sorted according to their objective function (supervised, captioning, self-supervised, image/text). The randomly initialized ViT-B has the poorest base performance but benefits the most (leftmost column in each panel). Conversely, SigLIP-So400m---the teacher model---has the strongest base performance and benefits the least but consistently (rightmost column in each panel). Vertical arrows indicate improvements in performance compared to the base performance.}}
    \label{fig:breeds_full}
\end{figure}

To measure whether the global category structure induced by fine-tuning on AligNet helps to alleviate model problems with distribution shifts between the training and the test set, we evaluate our models on the BREEDS benchmarks~\cite{santurkar2020breeds}. These were specifically designed to test generalization under input distribution shifts, by constructing datasets where the samples in the training and test sets are sampled from different subpopulations. For example, the training set for the ``dog'' class contains some selected dog breeds in the training set (e.g., Dalmatians and Schnautzers), whereas a disjoint set of dog breeds is used for the test set (e.g.. Poodles and German Shepards).

In this section we expand on the results presented in Sec.~\ref{sec:ml_results} in the main text and evaluate all of our AligNet fine-tuned student models (see Sec.~\ref{appx:student_models} for details) on BREEDS. For each model, we train a linear probe on the four BREEDS training sets---``entity13'', ``entity30'', ``living17'', and ``nonliving26''---, and evaluate that probe on the corresponding test sets.

AligNet fine-tuning consistently improves performance for all model types across all of the BREEDs benchmarks. We find that the poorer the baseline performance of a student model is, the more that model benefits from fine-tuning on AligNet. For example, the baseline performance of the randomly initialized ViT-B model (which we refer to as ``Scratch'') is by far the worst but after AligNet fine-tuning that model performs as well as the pretrained student models for ``entity13'' and ``entity30''. For the ``living17'' and ``nonliving26'' datasets its performance is slightly worse (but much less than before fine-tuning) than that of the other models, which may be due to the missing pretraining step. Conversely, SigLIP-So400m---the teacher model---has the strongest base performance and thus benefits the least from finetuning. It is striking, however, that even the teacher model consistently improves on BREEDS although that is the model that we used to generate AligNet. For ``entity30'' some of the student models even outperform the teacher model. This is suggestive evidence that the human part of the AligNet data plays a larger role for these kind of analyses than the teacher model part. 

Another interesting observation is that larger supervised ViT models have worse base performance than smaller supervised ViT models but benefit notably more from AligNet fine-tuning than the smaller models. Before AligNet fine-tuning the order of the models according to their base performance is ViT-S, ViT-B, ViT-L across all four datasets. This order is reversed after AligNet fine-tuning for all four datasets. Thus, similarly to the human alignment results presented in \cref{fig:rsa_scatter_vits}, there seems to exist a small scaling effect where larger models benefit more from AligNet fine-tuning, at least for the models that are not the teacher itself. We show all model results for the BREEDS dataset in \cref{fig:breeds_full}.

}}

{

\subsubsection{Additional Experimental Results}

\paragraph{CLIP.}
\label{appx:clip}

\begin{figure}[ht!]
    \centering
    \includegraphics[width=\textwidth]{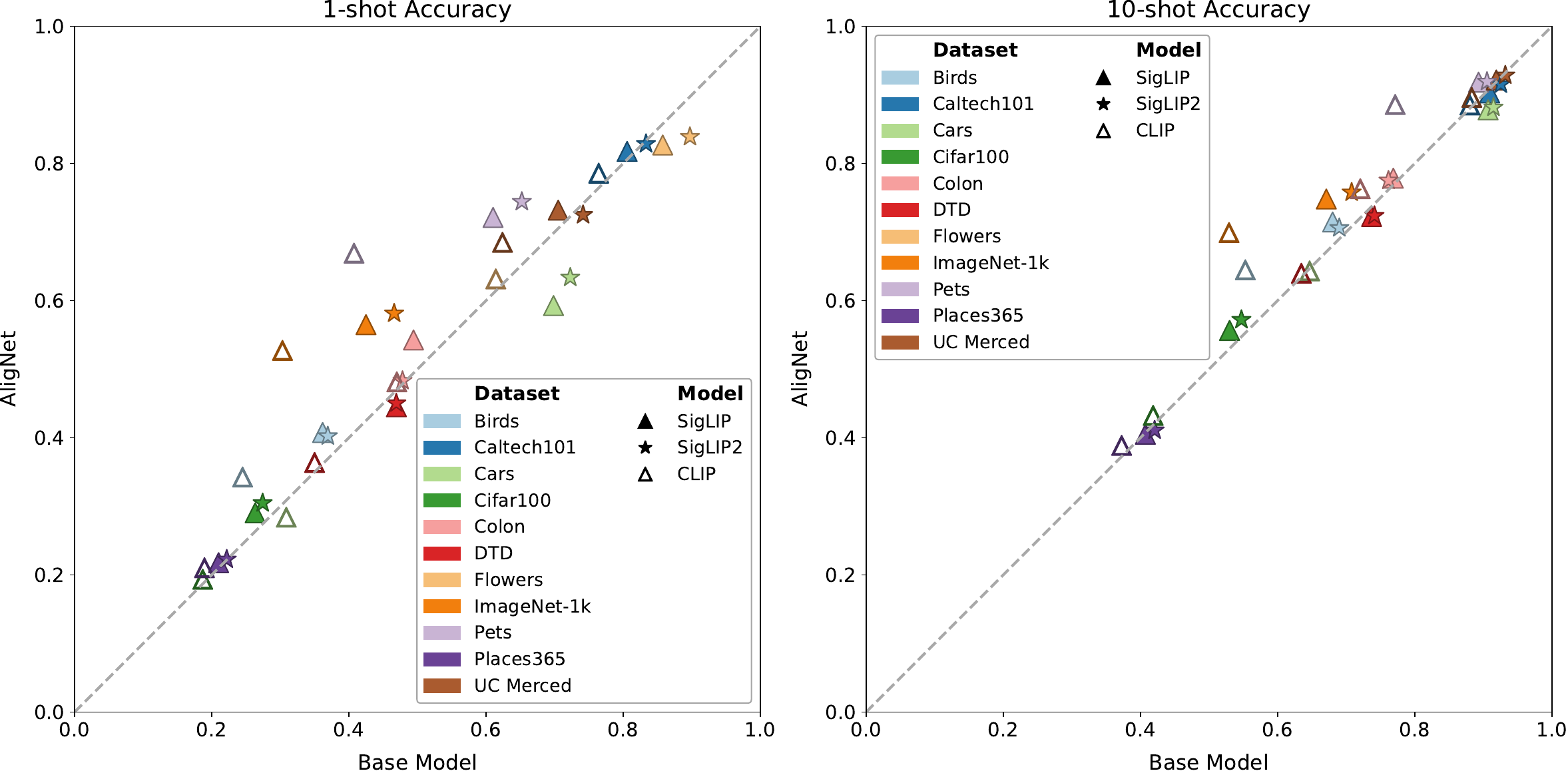}
    \caption{{Results for 1-shot and 10-shot linear probing on a number of common few-shot evaluation datasets for CLIP, SigLIP, and SigLIP2.}}
    \label{fig:fewshot_clip}
\end{figure}

In the main paper we have focused on SigLIP~\cite{zhai2023siglip} as the representative model for image/text contrastive models because it is a very strong model that significantly improved upon the more widely known CLIP~\cite{radford2021learning} model.
Here we show results for a CLIP ViT-B student model that we finetuned on AligNet using the same setup and hyperparameters as we did for the SigLIP model.
As can be seen in \cref{fig:fewshot_clip}, for few-shot learning the CLIP model generally benefits more from AligNet finetuning than SigLIP, while its final performance remains below that of the SigLIP model.
The same pattern, can be seen for ImageNet and out-of-distribution performance (see \cref{fig:imagenet_ood}), and distribution shifts (see \cref{fig:breeds_full}).
This further confirms the benefits of our method, and also justifies the focus on SigLIP in the main paper.

\paragraph{SigLIP2}
\label{appx:siglip2}
The opposite pattern emerges when comparing SigLIP with the more recent SigLIP2~\cite{tschannen2025siglip}: SigLIP2 has even stronger baseline performance than SigLIP and unsurprisingly benefits less from finetuning on AligNet.
As can be seen in \cref{fig:fewshot_clip} and \cref{fig:breeds_full}, it still does benefit in most cases for fewshot-generalization and distribution shift cases.
For the out-of-distribution experiments in \cref{fig:imagenet_ood}, on the other hand, the effect is negligible or even detrimental.
Given the strong baseline performance of SigLIP2 it would be fruitful to use the largest SigLIP2 as an even stronger teacher model, which we leave for future work.
}

{

\subsubsection{Training on AligNet from Scratch}
\label{appx:scratch}

AligNet finetuning provides consistent improvements across many different models, settings, and datasets. This begs the question whether pretraining is actually needed, or if we can train a randomly initialized model solely on AligNet and achieve similar results. To answer this question, we trained a ViT-B/16 architecture from scratch on the AligNet dataset using the KLD triplet objective outlined in \cref{eq:alignet_objective}.
We optimize for 1M steps with the Adam optimizer using a cosine decay learning rate schedule with a peak learning rate of $0.0003$, $\beta_1=0.9$, $\beta_2=0.999$, weight decay of $0.1$,  $\tau^{\dagger}=1000$, and a batch size of $1024$.
These hyperparameters were determined via grid search. 

The results are shown in \cref{fig:fewshot,fig:imagenet_ood,fig:breeds_full,fig:ud_ablation}, with the model listed as ``Scratch''.  They roughly fall into two clusters: The first one concerns the distribution-shift experiments on the Breeds dataset (see \cref{fig:breeds_full}), and odd-one-out accuracy on the THINGS dataset (see \cref{fig:ud_ablation}a).
Here our Scratch model performs almost as well as the other pretraining + finetuning models. This means that these tasks are quite closely related to the benefits that the AligNet fine-tuning provides.

The seconds cluster comprises the few-shot (see \cref{fig:fewshot}) and ImageNet out-of-distribution results (see \cref{fig:imagenet_ood}).
While the Scratch model improves substantially over training, on these tasks its performance clearly falls far short of that of the other pretrained models.
For example it achieves approx. $27\%$ 1-shot accuracy on ImageNet compared to the $56\%$ accuracy of the same-sized SigLIP ViT-B model.
Pretraining on other vision tasks is thus still important to augment AligNet training and clearly instills important complementary capabilities in a neural network model.
}

{

\subsubsection{Perceptual patch similarity evaluations} \label{appx:bapps}
\begin{figure}[ht!]
    \centering
    \includegraphics[width=0.66\textwidth]{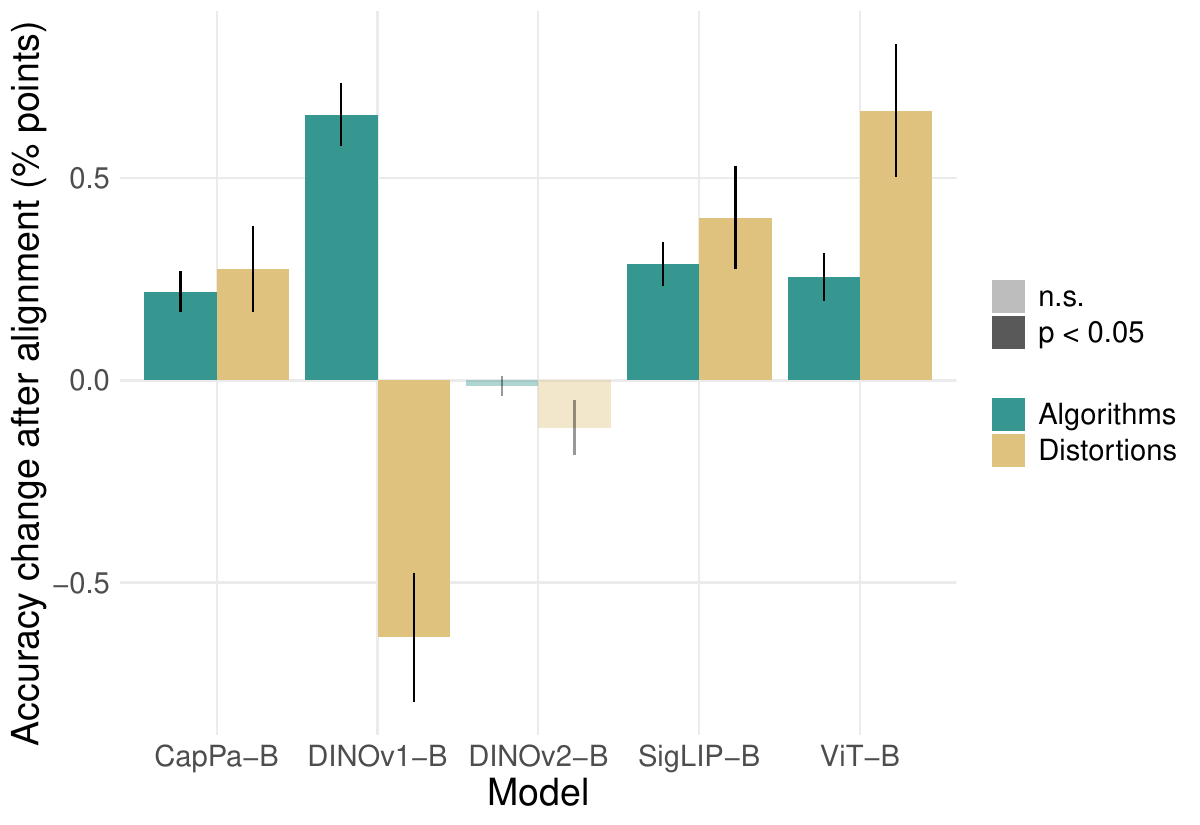}
    \caption{ Performance change on the BAPPS dataset after alignment. All changes are relatively small in absolute magnitude. Most of the AligNet models improve across both subsets of the benchmark, with the exception of the DINOv1 model, which shows significant improvements on Algorithms, but worsens on Distortions, and DINOv2, which changes nonsignificantly. These results suggest that AligNet tuning may slightly improve perceptual judgment performance in most cases. (Colors denote split subsets from the original work; transparent bars are non-signficant changes.)}
    \label{fig:bapps}
\end{figure}

\begin{table}[ht!]
\scriptsize 
\centering
\begin{tabular}
{ll|cc|cccc}
\toprule
& & \multicolumn{2}{c}{\textbf{Distortions}} & \multicolumn{4}{c}{\textbf{Algorithms}}\\
\multicolumn{2}{c}{\textbf{Model}} & cnn & traditional & color & deblur & frameinterp & superres  \\
\midrule
CapPa-B & base & 84.62 & 59.26 & 63.62 & 60.83 & 54.87 & 65.33\\
CapPa-B & AligNet & 84.96 & 59.47 & 64.22 & 60.93 & 54.98 & 65.40\\
DINOv1-B & base & 85.23 & 60.28 & 63.79 & 60.96 & 55.08 & 65.47\\
DINOv1-B & AligNet & 84.87 & 59.36 & 65.59 & 60.97 & 55.72 & 65.65\\
DINOv2-B & base & 82.84 & 61.36 & 63.79 & 59.75 & 54.03 & 63.46\\
DINOv2-B & AligNet & 82.71  & 61.25 & 63.83 & 59.65 & 54.03 & 63.46\\
SigLIP-B & base & 84.34 & 59.39 & 62.78 & 61.01 & 54.18 & 65.10\\
SigLIP-B & AligNet & 84.79 & 59.75 & 63.69 & 60.95 & 54.45 & 65.13\\
ViT-B & base & 84.92 & 59.26 & 63.37 & 61.05 & 56.09 & 65.40\\
ViT-B & AligNet & 85.17 & 60.34 & 64.64 & 61.07 & 55.83 & 65.40\\
\bottomrule
\end{tabular}
\caption{ Evaluations of models on the BAPP dataset, before and after AligNet tuning.}
\label{tab:bapps_results}
\end{table}

We also evaluated our AligNet models on the BAPPS Berkeley-Adobe Perceptual Patch Similarity (BAPPS) dataset \citep{Zhang_2018_CVPR}. This benchmark consists of human similarity judgments, but focuses on a different level of analysis---perceptual similarity of small image patches under distortion. Accordingly, we follow the original work in analyzing representations aggregated across the early layers of the models (specficially, we used the the first 5 transformer layers, analogous to the 5 convolutional layers used by the original authors). Analyses on this task therefore present a different test of how our alignment procedure affects model capacities---including the effect at earlier layers.

We show the changes between the base and AligNet models in \cref{fig:bapps} and the absolute results in \cref{tab:bapps_results}. In general, the effects are very small---presumably both because this type of perceptual similarity is relatively different from the semantic structure we focused on, and because our alignment procedure is only explicitly targeted at representations much later in the model. Nevertheless, we generally see significant improvements (via exact binomial tests on the proportion of changed answers where AligNet improves over the base model) for the models across both subsets of the dataset. The exception is the DINO models --- DINOv1 gets significantly worse at the Distortions subset, though it still improves significantly on Algorithms, while DINOv2 changes nonsignificantly for both. We find these results to be overall promising---they suggest that AligNet tuning is not dramatically harming other types of perceptual judgment performance, and in fact may slightly improve it for most models.
}

\subsubsection{Ablations}
\label{appx:ablations}

We performed several ablations to examine how the effects of AligNet change if we modify some of its central parameters.

\paragraph{Model Capacity.} To investigate if AligNet finetuning affects models of different capacity/size in a similar way, we compared the performance of the ViT-B model (87M parameters) with the performances of a smaller ViT-S (22M parameters) and a larger ViT-L (305M parameters) model respectively. The results in  \cref{fig:fewshot} and \cref{fig:imagenet_ood} indicate that the positive impact of AligNet fine-tuning increases with model scale. For example, the average 10-shot performance improved by $2.4\%$ points for ViT-S, by $6.3\%$ points for ViT-B and by $8.3\%$ points for ViT-L. We observe similar results for all datasets we considered. Thus, we hypothesize that a larger model capacity makes it easier for a model to integrate the additional global coarse-grained information provided by AligNet's human-like similarity judgments.
\begin{figure}[ht!]
    \centering
    \includegraphics[width=\textwidth]{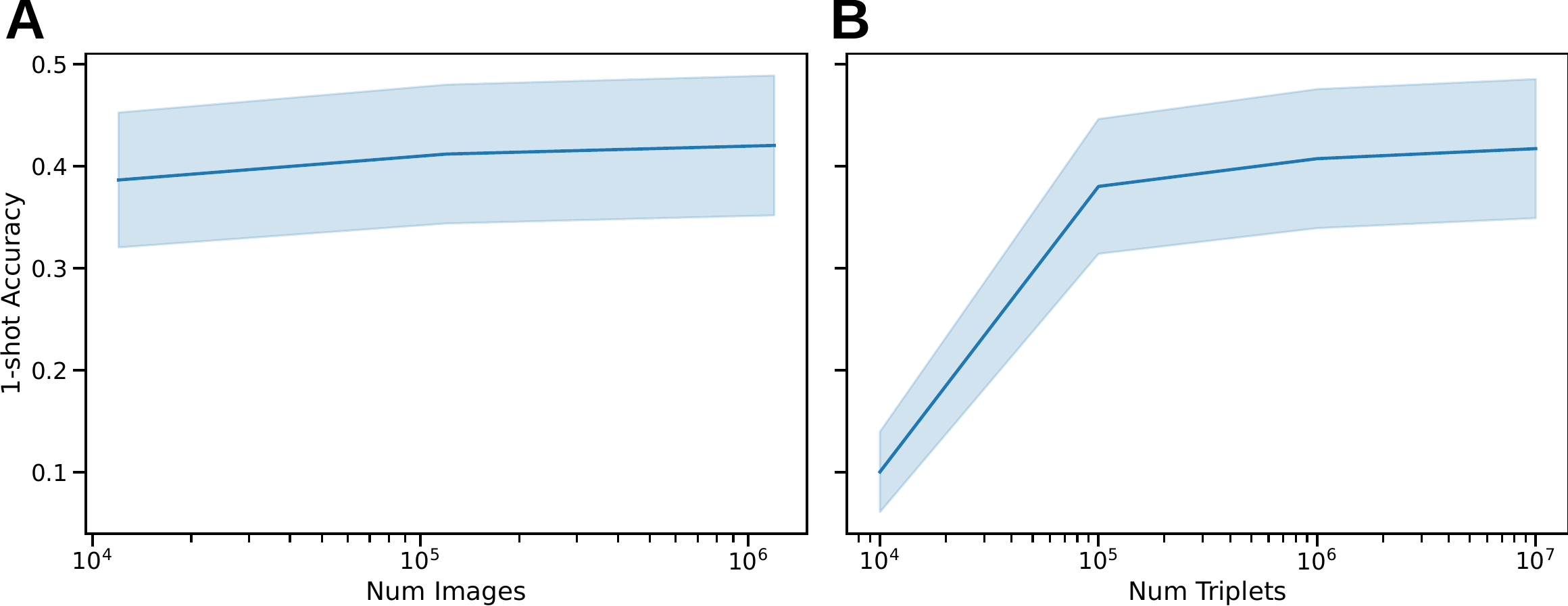}
    \caption{Accuracy in 1-shot linear probing over a number of common few-shot evaluation datasets when scaling \textbf{A} The number of images or \textbf{B} the number of triplets. Error bars show the standard error when averaging over datasets.}
    \label{fig:scalingablation}
\end{figure}

\paragraph{Increasing the Number of Images or Triplets.}
We also wanted to see how AligNet scales with the amount of available training data. We individually varied the number of ImageNet images we used to construct our dataset over 3 orders of magnitude, and the number of triplets that we sampled across 4 orders of magnitude. We then looked at the 1-shot results across all the few-shot datasets described in \cref{appx:few_shot} (Outcomes for 10-shot were not qualitatively different). The results in \cref{fig:scalingablation}a show that AligNet performance only increases slightly as the number of available base images increases. However, as we can see from  \cref{fig:scalingablation}b, the number of generated triplets does influence the outcome, though we hit diminishing returns eventually. We do not expect to be able to increase AligNet performance much further be simply increasing the amount of available data.

\begin{figure}
    \centering
    \includegraphics[width=\textwidth]{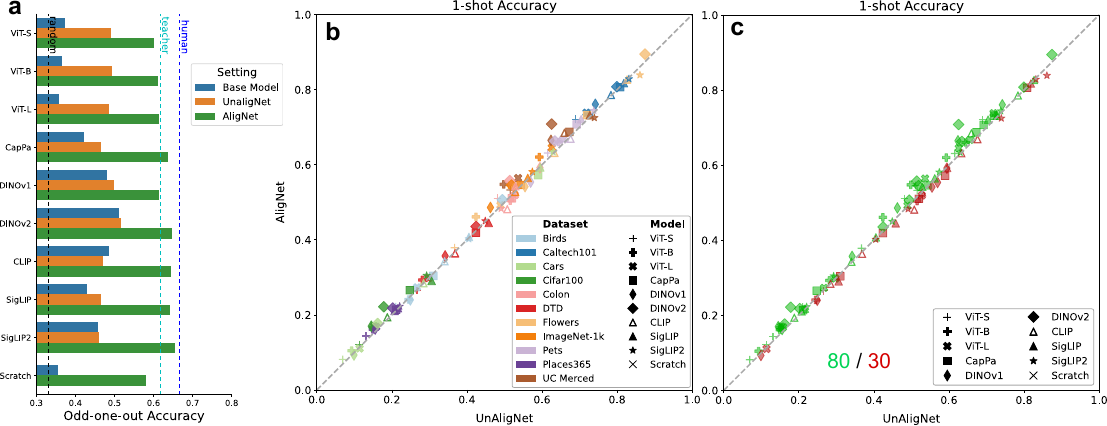}
    \caption{\textbf{a:} Odd-one-out accuracies on the THINGS dataset for different models in different settings.
    This figure is similar to (an extended version of) the THINGS plot in \cref{fig:fig2}A.
    ``AligNet'' and ``Base Model'' refer to the pretrained student model, without and without finetuning on AligNet respectively. 
    UnAlignet refers to a student finetuned on a variant of the AligNet dataset with an unaligned teacher.
    \textbf{b:} Comparing 1-shot accuracy on 11 datasets after fine-tuning on UnaligNet ($x$-axis) with fine-tuning on AligNet ($y$-axis).
    Most points lie slightly above the diagonal indicating that alignment of the teacher leads to a small but significant improvement in 1-shot performance of the student.
    This ablation shows the importance of aligning the teacher model for getting aligned student models.
    \textbf{c:} A variation on \textbf{b} which colors each marker as green if it represents an improvement from AligNet finetuning, vs red for a degradation of performance. Note that 80 out of the 110 model/data combinations led to an improvement.
    }
    \label{fig:ud_ablation}
\end{figure}

\paragraph{Using an Unaligned Teacher Model.}
An important part of the AligNet pipeline is to align the teacher model to human similarity judgments using a specially trained affine transformation.
Here we investigate the effect of this step, by generating a variant of AligNet without this transformation that we call UnAligNet.
For this dataset we do not use any human odd-one-out data, so finetuning on UnAligNet is equivalent to distilling the teacher model using our triplet sampling procedure.
\cref{fig:ud_ablation}A compares the odd-one-out accuracy of various student models when fine-tuning on AligNet vs on UnaligNet.
As expected, we find that using an unaligned teacher leads to substantially worse alignment of the student models. 
Though it should be noted that some models (especially the ViTs trained on ImageNet) can already profit even from an unaligned teacher.

It is perhaps unsurprising that an aligned teacher improves the alignment of the student models.
A more interesting question is, whether an aligned teacher also improves the students performance on other tasks such as few-shot classification. 
\cref{fig:ud_ablation}B thus compares 1-shot performance after training on UnaligNet ($x$-axis) with performance after training on AligNet ($y$-axis).
The first thing to note here, is that the differences between AligNet and UnaligNet are much smaller than the differences between AligNet and the Base Model (cf. \cref{fig:fewshot}).
This shows that the few-shot improvements are in large part due to the teacher model and our distillation method.
But importantly, the vast majority of remaining improvements are positive, which means that using an aligned teacher gives a consistent advantage over an unaligned teacher.

We confirm the significance of this trend by fitting a repeated measurements ANOVA to the accuracy on the different datasets, and treating the choice of model and the fine-tuning method (AligNet vs UnaligNet) as conditions.
Based on this ANOVA, the null-hypothesis that the fine-tuning do not affect the 1-shot accuracy, can be rejected at the $p < 0.01$ level. 

\begin{figure}
    \centering
    \includegraphics[width=\textwidth]{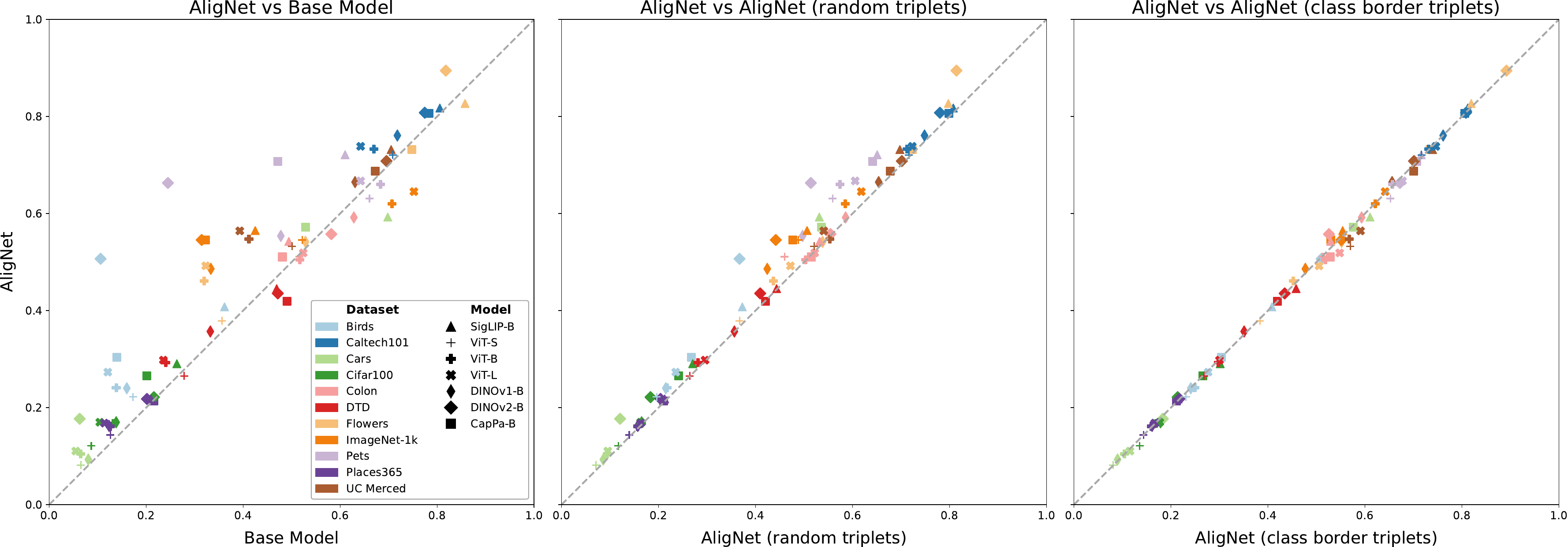}
    \caption{1-shot accuracy on various image datasets, comparing performance of an AligNet trained model ($y$-axis) with three other reference models.
    \textbf{Left:} Again comparing base-model ($x$-axis) with AligNet ($y$-axis) for reference.
    \textbf{Middle:} Comparing students fine-tuned using random triplets ($x$-axis) with AligNet ($y$-axis).
    \textbf{Right:} Comparing students fine-tuned using class-border triplets ($x$-axis) with AligNet ($y$-axis).}
    \label{fig:triplet_sampling}
\end{figure}

\paragraph{Investigating different Triplet Sampling Schemes.}

Sampling triplets is an important part of the AligNet pipeline, and the choice of triplets can greatly influence the effect and efficacy of the student finetuning.
Here we compare three different sampling strategies: 
\begin{itemize}
    \item \textbf{Cluster-border sampling}: This is the strategy used for AligNet. It is unsupervised and uses a k-means clustering of the teacher representations to sample triplets with two nearby images (same cluster) and one far away image (different cluster).
    \item \textbf{Class-border sampling}: This strategy is analogous to cluster-border sampling, but instead of clusters it uses the ImageNet labels to sample two nearby images (same class) and one far away image (different class).
    \item \textbf{Random sampling}: As a control we also compare to fully random sampling of triplets, i.e. each image is chosen with equal probability from the full set of all images.
\end{itemize}

\cref{fig:triplet_sampling} shows the 1-shot accuracy of student models when comparing different sampling strategies.
In all three panels, the $y$-axis corresponds to the performance of a model after AligNet-fine-tuning.
The left panel, again shows how this compares to the performance of the (not fine-tuned) base model ($x$-axis).
The middle panel compares shows how AligNet performance compares to fine-tuning on random-triplets ($x$-axis).
Finally, the right panel shows comparison between AligNet and fine-tuning on triplets sampled from class-borders.
Notice that the difference in the middle panel between random triplet sampling is consistently and substantially worse than cluster-based sampling.
This clearly highlights the importance of the triplet sampling for down-stream performance.
The panel on the right, on the other hand, shows that there is virtually no difference between cluster-border sampling and class-border sampling.
So when label information is available, that can effectively be used for sampling triplets, but even if there is not, the clustering based sampling will work just as well.

\begin{figure}
    \centering
    \includegraphics[width=0.48\textwidth]{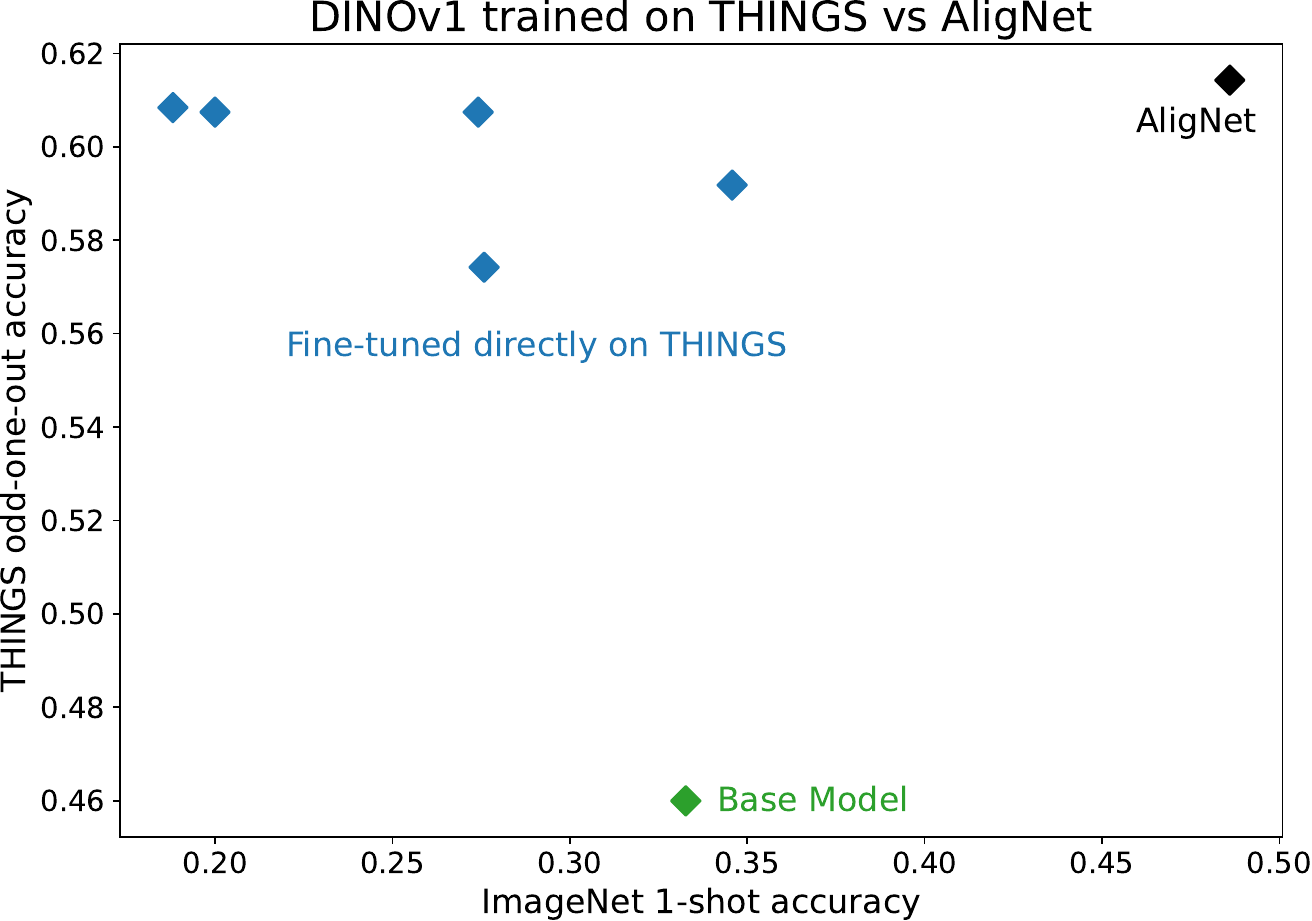}
    \includegraphics[width=0.48\textwidth]{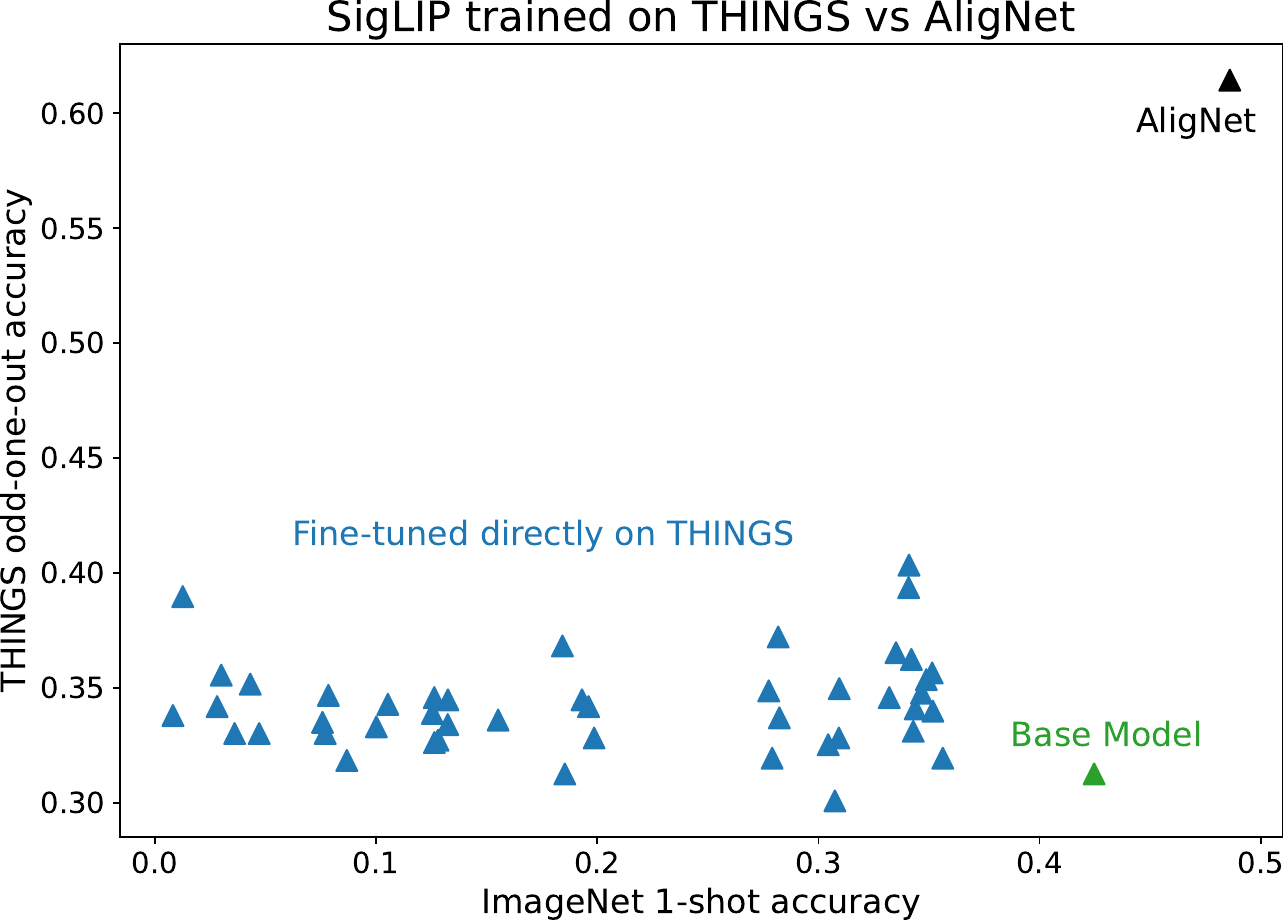}
    \caption{Performance on two different tasks: ImageNet 1-shot accuracy ($x$-axis) and odd-one-out accuracy on THINGS ($y$-axis) for two models (DINOv1 on the left, SigLIP-B on the right). 
    The green markers correspond to the base model, and the black marker to the model fine-tuned on AligNet. 
    The different blue markers in the scatterplot correspond to different hyperparameter settings (learning rate, weight decay, \dots) for fine-tuning directly on THINGS. }
    \label{fig:things_direct}
\end{figure}

\paragraph{Fine-tuning directly on the THINGS data.}
We have stressed that fine-tuning models to better align them with human similarity judgments depends on having sufficiently large datasets, which do not exist yet. 
This argument has motivated our multi-stage approach of first linearly aligning a teacher model and then using this as a surrogate model to generate a large corpus of human-like triplet odd-one-out choices.
Yet, if we look at \cref{fig:scalingablation}, it seems that decreasing the number of images or the number of triplets by up to two orders of magnitude still retains most of the benefit on downstream task performance (in terms of few-shot accuracy). This raises the question whether our pipeline as a whole might be unnecessary, given that the THINGS dataset~\cite{hebart2023things-data} already contains 4.70 million human similarity judgments collected via online crowd- sourcing for 1854 object images.

Therefore, here we compare soft-alignment with using the THINGS odd-one-out triplets directly for fine-tuning student models. \cref{fig:things_direct} shows that fine-tuning directly on THINGS is consistently worse compared to training on AligNet in terms of ImageNet 1-shot accuracy ($x$-axis).
Importantly, it appears that in almost all cases the performance of the model actually decreases below that of the base model. It is further interesting to note that even in terms of THINGS odd-one-out accuracy, the AligNet trained models perform better than the any other model.
This is somewhat surprising, because it means that there clearly exists an overfitting problem when training on the THINGS data (even with strong $\ell_{2}$-regularization), which seems to mostly disappear when using AligNet for fine-tuning. One reason for this could be the differences in the sampling procedure (cluster-based vs. random sampling) and the loss function (soft-choices vs. hard-choices) between AligNet and THINGS.

\begin{figure}[ht!]
    \centering
    \includegraphics[width=1.0\textwidth]{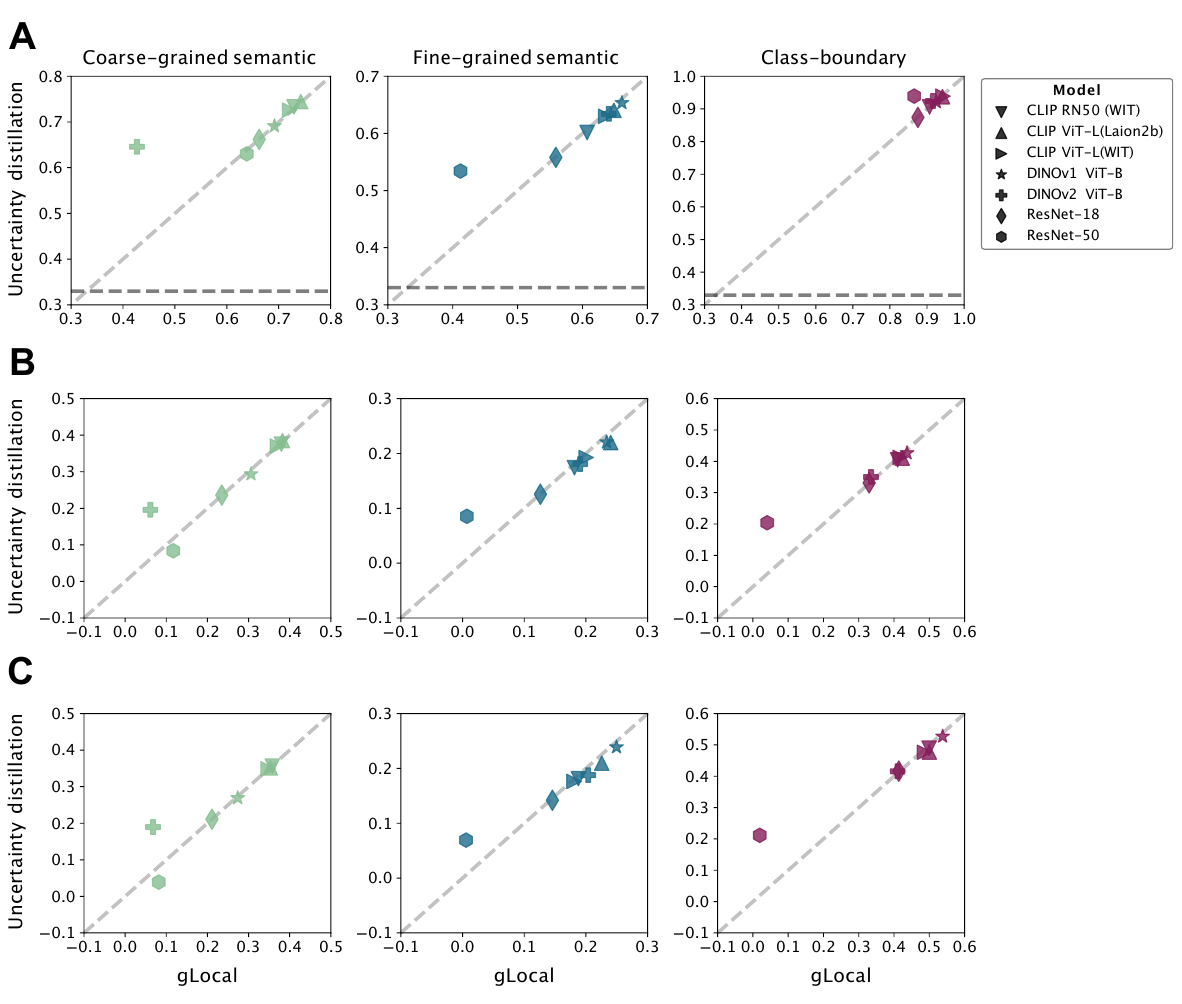}
    \caption{gLocal vs. UD on the Levels data. Here, we compare gLocal against UD for the different abstraction levels in the Levels dataset. \textbf{A}: Fraction of triplets where humans and models selected the same odd-one-out image. \textbf{B}: Spearman rank correlations between the human disagreement levels (using the five responses per triplet) and the model uncertainties over the triplets. \textbf{C}: Spearman rank correlations between the aggregated human response times (in log-space) and the model triplet uncertainties. Dashed horizontal lines in the panels of the top row indicate random guessing.}
    \label{fig:ud_vs_glocal}
\end{figure}

\noindent{\bf Uncertainty Distillation vs. gLocal.} Here, we compare the Uncertainty Distillation (UD) transformation that we introduced in SI.~\ref{appx:method-uncertainty-distill} against gLocal~\citep[cf.][]{muttenthaler2023improving}. The learning objectives of the two linear transformations are highly similar and the goal is the same: the representations of a pretrained vision foundation model are linearly transformed into a global coarse-grained human object similarity space using the THINGS dataset~\cite{hebart2020revealing} while the nearest neighbor structure of the model's representation space is preserved. However, the difference between gLocal and UD is that UD directly injects the human uncertainty estimates for the triplet odd-one-out choices via approximate Bayesian inference \citep[cf.][]{muttenthaler2022vice} into a model's representation space and, thus, uses soft rather than hard triplet choices (see SI.~\ref{appx:method-uncertainty} for details). We use the Levels datasets (see SI.~\ref{appx:levels}) for evaluating the UD and gLocal transformations, comparing the same set of models that were evaluated in \citet{muttenthaler2023improving}. The gLocal transformations for those models are publicly available via the Python library \texttt{thingsvision}~\cite{muttenthaler2021thingsvision} which we used for this analysis. In \cref{fig:ud_vs_glocal} we see that the UD transformation is either equally accurate or better in aligning the model's representations with the human similarity judgments, and predicting their disagreement levels and aggregated RTs respectively in the different abstraction level compared to gLocal. As before, human RTs are measured in log-space and model uncertainties are measured as discrete Shannon entropy over the triplet odd-one-out choices. Note that UD is computationally more efficient than gLocal because it does not require an extra set of images---in addition to the THINGS data---to evaluate the preservation of a representation's nearest neighbor structure which the multi-objective of gLocal necessitates~\citep[cf.][]{muttenthaler2023improving}.

\subsection{Qualitative Analysis of Representations}

\subsubsection{Principal Components Analysis of Representations}
\label{appx:pca}

\begin{figure}[ht!]
    \centering
    \includegraphics[width=\textwidth]{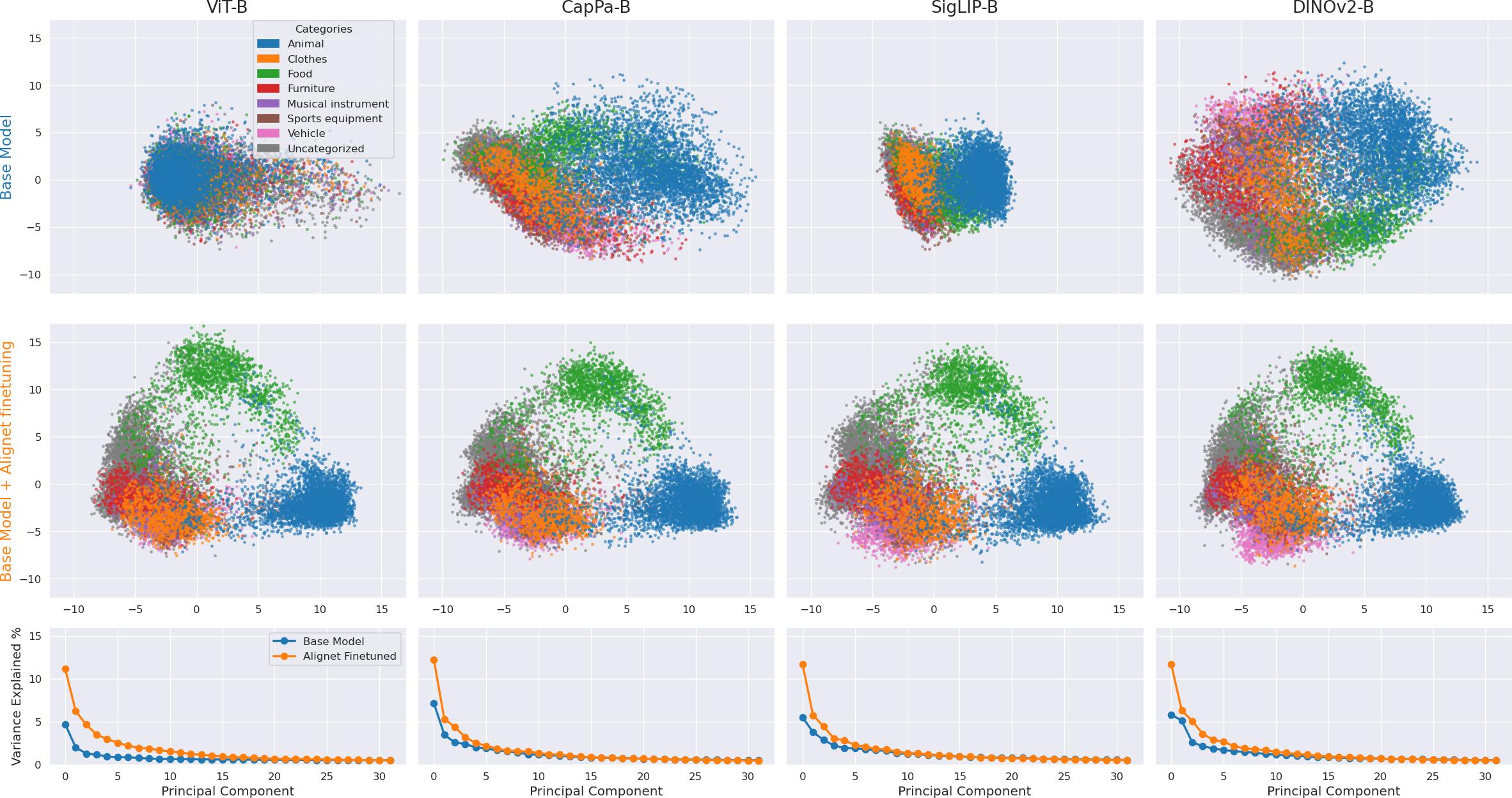}
    \caption{Projection of the representations of 85k ImageNet images from four different models (columns) onto their first two principal components both before (first row) and after fine-tuning on AligNet (second row).
    Colors correspond to high-level categories of the images.
    The third row visualizes the proportion of explained variance for the first 32 principal components.
    }
    \label{fig:pca}
\end{figure}

\begin{figure}[ht!]
    \centering
    \includegraphics[width=\textwidth]{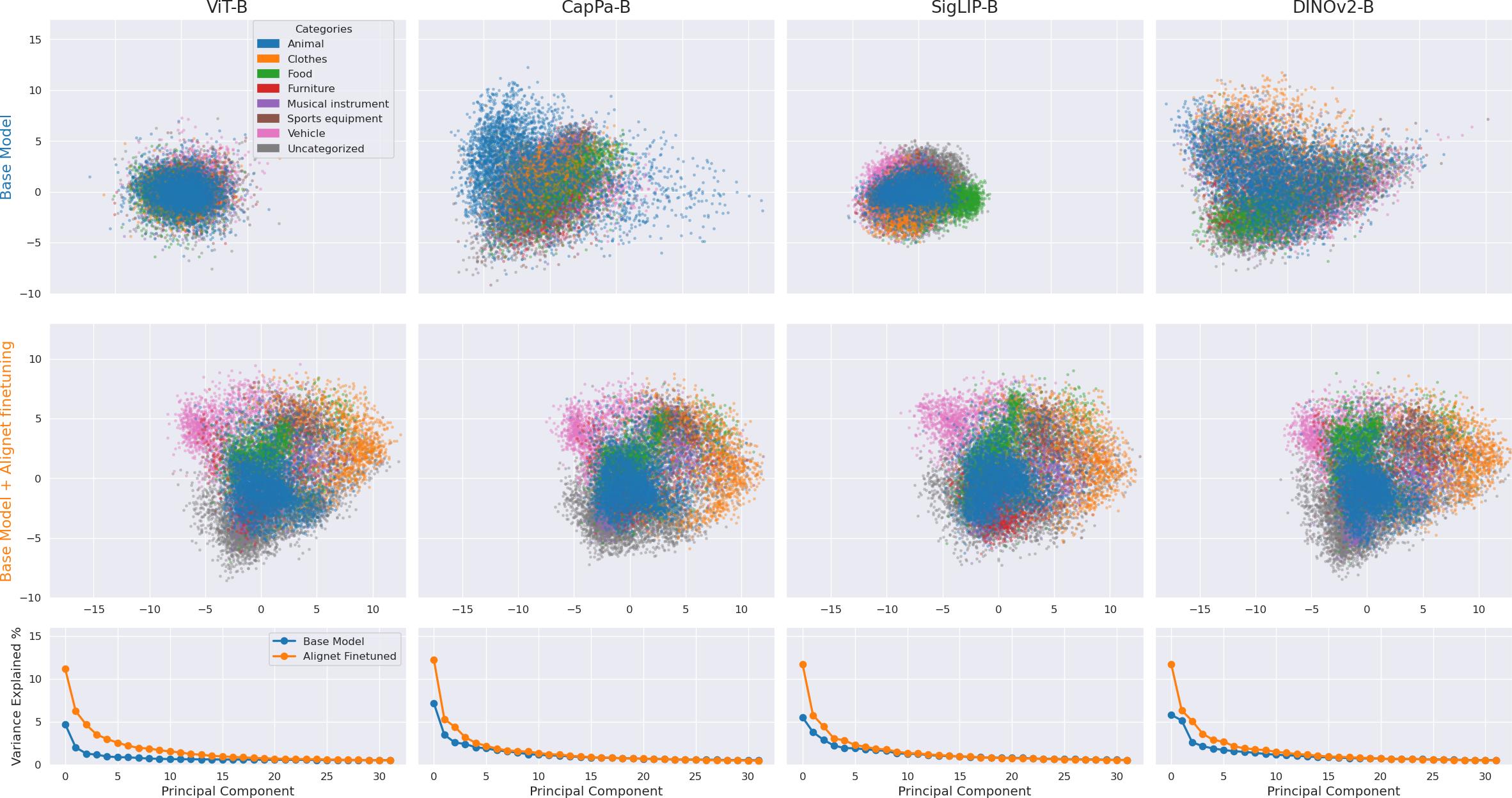}
    \caption{Projection of the representations of 85k ImageNet images from four different models (columns) onto their first two principal components both before (first row) and after fine-tuning on AligNet (second row).
    Similar to \cref{fig:pca} but for components 3 \& 4.
    }
    \label{fig:pca34}
\end{figure}

In \cref{fig:pca} we show PCA projections of the space of image representations from four different models before and after fine-tuning on AligNet.
It shows that the large-scale structure of the representational space becomes more structured and interpretable. Notice for example that animals (blue) and food (green) are clearly separated from artifacts such as furniture (red), musical instruments (purple) and clothing items (orange) which are much more clustered together. The same structuring effect --- though to a lesser degree --- can be seen for principal components three and four (see \cref{fig:pca34}).

This increased structuring within the first few principal components is also reflected in the amount of variance explained by them. 
The third row of \cref{fig:pca} clearly shows a notable increase of the proportion of the total variance of the representations that can be explained with only the first 5-15 principal components.
It is also striking how different the global structure of the four models is before fine-tuning on AligNet, and how similar it becomes afterwards.

\begin{figure}[ht!]
    \centering
    \includegraphics[width=\textwidth]{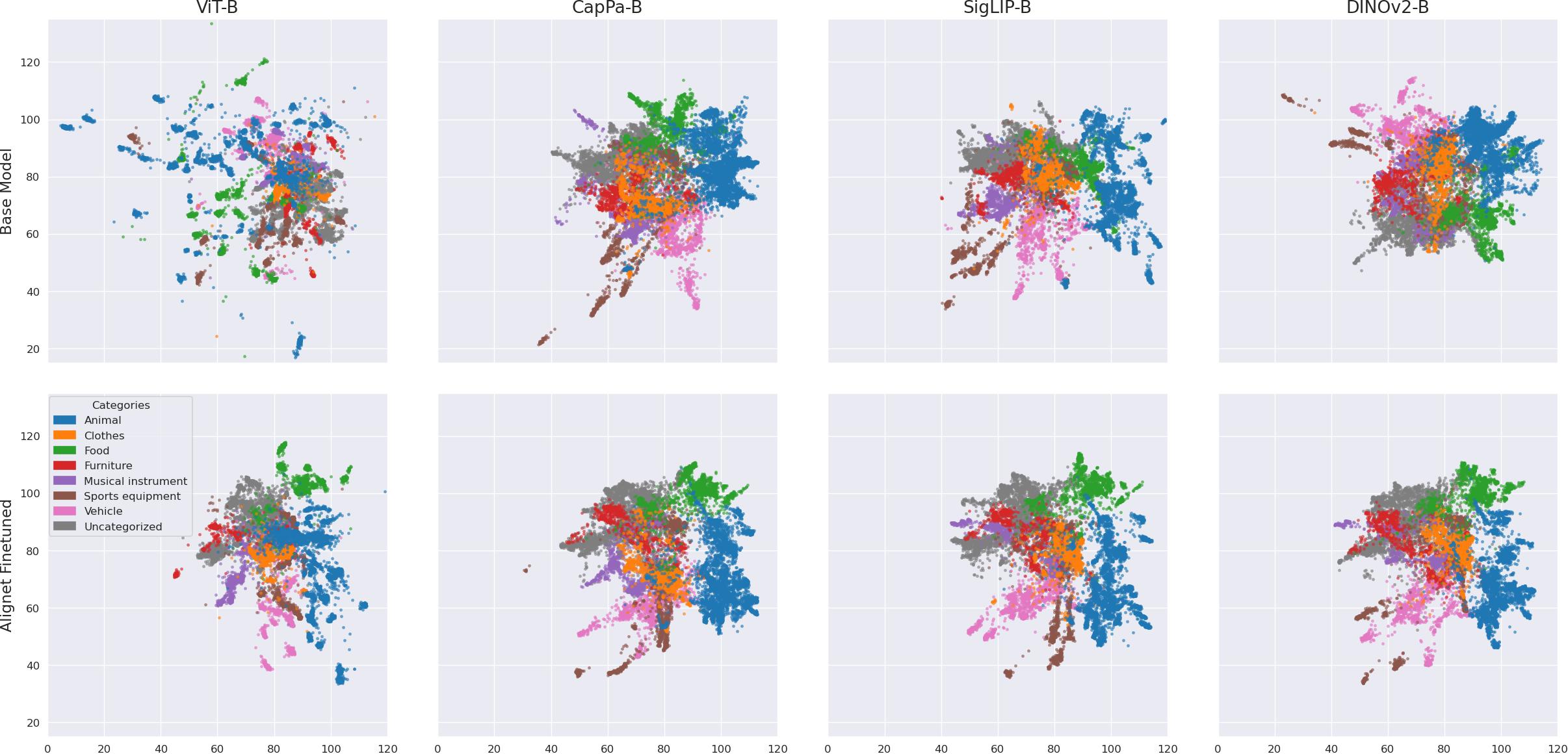}
    \caption{Projection of the representations of 85k ImageNet images from four different models (columns) onto 2 dimensions using TriMap~\cite{amid2019trimap} both before (first row) and after fine-tuning on AligNet (second row).
    Colors correspond to high-level categories of the images.}
    \label{fig:trimap}
\end{figure}

In \cref{fig:trimap} we further show projections of the representations using TriMap~\cite{amid2019trimap}, which is a non-linear dimensionality reduction technique designed to better preserve both the global and the local structure of the data.
While the coarse structure is similar to that of the first two PCs (see \cref{fig:pca}), the fine-grained structure reveals some interesting details.
Note for example that in the TriMap projection the ViT-B exhibits a lot of fine-grained structure in the form of tiny clusters.

\FloatBarrier
\subsubsection{Detailed results on alignment with the semantic hierarchy}
\label{appx:eval_hierarchy}

\begin{figure}[ht!]
    \centering
    \includegraphics[width=\textwidth]{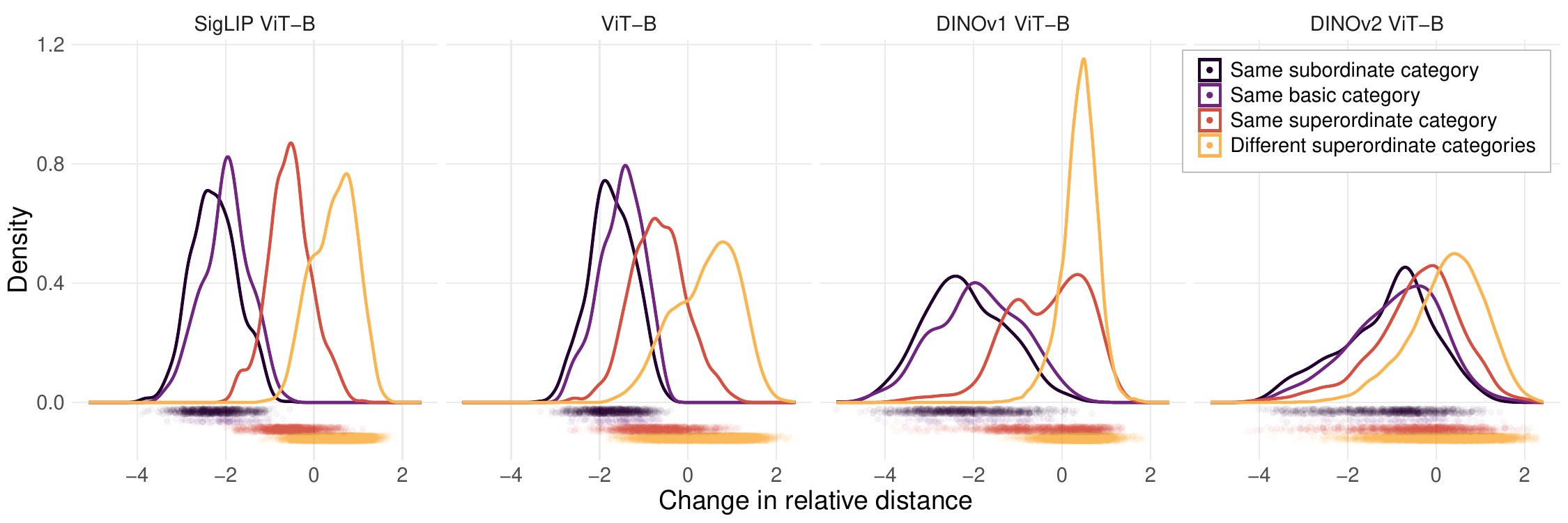}
    \caption{Changes in relative distances between stimuli reflect the superordinate category structure. Representations within the same superordinate category tend to end up relatively closer together than those in different superordinate categories. {Representations of images in the same basic-level category end up closer together, and those in the same subordinate category move even closer together.}}
    \label{fig:cluster_changes}
\end{figure}

In \cref{fig:cluster_changes} we show the distribution of changes in relative distance between the representations of pairs of ImageNet images for four models. We measure this distance change relative to other changes (by \(z\)-scoring), because relative distances are more meaningful than absolute ones (e.g., scaling all representations by a factor of two would change absolute distances, but not relative ones), and absolute scales of all representations tend to increase during training.

In general, images from classes that come from the same superordinate category (such as two different species of bird) tend to end up relatively closer together, while those from different superordinate categories tend to end up farther apart. {{Images that come from the same basic- or subordinate-level category move even closer together.}} The exact distribution of changes depends on the prior representation structure of the models; the effect is stronger for models that had lower-quality initial representations. 

\begin{figure}[htp!]
    \centering
    \includegraphics[width=\textwidth]{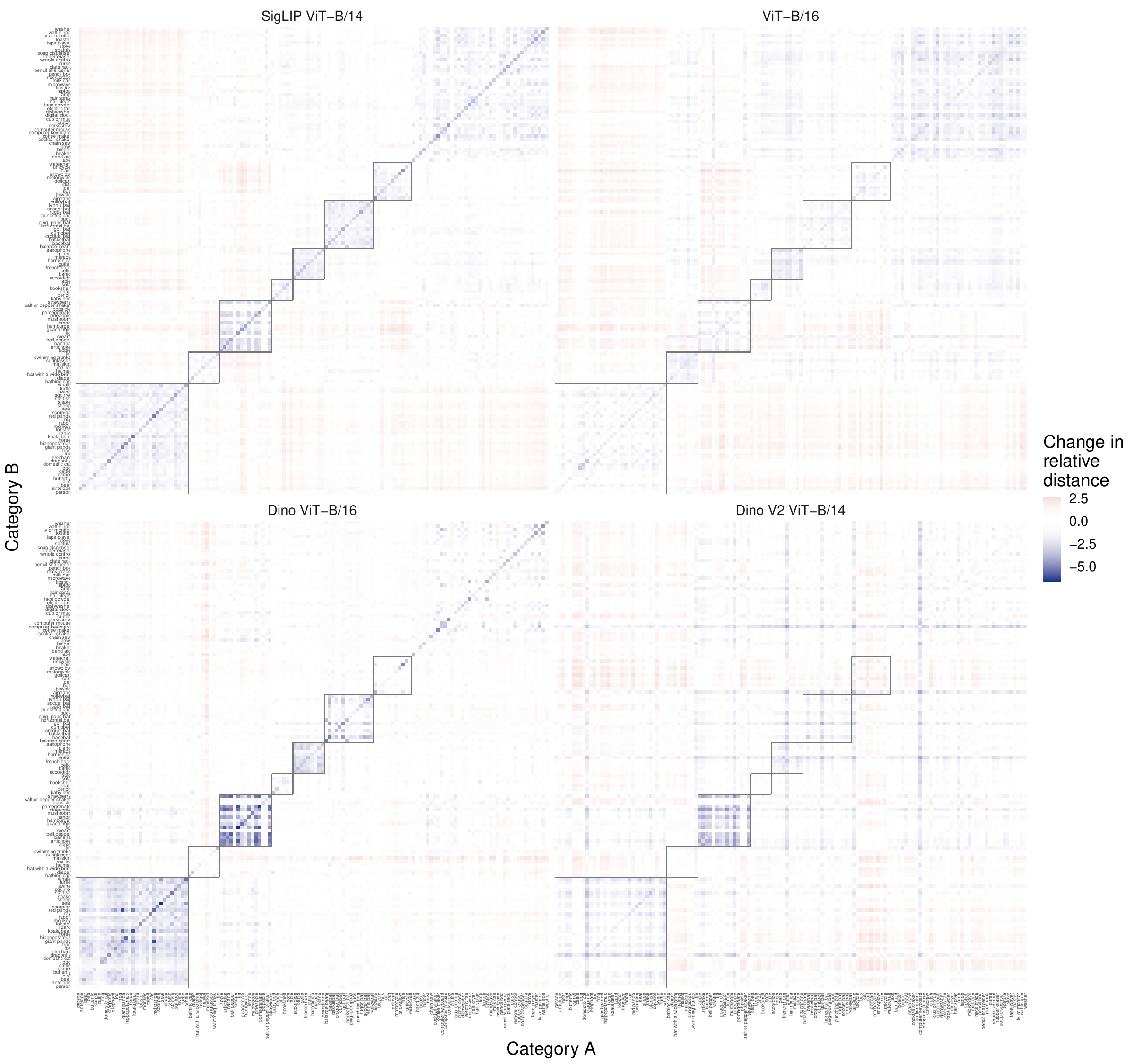}
    \caption{Changes in relative distances between stimuli are structured by higher-level semantics. After alignment, representations end up relatively more similar to one another within broad categories (diagonal blocks)---particularly human-salient ones like ``animals'' or ``food''---and less similar between these categories.}
    \label{fig:cluster_changes_matrix}
\end{figure}

In \cref{fig:cluster_changes_matrix} we show a more detailed visualization, showing a matrix of average changes in the relative distances between a subset of the \emph{basic-level} categories, grouped into several higher-level semantic categories (animals, clothes, food, furniture, musical instruments, sports equipment, and vehicles), which are plotted as blocks on the diagonals. The block structure of the changes is clear, with overall increases in similarity within these broad categories, and decreases in similarity between them.

Taken together, these results show qualitatively that our alignment process is working as intended---it is reorganizing the representation space of the model in accordance with the structure of human semantic knowledge.

{{
\textbf{Verifying the statistical significance of the changes.}
To quantify the statistical significance of the representational changes, we fit mixed-effects linear regressions accounting for the non-independence of the representational changes across different clusters. We use the same-subordinate condition as the reference level, as this offers the most stringent comparison for the same-basic condition (i.e. the effect size of the difference between the two conditions is visibly the smallest). All effects were generally highly significant, full results are presented in \cref{tab:cluster_changes_stats}.
\begin{table*}[htbp] 
\begin{subtable}{\textwidth}
\begin{Verbatim}[fontsize=\tiny]
                                                 Estimate Std. Error          df   t value  Pr(>|t|)    
(Intercept)                                      -2.25018    0.01612  2344.28109  -139.562   < 2e-16 ***
relationshipSame basic category                   0.25678    0.04675  2344.28108     5.493  4.39e-08 ***
relationshipSame superordinate category           1.62585    0.03372   619.98263    48.220   < 2e-16 ***
relationshipDifferent superordinate categories    2.72148    0.03229   520.51180    84.275   < 2e-16 ***
\end{Verbatim}
\caption{{SigLIP-B}}
\end{subtable}
\begin{subtable}{\textwidth}
\begin{Verbatim}[fontsize=\tiny]
                                                 Estimate Std. Error         df  t value  Pr(>|t|)    
(Intercept)                                      -1.73190    0.01931 2280.89874  -89.667   < 2e-16 ***
relationshipSame basic category                   0.22030    0.05600 2280.89872    3.934  8.62e-05 ***
relationshipSame superordinate category           0.98145    0.04498  963.74930   21.818   < 2e-16 ***
relationshipDifferent superordinate categories    2.13518    0.04384  870.21854   48.701   < 2e-16 ***
---
Signif. codes:  0 ‘***’ 0.001 ‘**’ 0.01 ‘*’ 0.05 ‘.’ 0.1 ‘ ’ 1
\end{Verbatim}
\caption{{ViT-B}}
\end{subtable}
\begin{subtable}{\textwidth}
\begin{Verbatim}[fontsize=\tiny]
                                                 Estimate Std. Error          df  t value  Pr(>|t|)    
(Intercept)                                      -2.18517    0.02733  1127.92111   -79.95   < 2e-16 ***
relationshipSame basic category                   0.29318    0.07925  1127.92113     3.70  0.000226 ***
relationshipSame superordinate category           1.75272    0.07325   819.54322    23.93   < 2e-16 ***
relationshipDifferent superordinate categories    2.59777    0.07268   793.50251    35.74   < 2e-16 ***
\end{Verbatim}
\caption{{Dino ViT-B}}
\end{subtable}
\begin{subtable}{\textwidth}
\begin{Verbatim}[fontsize=\tiny]
                                                 Estimate Std. Error          df  t value  Pr(>|t|)    
(Intercept)                                      -1.11179    0.03202  1312.37727  -34.719   < 2e-16 ***
relationshipSame basic category                   0.19989    0.09285  1312.37733    2.153    0.0315 *  
relationshipSame superordinate category           0.63283    0.08885  1098.25627    7.122  1.92e-12 ***
relationshipDifferent superordinate categories    1.40199    0.08847  1079.24964   15.847   < 2e-16 ***
\end{Verbatim}
\caption{{Dino V2 ViT-B}}
\end{subtable}
\caption{{Statistical significance of the representation reorganization for the AligNet models, via mixed-effects linear regressions. (Note that same-subordinate category is the reference level; thus, the statistics on the intercept denote the magnitude and significance of the changes in relative distances for that category, and the other coefficients and statistics are relative to that change.}} \label{tab:cluster_changes_stats}
\end{table*}

\textbf{Reorganization across living and nonliving categories}
Given that the living vs. nonliving distinction is extremely salient to humans, and likely plays a large role in the odd-one-out-judgments, we also computed similar analyses to the above focusing specifically on this distinction. Specifically, we evaluated how representation distances change within- and between the living and nonliving categories. We show the results in \cref{fig:living_nonliving_cluster_changes}---the changes are qualitatively as expected. These differences are also all statistically significant for all models (all \(t\)s \(>2.7\), all \(p\)s \(< 0.01\)) in mixed-effects regressions.

\begin{figure}[ht!]
    \centering
    \includegraphics[width=\textwidth]{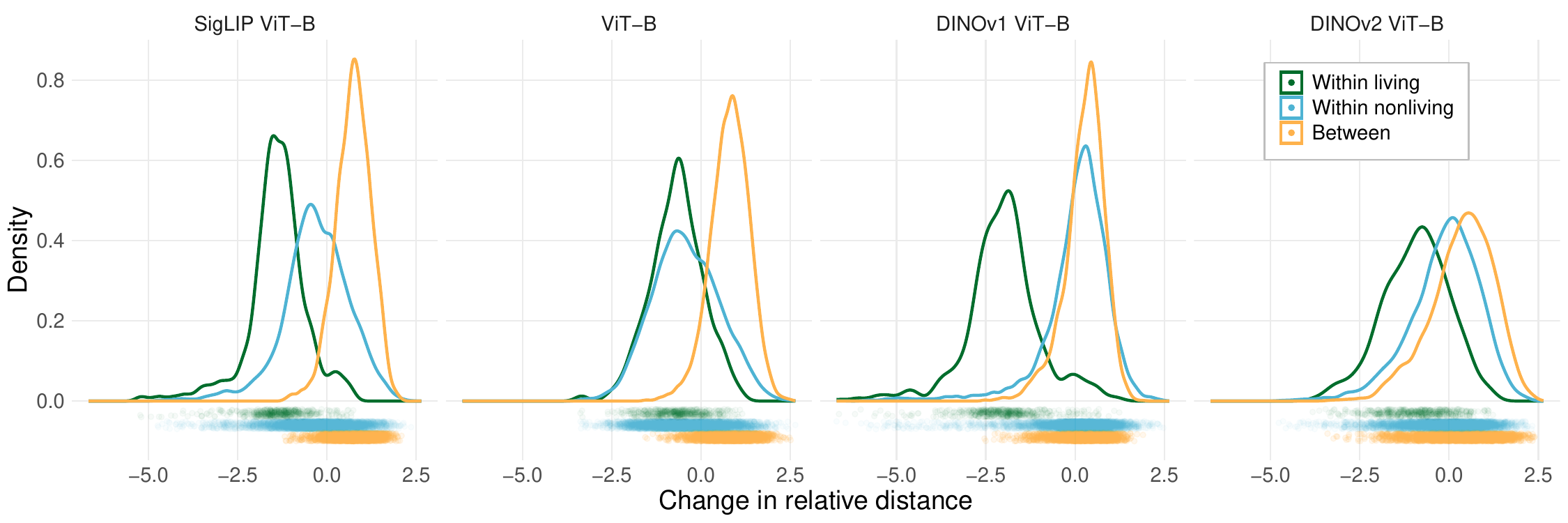}
    \caption{{Changes in relative distances between living and nonliving entities after alignment. Living entities tend to cluster more closely together after alignment, nonliving entities do so more weakly, and the two categories tend to move apart from one another.}}
    \label{fig:living_nonliving_cluster_changes}
\end{figure}

\textbf{Alignment across layers}
Since our alignment procedure focuses on aligning the final representations of the model, it is not obvious that we would see similar reorganization at earlier layers. We therefore performed similar representation-change analyses to the above across the representation layers of the model (using the mean representation across spatial positions). We show the results in \cref{fig:cluster_changes_by_layer}---generally, we see a growing degree of reorganization across the layers of the model.

\begin{figure}[ht!]
    \centering
    \includegraphics[width=0.6\textwidth]{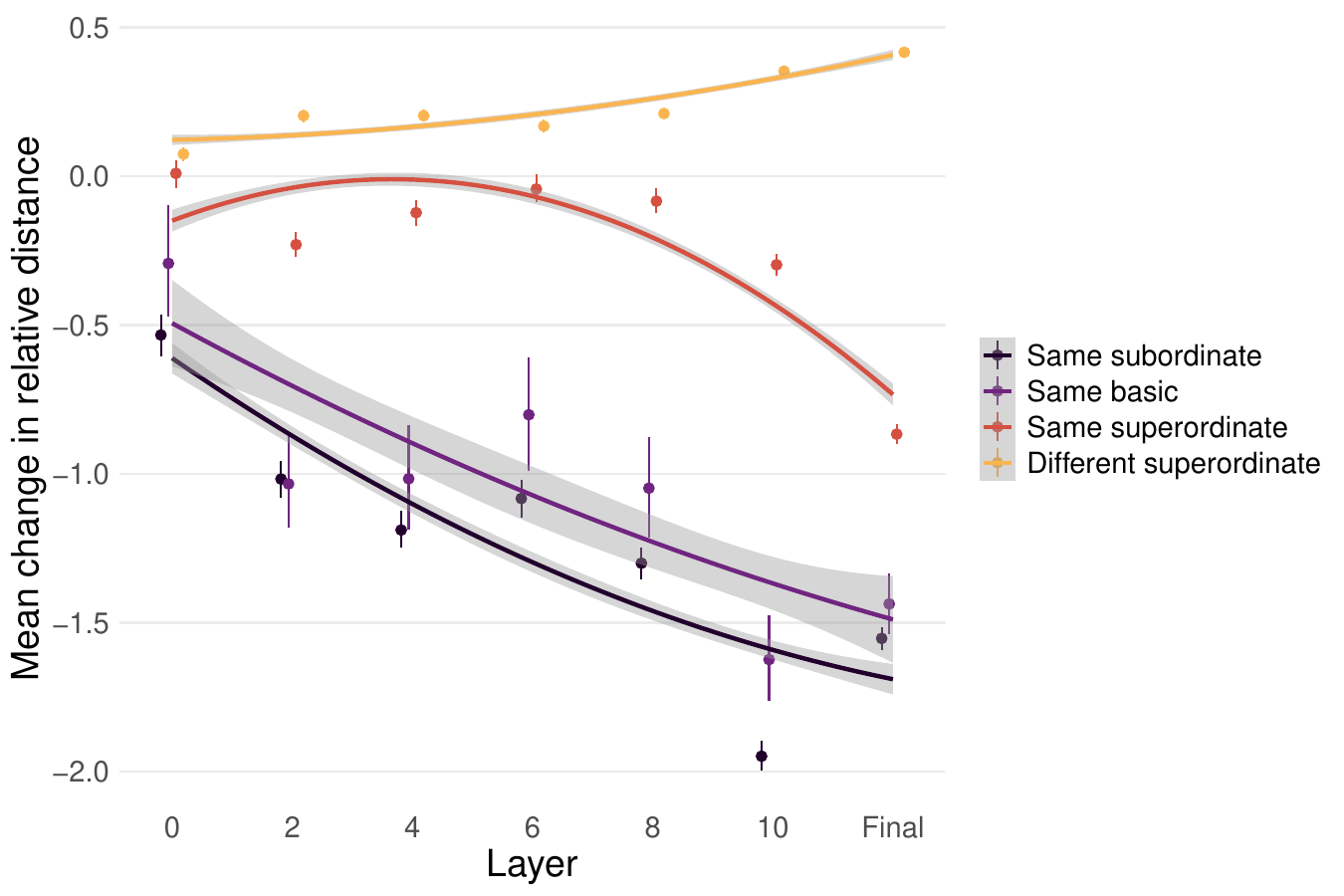}
    \caption{{Changes in relative distances between stimuli across the layers of the ViT-B/16 model. We see generally consistent directions of the effects across layers, with the magnitude of the changes increasing at higher layers. (Errorbars on points are bootstrap 95\%-CIs. Curves are quadratic regression fits, to illustrate the overall trends.)}}
    \label{fig:cluster_changes_by_layer}
\end{figure}

\textbf{Comparing representation changes to UnaligNet}
Finally, we compare the representation changes in the UnaligNet ablation (see above) for ViT-B. We show the results in \cref{fig:unalignet_cluster_changes}. This reorganization is significantly different from that observed in the AligNet version (\cref{tab:unalignet_cluster_changes_stats}). Surprisingly, the UnaligNet procedure results in \emph{greater} divergence amongst images within the same basic- or subordinate-level category, compared to the original representations. To gain some qualitative insight into this effect, we compared pairs of images within the same subordinate category on which AligNet and UnaligNet maximally disagree (\cref{fig:unalignet_alignet_max_disagreement}). The results suggest that perhaps UnaligNet representations are focusing more on superficial features that distinguish images, such as color or style. We validate these results with a human experiment in the next section.}}

\begin{figure}[ht!]
    \centering
    \includegraphics[width=0.5\textwidth]{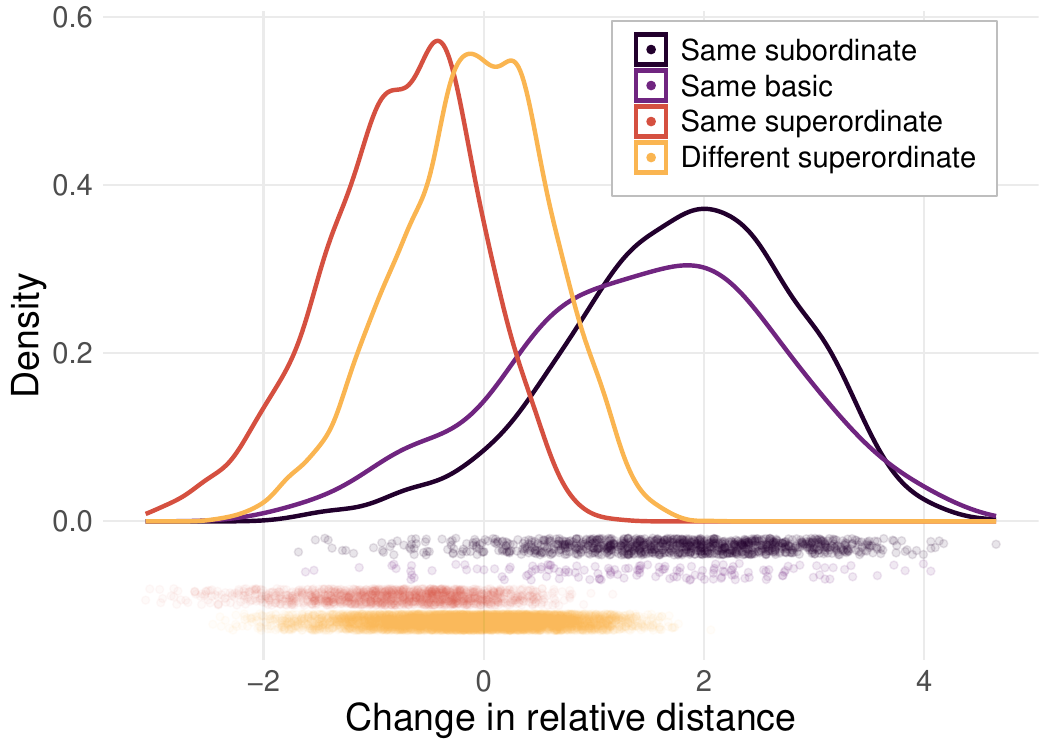}
    \caption{{Changes in relative distances between stimuli in the UnaligNet ViT-B. We see reorganization at the basic and subordinate levels that is the \emph{opposite} of what would be desired; viz. these categories are moving farther apart, with changes that are larger in magnitude than the changes in the superordinate categories.}}
    \label{fig:unalignet_cluster_changes}
\end{figure}

\begin{table*}[htbp] 
\begin{subtable}{\textwidth}
\begin{Verbatim}[fontsize=\tiny]
                                                         Estimate   Std. Error         df t value  Pr(>|t|)    
(Intercept)                                               1.776e+00  2.470e-02  1.948e+03   71.89   < 2e-16 ***
versionAligNet                                           -3.328e+00  2.013e-02  1.906e+04 -165.38   < 2e-16 ***
relationshipSame basic                                   -3.288e-01  7.162e-02  1.948e+03   -4.59  4.71e-06 ***
relationshipSame superordinate                           -2.606e+00  6.175e-02  1.076e+03  -42.20   < 2e-16 ***
relationshipDifferent superordinate                      -1.877e+00  6.086e-02  1.015e+03  -30.84   < 2e-16 ***
versionAligNet:relationshipSame basic                     4.447e-01  5.836e-02  1.906e+04    7.62  2.65e-14 ***
versionAligNet:relationshipSame superordinate             3.232e+00  2.573e-02  1.906e+04  125.60   < 2e-16 ***
versionAligNet:relationshipDifferent superordinate        3.846e+00  2.143e-02  1.906e+04  179.48   < 2e-16 ***
\end{Verbatim}
\end{subtable}
\caption{{Statistical significance of the differences in representation changes between UnaligNet and AligNet (interaction of changes by training version). All interactions are highly significant. (Note that the relationship is dummy coded, so the simple effect of versionAligNet shows the difference in the reorganization of the same-subordinate category between the model versions.)}} \label{tab:unalignet_cluster_changes_stats}
\end{table*}

\begin{figure}[ht!]
    \centering
    \begin{subfigure}{0.33\textwidth}
    \centering
    \includegraphics[width=0.45\linewidth]{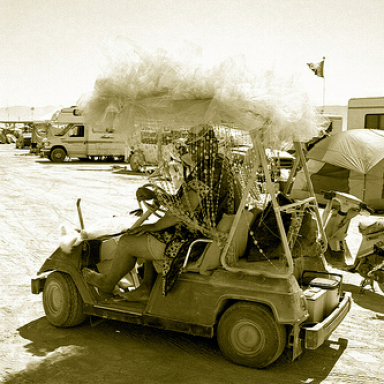}%
    \includegraphics[width=0.45\linewidth]{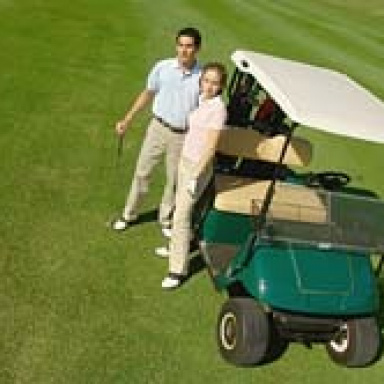}%
    \caption{Golf cart}
    \end{subfigure}%
    \begin{subfigure}{0.33\textwidth}
    \centering
    \includegraphics[width=0.45\linewidth]{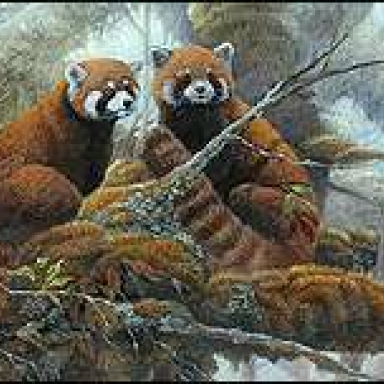}%
    \includegraphics[width=0.45\linewidth]{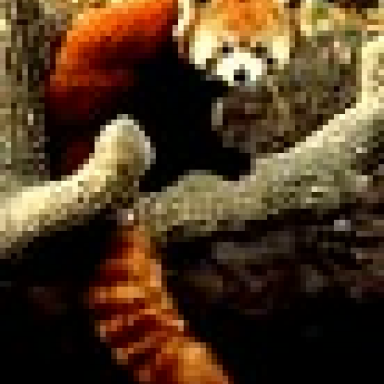}%
    \caption{Red panda}
    \end{subfigure}%
    \begin{subfigure}{0.33\textwidth}
    \centering
    \includegraphics[width=0.45\linewidth]{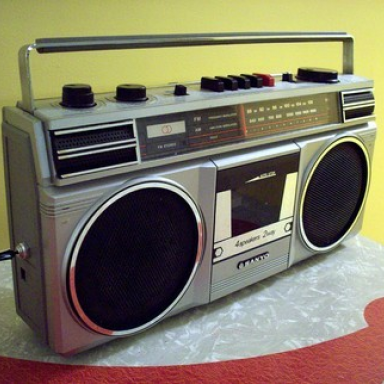}%
    \includegraphics[width=0.45\linewidth]{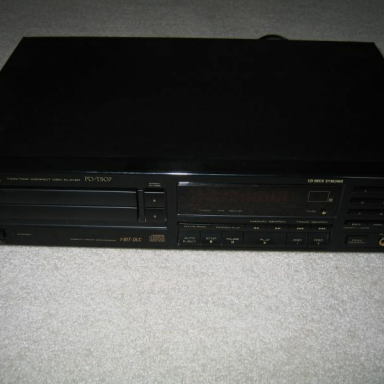}%
    \caption{Tape player}
    \end{subfigure}%
    \caption{{Pairs of same-category images for which the representations of AligNet and UnaligNet maximally disagree. AligNet groups these images more closely together due to their shared semantic content, whereas UnaligNet represents them fairly differently, perhaps on the basis of more superficial features like color vs. black \& white, pixelization, style, etc.}}
    \label{fig:unalignet_alignet_max_disagreement}
\end{figure}

\FloatBarrier
{
\subsubsection{Human validation of the representational differences between AligNet and UnaligNet}
\label{appx:alignet_unalignet_disagreement_humans}

In order to further ascertain the origin of the differences between the AligNet and UnaligNet tuned versions of ViT-B, we compared representations within each ImageNet category to find the largest disagreements between the two models. Specifically, within each category we identified the pair of images for which the models' representations most strongly disagreed in both directions: the pair that the AligNet model represented much more similary than UnaligNet, and the pair that UnaligNet represented much more similarly than AligNet (Note that in this context, since both models start from the same base model, it is equivalent to compare similarities directly or to compare the change in similarity from the base model.) We took the top 100 disagreements (in magnitude) in both directions across categories, for a total of 200 image pairs --- half of which AligNet represents much more similarly than UnaligNet, and half of which UnaligNet represents much more similarly than AligNet.

We then ran a human online experiment via Prolific (\(N = 49\) participants) testing which images participants rated (on a 1-5 Likert scale) as semantically more similar in content, when explicitly instructed to ignore superficial visual features. We find strong alignment between the participants' judgments and the AligNet model --- the participants consistently gave higher similarity ratings to the images that AligNet represented as more similar. 
This effect was highly significant (paired \(t\)-test: \(t(48) = 20.260, p < 0.001\)); in fact, \emph{every} participant gave numerically higher similarity scores to the pairs AligNet rated as more similar. Most importantly, the correlation between model- and human-judged similarity is strongly positive for the aligned model (Spearman =0.641, p<.001) and negative for the unaligned model (Spearman =–0.122, p = 0.087).

These strong results support our claim that AligNet is uniquely changing the model's representation structure---even within low-level categories---to align more closely with human semantic knowledge.
}

\begin{figure}[ht!]
    \centering
    \includegraphics[width=0.7\textwidth]{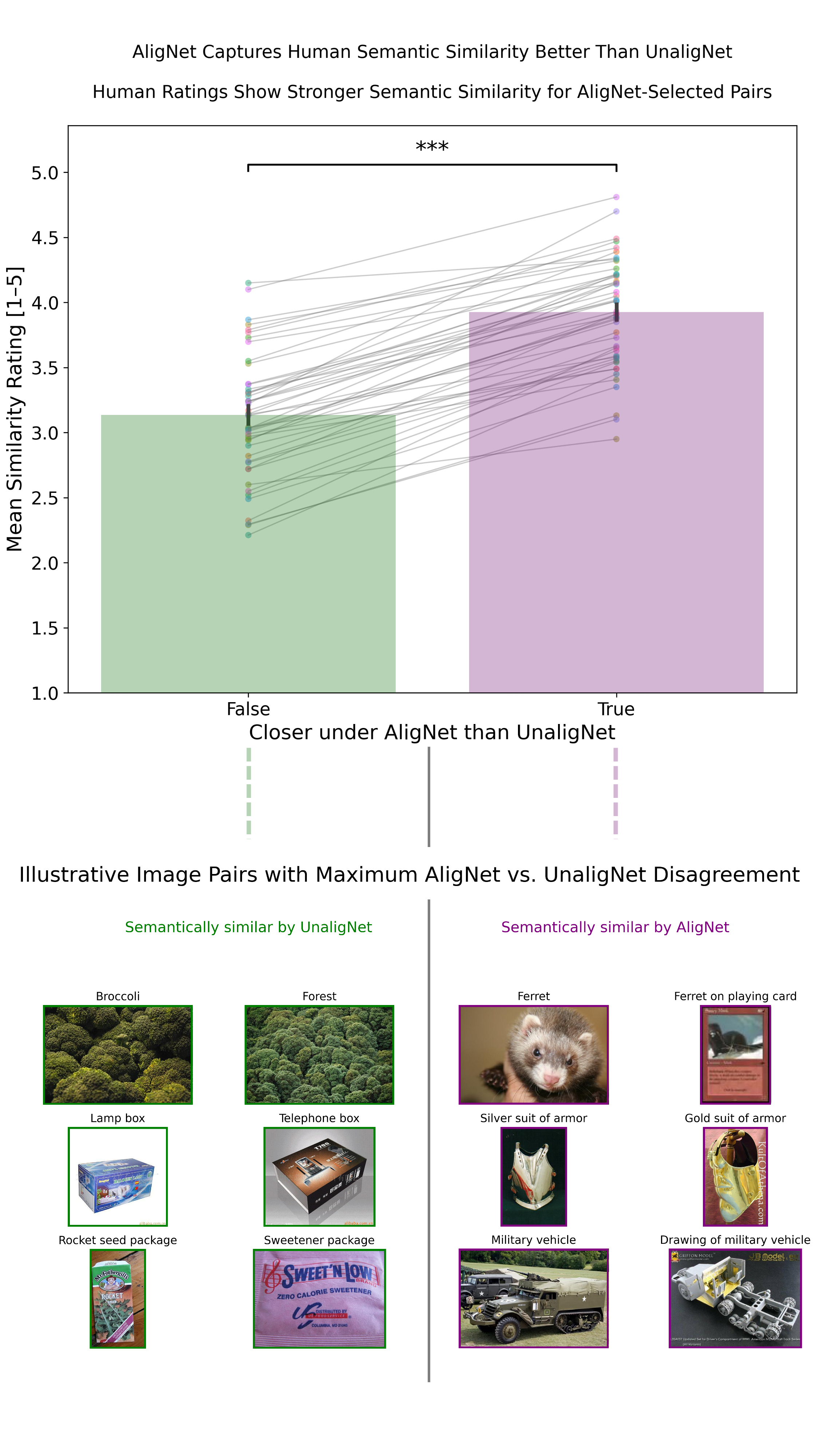}
    \caption{ Human ratings of semantic similarity on image pairs on which AligNet and UnaligNet maximally disagree. The human participants consistently rate the pairs which AligNet represents more similarly as in fact more semantically similar. Shown on the right are the three image pairs that received the lowest human similarity ratings, all of which contradict UnaligNet's prediction of high similarity.}
    \label{fig:alignet_unalignet_human_ratings}
\end{figure}

\FloatBarrier

\end{document}